\documentclass[letterpaper]{article} 
\usepackage[preprint]{aaai2027}  
\usepackage[hyphens]{url}  
\usepackage{graphicx} 
\usepackage{natbib}  
\usepackage{caption} 
\usepackage{algorithm}
\usepackage{algorithmic}

\usepackage{amssymb}
\usepackage{amsmath}
\usepackage{xcolor}

\usepackage{newfloat}
\usepackage{listings}
\DeclareCaptionStyle{ruled}{labelfont=normalfont,labelsep=colon,strut=off} 
\floatstyle{ruled}
\newfloat{listing}{tb}{lst}{}
\floatname{listing}{Listing}

\usepackage{booktabs}
\usepackage{multirow}

\title{Instruction-Conditioned Electromagnetic Spectrum Understanding via Budget-Adaptive Signal Tokenization}
\author{Lei Zhai, Zhihao Chang, Shuyuan Yang\thanks{Corresponding author: Shuyuan Yang.}, Zhixi Feng}
\affiliations{
School of Artificial Intelligence, Xidian University, Xi'an, China\\
zhailei@stu.xidian.edu.cn, \{changzhihao, syyang, zxfeng\}@xidian.edu.cn
}

\graphicspath{{Figures/}{supplementary/figures/}}

\begin{document}

\maketitle

\begin{abstract}
Electromagnetic spectrum monitoring increasingly requires flexible analysis beyond task-specific recognition and detection. Multimodal large language models offer a unified interface, but extending vision-language models (VLMs) to raw I/Q signals requires tokenization that balances fidelity against a strict budget. For signals, dense encoding causes token costs to grow with observation length, whereas fixed-resolution compression may discard short-duration or localized signal evidence. Thus, we propose \textbf{BATok}, a budget-adaptive signal tokenizer that adjusts token capacity to the input length while allocating that capacity according to the signal content. BATok constructs candidate representations from signal-derived features using lightweight multi-resolution branches, then combines a local energy prior with learnable queries to resample these representations into compact signal tokens. The number of tokens adapts to the input length while remaining strictly bounded. The resulting tokens are projected into the language embedding space of VLMs. We further introduce \textbf{EMSpec-Instruct}, a multimodal instruction dataset aligning raw I/Q signals, waterfall images, and language supervision for modulation recognition, structured detection, and language-conditioned signal grounding. Experiments show that BATok learns effective signal representations and achieves competitive performance across all tasks.

\end{abstract}

\section{Introduction}
\begin{figure}[t]
	\centering
	\includegraphics[width=\linewidth]{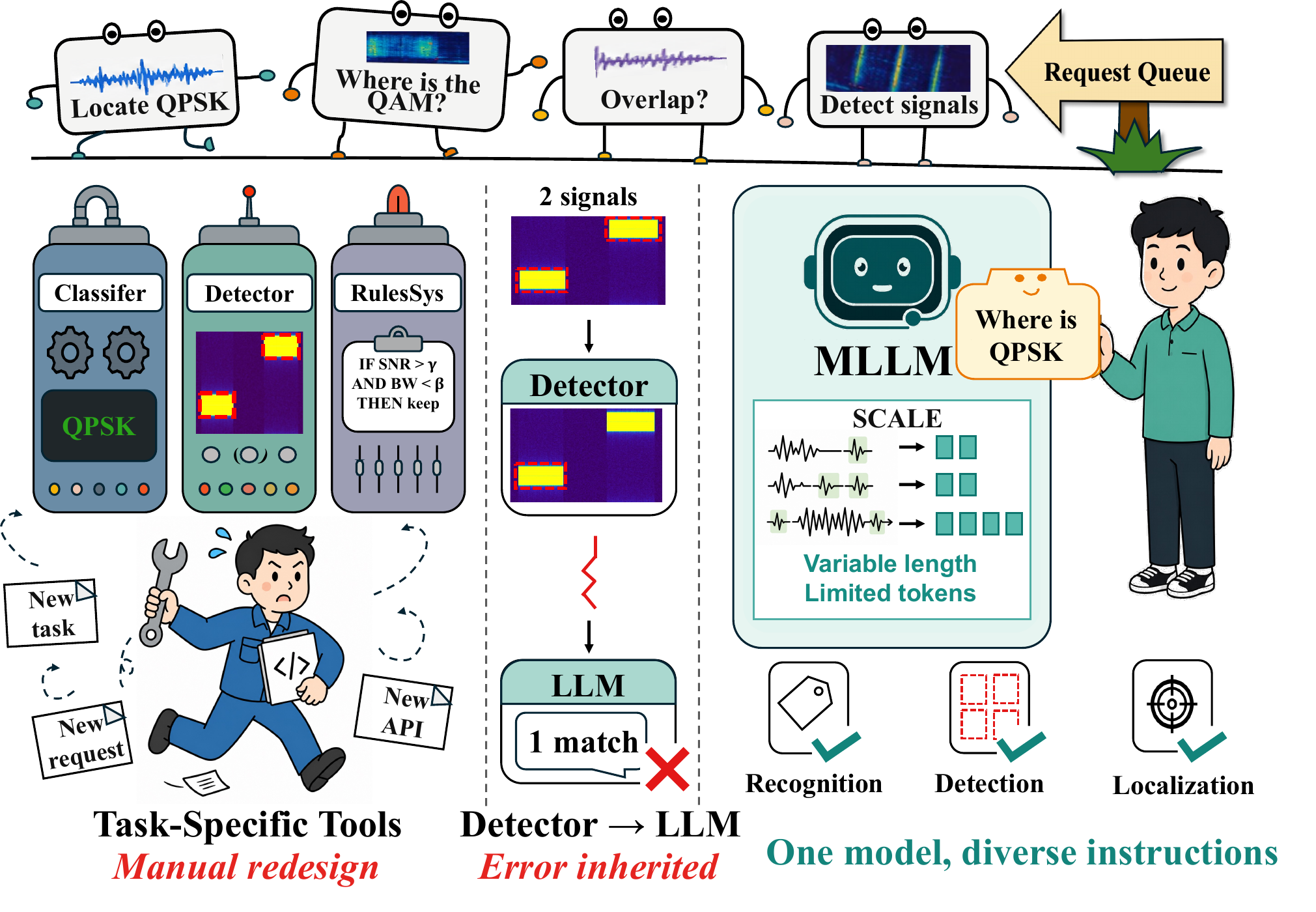}
	\caption{Motivation for unified instruction-conditioned spectrum
		understanding. Task-specific models require manual redesign, while detector--LLM cascades inherit upstream errors. Our framework couples signal tokenization with a VLM through a unified interface.}
	\label{fig:intro}
\end{figure}

Electromagnetic spectrum monitoring supports communications, spectrum
management, interference analysis, and dynamic spectrum
access~\cite{yucek2009survey,ding2018spectrum}. Beyond recognizing
predefined signal classes, practical monitoring also requires detecting,
localizing, and characterizing signals of interest across time and
frequency~\cite{ding2018spectrum,boegner2022wideband}. These practical
demands call for spectrum analysis systems that can flexibly respond to
different information needs rather than producing only a fixed set of
predictions. Figure~\ref{fig:intro} contrasts this unified objective with
task-specific pipelines.

Existing spectrum-analysis methods generally use task-specific
encoders with dedicated classification or detection
heads~\cite{oshea2016convolutional,shalu2026deepamc,boegner2022wideband}.
Although effective for individual objectives, they expose separate
interfaces for recognition, detection, and localization. Supporting
additional queries therefore typically requires manually designed
rules, post-processing, or new model components. A detector--LLM
cascade can provide a natural-language interface, but its answers
remain constrained by the labels, boxes, and attributes produced by
the upstream detector. In contrast, multimodal large language models
(MLLMs) provide a general interface for multimodal perception,
reasoning, and language generation~\cite{yin2024mllmsurvey, jiang2025largeai}.

Building on recent advances in audio-language
models~\cite{kimiteam2025kimiaudio,yang2025audiollmsurvey},
video-language
models~\cite{zhang2025videollama3,tang2026videosurvey},
and time-series--language
models~\cite{wang2025chattime,liu2025crossmodalitysurvey},
recent RF-oriented MLLMs connect electromagnetic observations to
language through visual representations, including waveform plots,
spectra, spectrograms, waterfall images, and constellation
diagrams~\cite{zou2026seeingradio,zou2026rfgpt,han2026pred,
	bara2026rfanalyzer}. However, these vision-aligned interfaces convert
numerical I/Q samples into fixed-resolution images before tokenization.
Such conversion may attenuate waveform-level amplitude, phase, and
transient evidence, while the visual token capacity is determined by
image resolution rather than the duration and content of the original
observation. Moreover, generic audio and time-series tokenizers are not
designed around the coupled physical structure of I/Q samples.

Existing signal encoders achieve strong performance on task-specific
signal recognition using CNN-based~\cite{du2022dncnet},
RNN-based~\cite{yao2025shortwave},
Transformer-based~\cite{zhai2025cdsit}, and
state-space-based~\cite{10705364, ma2025dwmtn} architectures.
Nevertheless, they typically produce global representations for fixed
prediction heads or temporal features whose granularity is determined by
predefined strides, pooling ratios, or patch sizes. This creates a capacity
mismatch for mixed-length observations: dense representations become
expensive for long signals, whereas a uniformly small output budget may
discard localized evidence. Consequently, existing encoders generally do
not adapt their output granularity jointly to input length and information
distribution. Integrating raw I/Q into an MLLM therefore requires a tokenizer
that adapts its representation capacity to signal length and content while
keeping the output strictly bounded.

To address these challenges, we develop an instruction-conditioned multimodal framework for electromagnetic spectrum understanding. At its core, we introduce \textbf{BATok}, a signal tokenizer that adapts the total token capacity to the input length and adapts temporal aggregation to the signal content for VLM conditioning.
BATok first constructs candidate signal representations using signal-derived amplitude, power, and phase features together with lightweight multi-resolution temporal branches. It performs evidence-guided query resampling to compress variable-length features into compact signal tokens, while a monotonic token-budget function allocates larger yet strictly bounded representation capacity to longer observations. The resulting tokens are projected into the language-model embedding space and jointly processed with waterfall-image tokens and textual instructions. To provide supervision for this setting, we construct \textbf{EMSpec-Instruct}, a multimodal instruction dataset that aligns raw I/Q observations and waterfall images with modulation recognition, structured detection, and language-conditioned signal grounding.

Our main contributions are summarized as follows:

\begin{itemize}
\item We construct \textbf{EMSpec-Instruct}, a multimodal instruction dataset aligning I/Q signals, waterfall images, and language supervision for modulation recognition, structured detection, and language-conditioned signal grounding.

\item We propose \textbf{BATok}, a lightweight signal tokenizer with a compact parameter footprint. It uses multi-resolution evidence-guided query resampling to convert mixed-length I/Q observations into informative tokens with an adaptive yet strictly bounded budget.

\item We integrate BATok with a VLM through continuous signal-token projection. Experiments  demonstrate the effectiveness and show that I/Q signals provide complementary waveform-level evidence beyond waterfall images.

\end{itemize}

\section{Related Work}
\label{sec:supp-related-work}

\subsection{Representation Learning from Raw I/Q Signals}
\label{sec:supp-iq-representation}

Deep automatic modulation classification established that useful
radio representations can be learned directly from complex I/Q
sequences. CNN-based methods learn local temporal and cross-channel
patterns~\cite{oshea2016convolutional,xu2020spatiotemporal}; DNCNet
further couples convolutional denoising and classification for robust
recognition~\cite{du2022dncnet}. Transformer-based models capture
long-range dependencies, as represented by SiT's dual-domain
interaction and IQFormer's multi-representation sequence
modeling~\cite{zhai2025cdsit,shao2025iqformer}. More recently, MAMC and
DWMTN employ selective state-space modeling, alone or with
Transformers, for efficient long-sequence processing
~\cite{10705364,ma2025dwmtn}; self-supervised methods such as RIS-MAE
learn transferable features through masked reconstruction
~\cite{liu2025rismae}. A broader survey is provided
in~\cite{shalu2026deepamc}. These architectures demonstrate the value of waveform-level
amplitude, phase, and temporal evidence, but predominantly produce
global or fixed-granularity features for task-specific heads. BATok
instead converts mixed-length I/Q observations into a bounded
continuous-token sequence suitable for multimodal language models.

\subsection{Connecting Non-Textual Sequences to Language Models}
\label{sec:supp-sequence-llm}

Non-textual observations are commonly connected to LLMs through
patch-based or query-based interfaces. Audio and video MLLMs use
learned cross-modal connectors~\cite{tang2024salmonn,zhang2023videollama},
while Time-LLM, ChatTime, and ChatTS introduce numerical sequence
representations for forecasting, understanding, and reasoning
~\cite{jin2024timellm,wang2025chattime,xie2024chatts}. These methods
establish the feasibility of continuous inputs, but generic sequence
interfaces do not explicitly preserve coupled I/Q evolution; moreover,
fixed-size patches yield token counts that grow with duration. Query-based resamplers instead map variable encoder features to a
fixed token set, as in Flamingo and BLIP-2
~\cite{alayrac2022flamingo,li2023blip2}. This bounds the interface but
assigns equal capacity to observations of different lengths. BATok
retains a bounded interface while adapting token count to valid length
and allocating that budget using multi-resolution features, learned
relevance, and an energy prior.

\subsection{RF-Oriented Language and Multimodal Models}
\label{sec:supp-rf-mllms}

Signal-oriented language and multimodal models follow two main directions.
Numerical approaches adapt language-model backbones to radio signal tasks:
RadioLLM reprograms I/Q patches and combines language and convolutional
features for classification and reconstruction, retaining a
representation-plus-task-head formulation~\cite{chen2025radiollm};
Large Spectrum Models discretize spectrum measurements for
decoder-only forecasting~\cite{lunar2026lsm}. Visual approaches reuse VLM interfaces by rendering RF observations.
Seeing Radio uses I/Q traces and spectrograms for modulation VQA
~\cite{zou2026seeingradio}; RF-GPT, PReD, and RF-Analyzer extend this
paradigm to broader spectrum understanding, electromagnetic tasks, and
synthetic-to-real waterfall analysis
~\cite{zou2026rfgpt,han2026pred,bara2026rfanalyzer}. Although effective
for language-conditioned RF analysis, their token layout and capacity
are inherited from the rendered image grid rather than the length and
information distribution of numerical I/Q observations.

\section{BATok: Budget-Adaptive Signal Tokenization}
\label{sec:signal_tokenizer}

\begin{figure*}[t]
\centering
\includegraphics[width=\linewidth]{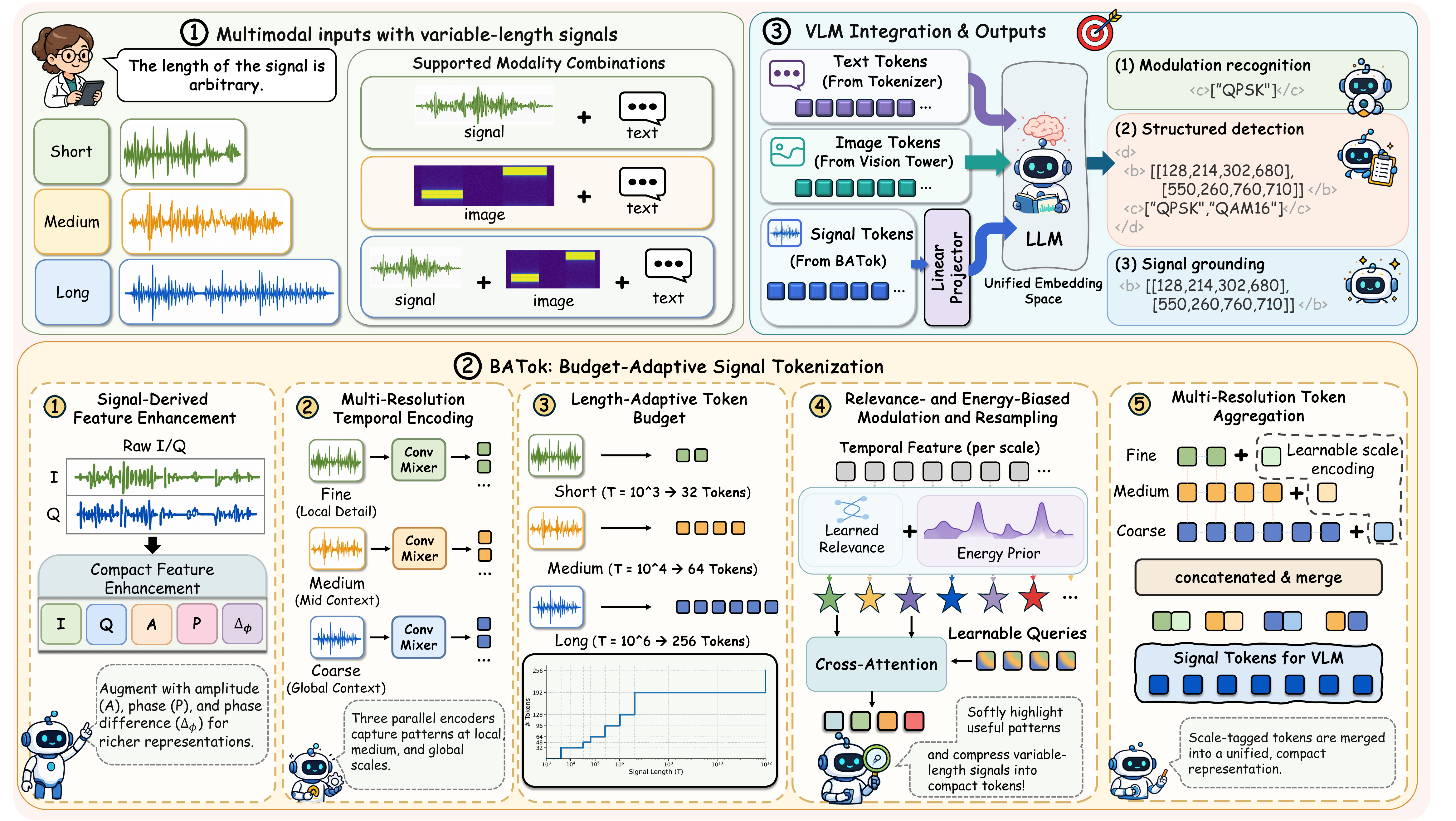}
\caption{Overview of BATok and its integration with a VLM.
Variable-length I/Q signals are augmented with signal-derived
features, encoded at multiple temporal resolutions, assigned a bounded
length-adaptive token budget, and resampled using learned relevance
and an energy prior. The resulting multi-resolution signal tokens are
aggregated, projected into the language embedding space, and jointly
processed with image and text tokens for unified spectrum
understanding.}
\label{fig:model}
\end{figure*}

Integrating raw I/Q signals into a VLM requires more than extracting waveform features. The representation must allocate sufficient capacity to observations of different lengths, preserve signal information that is distributed non-uniformly over time, and remain compatible with the bounded context of the language model. We address this problem at two levels: a length-dependent token budget controls the overall representation capacity, while signal-aware resampling determines how that capacity is distributed over the temporal observation. Figure~\ref{fig:model} summarizes the complete BATok pipeline and its integration with image and text tokens.

We propose \textbf{BATok}, which adapts the total token capacity to the input length and adapts temporal aggregation to the content for VLM conditioning.

Given a raw I/Q sequence $X=\{x_t\}_{t=1}^{T}$ with $x_t=I_t+jQ_t$, BATok learns the mapping
\begin{equation}
f_{\theta}:
\mathbb{R}^{T\times2}
\rightarrow
\mathbb{R}^{K(T)\times d_s},
\qquad
K(T)\leq K_{\max},
\end{equation}
where $T$ is the number of valid input samples, $K(T)$ is a length-dependent token budget, $K_{\max}$ is the maximum signal-token budget allowed by the VLM context, and $d_s$ is the signal-token dimension. The budget function $K(T)$ controls the global representation capacity, while signal-aware resampling determines how that capacity is distributed over the observation.

BATok first augments the raw I/Q sequence with signal-derived features and constructs candidate temporal representations at complementary resolutions. It then combines learned feature relevance with a local energy prior to modulate the branch representations and uses learnable queries to resample them into a compact set of scale-specific tokens. The resulting tokens are tagged by their temporal scale and assembled into the final signal representation. 

\subsection{Signal-Derived Feature Enhancement}
\label{subsec:feature_enhancement}
Raw I/Q samples provide the Cartesian components of a complex baseband signal, with amplitude and phase implicitly determined by their joint variation. Although a
sufficiently expressive network can in principle recover these
quantities from $I_t$ and $Q_t$, requiring a lightweight tokenizer to
learn the corresponding nonlinear transformations may unnecessarily
consume model capacity. We therefore retain the I/Q components
and augment them with signal-derived features that expose complementary
views of the underlying waveform.

For each sample, we construct
\begin{equation}
u_t=
\left[I_t, Q_t, A_t, P_t, \Delta\phi_t \right],
\end{equation}
where $A_t=\sqrt{I_t^2+Q_t^2}$, $P_t=I_t^2+Q_t^2$, and $\Delta\phi_t=
\angle\left(x_tx_{t-1}^{*}\right)$ denote the instantaneous amplitude, power, and inter-sample phase difference, providing complementary descriptions of the signal. These derived features provide complementary descriptions of envelope variation, relative signal activity, and local phase evolution. Although $A_t$ and $P_t$ are deterministically related, their different nonlinear scales expose waveform variations from complementary perspectives. 

The enhancement does not replace or discretize the original waveform;
instead, it preserves $I_t$ and $Q_t$ while providing explicit
signal-domain coordinates for envelope, energy, and phase evolution.
The resulting sequence $U=\{u_t\}_{t=1}^{T}$ is projected into a shared
hidden space by a lightweight convolutional stem:
\begin{equation}
H_0=
\operatorname{GELU}
\left(
\operatorname{GN}
\left(
\operatorname{Conv1D}(U)
\right)
\right),
\end{equation}
where $\operatorname{GN}$ denotes Group Normalization~\cite{wu2018groupnorm} and
$H_0\in\mathbb{R}^{T\times d_h}$ is passed to the subsequent
multi-resolution temporal encoder. In the VLM experiments,
$d_h=d_s=256$.

\subsection{Multi-Resolution Temporal Encoding}
\label{subsec:multi_res_encoding}

Electromagnetic observations may contain informative structures over substantially different temporal extents. Fine temporal resolution helps preserve localized waveform variations, whereas coarser representations provide broader temporal coverage and stronger sequence compression. A single receptive field and downsampling ratio may offer an inadequate trade-off between local detail and contextual coverage. BATok instead constructs complementary temporal representations at multiple resolutions before adaptive resampling.

Let $\mathcal{B}$ denote the set of temporal branches operating at different resolutions. Each branch applies a strided depthwise-separable one-dimensional convolution:
\begin{equation}
H_b=
\operatorname{DSConv}_b(H_0)\in\mathbb{R}^{T_b\times d_h},
\qquad
b\in\mathcal{B},
\end{equation}
where $T_b$ is the effective length after downsampling of branch $b$. The receptive field and downsampling ratio are set to increase progressively across the branches, producing different trade-offs among temporal detail, contextual coverage, and representation length. The fine-resolution branch retains localized waveform variations, the intermediate-resolution branch captures structures over moderate temporal extents, and the coarse-resolution branch summarizes broader contexts under stronger compression.

After downsampling, each branch is refined by the lightweight temporal and channel mixing without changing its sequence length:
\begin{equation}
\begin{aligned}
\tilde{H}_b &=
H_b +
\operatorname{DWConv}_b
\left(
\operatorname{LN}(H_b)
\right), \\
H_b' &=
\tilde{H}_b +
\operatorname{GLU}
\left(
\operatorname{PWConv}_b
\left(
\operatorname{LN}(\tilde{H}_b)
\right)
\right).
\end{aligned}
\end{equation}
Here, $\operatorname{LN}$ denotes Layer Normalization. $\operatorname{DWConv}$ and $\operatorname{PWConv}$ follow the depthwise and pointwise operations in depthwise separable convolution~\cite{howard2017mobilenets}, and $\operatorname{GLU}$ denotes the gated linear unit~\cite{dauphin2017glu}. This factorized design enriches each branch representation with linear complexity in the sequence length, avoiding the quadratic cost of global self-attention~\cite{vaswani2017attention}. The feature sequences
$\{H_b'\}_{b\in\mathcal{B}}$
form a multi-resolution representation pool for the subsequent adaptive resampling stage, rather than being directly forwarded to the VLM.

\subsection{Adaptive Token Budgeting and Resampling}
\label{subsec:adaptive_resampling}

The multi-resolution encoder produces branch features whose temporal lengths still vary with the input length. BATok controls the overall representation capacity through a length-dependent but strictly bounded token budget and subsequently allocates this capacity over time through signal-aware query resampling.

\paragraph{Length-adaptive token budget.}
A fixed token budget imposes increasingly severe compression as the signal length grows, which may preserve redundant capacity for short inputs while discarding more temporal detail from long inputs. Conversely, assigning tokens in direct proportion to the input length causes the representation cost to grow continuously with $T$ and may exceed the bounded context available to the VLM. Therefore, the total number of signal tokens is determined by a monotonic piecewise-constant function:
\begin{equation}
K(T) = \operatorname{clip} \left( K_{\min} + \sum_{j=1}^{J} \Delta_j
\mathbb{I}(T>\tau_j),  K_{\min}, K_{\max} \right),
\label{eq:token_budget}
\end{equation}
where $\tau_j$ are increasing length thresholds and
$\Delta_j\geq 0$ are budget increments. $\mathbb{I}(\cdot)$ denotes the indicator function. Consequently, longer signals are assigned more tokens, while the resulting token sequence remains compatible with the bounded context of the VLM. The piecewise-constant schedule also assigns inputs within the same threshold interval to use the same token budget, avoiding unnecessary changes in token count for small variations.

The total budget is distributed across the resolution branches according
to predefined proportions ${\rho_b}$, where
$\rho_b\geq 0$ and
$\sum_{b\in\mathcal B}\rho_b=1$. Integer branch budgets $K_b = \lfloor \rho_b K(T) \rfloor$ are obtained
from these proportions and adjusted to preserve the total token count $\sum_{b\in\mathcal B}K_b=K(T)$.

\paragraph{Relevance- and Energy-Biased Modulation.}
Although all valid temporal positions are retained, they may contribute differently to the compact representation. A learned relevance score captures task-dependent information from the latent features, while local energy provides a complementary indication of temporal signal activity. However, high-energy regions may also arise from interference, impulsive noise, or unrelated transmissions. BATok therefore uses energy only as a soft bias rather than as a direct criterion for selecting temporal positions. 

For the $t$-th feature $h'_{b,t}$ of branch $b$, BATok combines a learned relevance score with a local energy prior:
\begin{equation}
\begin{aligned}
    E_{b,t}& =
\operatorname{AvgPool}_{k_b,s_b}(I^2+Q^2)_t\\
r_{b,t} &= \operatorname{MLP}(h'_{b,t}) + \lambda_b e_{b,t},
\end{aligned}
\label{eq:importance_score}
\end{equation}
where $k_b$ and $s_b$ denote the pooling kernel size and stride aligned with the temporal resolution of branch $b$. $e_{b,t}$ is the normalization of $E_{b,t}$. The term $\operatorname{MLP}(h'_{b,t})$ denotes the learned relevance score, while $\lambda_b$ is a learnable coefficient controlling the contribution of the normalized energy prior.

The combined scores are normalized over valid temporal positions and used to modulate the branch features:
\begin{equation}
\bar H_b
=
H_b'
\odot
\left[
1+
\operatorname{clip}
\left(
\sqrt{T_b} \operatorname{softmax}(r_b),
0,c
\right)
\right],
\label{eq:soft_enhancement}
\end{equation}
where $c$ bounds the maximum enhancement magnitude, and $\odot$ denotes element-wise multiplication with broadcasting over the channel dimension. The factor $\sqrt{T_b}$ compensates for the reduced magnitude of normalized scores in longer sequences, while clipping prevents a small number of positions from being amplified. This modulation does not remove low-scoring positions; instead, it introduces a soft bias before query-based resampling, allowing the subsequent cross-attention module to consider the complete valid sequence.

\paragraph{Query-based resampling.}
Learnable-query interfaces have been used to compress variable-size perceptual features before language-model conditioning~\cite{alayrac2022flamingo,li2023blip2}. BATok adapts this interface to raw I/Q by assigning a length-dependent number of queries to each temporal scale. Each modulated branch is converted into a bounded set of tokens through cross-attention with learnable queries and then refined by a lightweight residual feed-forward layer:
\begin{equation}
\begin{aligned}
Z_b
&=
\operatorname{CrossAttn}
\left(
Q_b^{(K_b)},
\bar H_b,
\bar H_b
\right),
\\
\widehat Z_b
&=
Z_b+
\operatorname{FFN}
\left(
\operatorname{LN}(Z_b)
\right),
\end{aligned}
\label{eq:query_resampling}
\end{equation}
where
$Q_b^{(K_b)}=\{q_{b,i}\}_{i=1}^{K_b} \in \mathbb{R}^{K_b \times d_h}$
contains the first $K_b$ queries from the learnable query pool of branch $b$. 
This process compresses each variable-length branch representation into $K_b$ scale tokens, keeping the signal-token length adaptive yet bounded while emphasizing active signal regions and locally informative temporal structures.

\subsection{Multi-Resolution Token Aggregation}
\label{subsec:scale_fusion}

The scale-tagged tokens are concatenated and refined by a shared token-wise residual projection. Explicit cross-scale interactions are subsequently modeled by the VLM. Since these tokens are extracted at different temporal resolutions, they contain distinct scale-specific biases. We therefore add a learnable scale encoding $\eta_b \in \mathbb{R}^{d_s}$ to each branch:
\begin{equation}
\tilde{Z}_b = \hat{Z}_b + \eta_b,
\quad b \in \mathcal{B}.
\end{equation}

The scale-aware tokens are concatenated and passed through a shared token-wise residual projection:
\begin{equation}
Z =
\operatorname{Concat}_{b\in\mathcal{B}}(\tilde{Z}_b),
\end{equation}
\begin{equation}
Z_{\mathrm{sig}}
=
Z
+
\operatorname{GELU}
\left(
\operatorname{Linear}
\left(
\operatorname{LN}(Z)
\right)
\right).
\end{equation}
where $Z \in \mathbb{R}^{K(T) \times d_s}$. The final representation $Z_{\mathrm{sig}} \in \mathbb{R}^{K(T) \times d_s}$ serves as compact multi-resolution signal tokens, which are further projected into the VLM language embedding space.

\section{VLM Integration and Training}
\label{sec:vlm_training}

Following common practice in multimodal large language models~\cite{alayrac2022flamingo,li2023blip2,liu2023llava15}, the signal tokens are not discretized into textual symbols, but are mapped into the VLM language embedding space as continuous multimodal tokens. After obtaining compact signal tokens $Z_{\mathrm{sig}} \in \mathbb{R}^{K(T) \times d_s}$ from BATok, we project them into the VLM embedding space using a two-layer projector:
\begin{equation}
H_s =
W_2\operatorname{GELU}
\left(
W_1\operatorname{LN}(Z_{\mathrm{sig}})
\right)
\in \mathbb{R}^{K(T) \times d_{\ell}},
\end{equation}
where $W_1$ maps the signal-token dimension $d_s$ to the VLM language embedding dimension $d_{\ell}$, and $W_2$ preserves $d_{\ell}$. The projected embeddings $H_s$ are fed into the VLM together with available image embeddings and textual instruction embeddings. Image embeddings are obtained from the original visual encoder of the VLM, while text embeddings are produced by its token embedding layer. Therefore, the model can support three input configurations: image-text, signal-text, and image-signal-text inputs.

We train the integrated model with standard supervised fine-tuning. Given $\mathcal{D}
=
\{(X^{(n)}, V^{(n)}, P^{(n)}, Y^{(n)})\}_{n=1}^{N}
$, where $X^{(n)}$ and $V^{(n)}$ denote optional signal and waterfall-image inputs, $P^{(n)}$ is the textual instruction, and $Y^{(n)}$ is the target response, we optimize the response-only autoregressive objective:
\begin{equation}
\mathcal{L}_{\mathrm{sft}}
=
-
\sum_{n=1}^{N}
\sum_{i=1}^{|Y^{(n)}|}
\log
p_{\phi}
\left(
y_i^{(n)}
\mid
X^{(n)}, V^{(n)}, P^{(n)}, y_{<i}^{(n)}
\right).
\end{equation}
The loss is computed only on response tokens, while image, signal, and instruction tokens serve as conditioning context.

\section{EMSpec-Instruct Dataset}

Existing electromagnetic datasets are primarily designed for task-specific recognition or detection, typically providing class labels for individual I/Q segments or bounding-box annotations on time-frequency representations~\cite{tekbiyik2020hisarmod,boegner2022torchsig,boegner2022wideband}. Such supervision does not directly support instruction-conditioned recognition, structured multi-signal detection, and language-conditioned signal grounding. We therefore construct \textbf{EMSpec-Instruct}, a multimodal instruction dataset that associates raw I/Q observations, waterfall images, structured signal metadata, and natural-language queries.

The dataset integrates both single-signal and multi-signal
observations. The single-signal stream combines
\textsc{HisarMod-Sub}, a subset of HisarMod2019.1 (hereafter
HisarMod2019)~\cite{tekbiyik2020hisarmod}, and
\textsc{EMSpec-Sim-Sub}, a subset of our single-signal simulation
benchmark \textsc{EMSpec-Sim}, for modulation recognition. The
	multi-signal stream combines \texttt{SpaceNet}~\cite{spacenet_dataset_chaspark} and a multi-signal
simulated dataset constructed with RF simulation utilities~\cite{boegner2022torchsig}, supporting structured detection and signal
grounding. The simulated data complement the real datasets by
expanding modulation coverage and introducing diverse acquisition
lengths, signal densities, and time-frequency overlap patterns.

\begin{figure}[!htb]
	\centering
	\includegraphics[width=\linewidth]{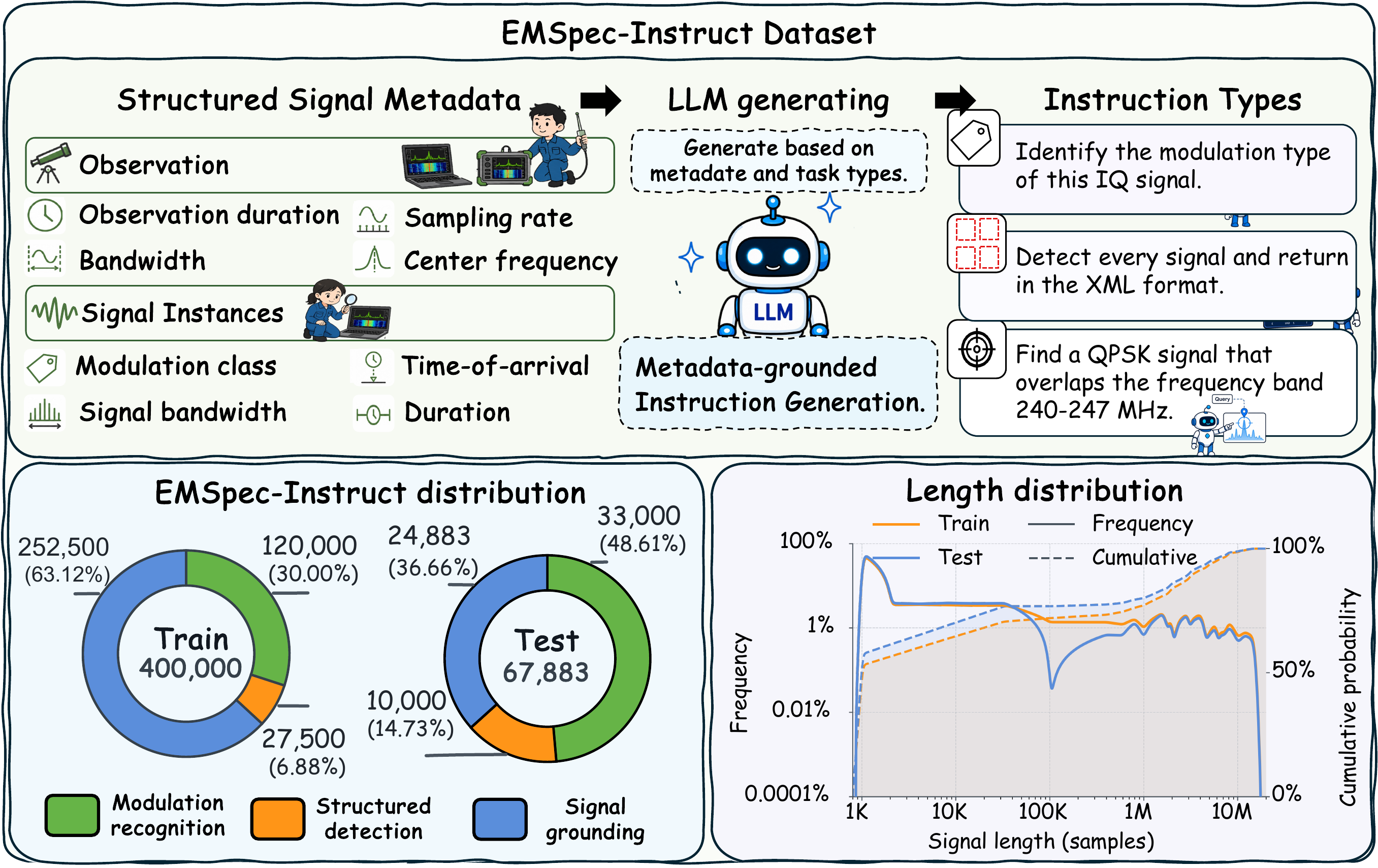}
	\caption{Construction and statistics of EMSpec-Instruct. Structured observation- and instance-level metadata are converted into metadata-grounded instructions. The bottom panels summarize the task composition of the training and test sets and the distribution of signal lengths.}
	\label{fig:fig3}
\end{figure}

To associate electromagnetic observations with language supervision, we first derive task conditions from the structured metadata and then use Qwen3.5-Max~\cite{qwen2026qwen35} to generate linguistically diverse instruction records. For modulation recognition and structured detection, Qwen3.5-Max generates the question while the compact answer is filled directly from the class label or complete scene annotation. For language-conditioned signal grounding, Qwen3.5-Max generates both the metadata-conditioned question and answer; the result is retained only when its returned boxes agree exactly with the programmatically computed match set. EMSpec-Instruct contains three task families. \textbf{Modulation recognition} predicts the modulation category of a single signal. \textbf{Structured detection} identifies all signal instances in an observation and returns their bounding boxes and class labels in a predefined format. \textbf{Language-conditioned signal grounding} retrieves signals satisfying natural-language constraints, including modulation type, time interval, frequency interval, their combinations, and time-frequency overlap. Positive queries return the boxes of all matched signals, while unmatched conditions produce an empty result. Figure~\ref{fig:fig3} summarizes this construction pipeline together with the resulting task and signal-length distributions.

The final dataset contains 400,000 training instructions and
67,883 test instructions, as in Figure~\ref{fig:fig3}. We construct aligned multimodal and waterfall-only configurations for controlled comparison between image-only VLMs and models additionally equipped with the BATok. All partitions are defined at the underlying observation level before instruction generation, ensuring that the same I/Q sequence, its derived waterfall image, and associated paraphrased instructions do not cross the training and test splits.
Further construction details, source statistics, task distributions,
and validation results are provided in Supplementary
Sections~S2 and~S3.3.

\begin{table*}[!htb]
	\centering
	\setlength{\tabcolsep}{2.2pt}
	\begin{tabular*}{\textwidth}{@{\extracolsep{\fill}}l c c c c c c c c c c c@{}}
		\toprule
		& &
		\multicolumn{3}{c}{Modulation Recognition} &
		\multicolumn{3}{c}{Structured Detection} &
		\multicolumn{4}{c}{Signal Grounding} \\
		\cmidrule(lr){3-5}\cmidrule(lr){6-8}\cmidrule(lr){9-12}
		Backbone & BATok &
		\shortstack{OA\\(\%) $\uparrow$} &
		\shortstack{Macro-F1\\(\%) $\uparrow$} &
		\shortstack{Parse\\(\%) $\uparrow$} &
		\shortstack{F1@0.5\\(\%) $\uparrow$} &
		\shortstack{mIoU\\(\%) $\uparrow$} &
		\shortstack{Parse\\(\%) $\uparrow$} &
		\shortstack{F1@0.5\\(\%) $\uparrow$} &
		\shortstack{mIoU\\(\%) $\uparrow$} &
		\shortstack{Parse\\(\%) $\uparrow$} &
		\shortstack{Empty Acc.\\(\%) $\uparrow$} \\
		\midrule
		Qwen3-VL-2B & 
		& 33.90 & 35.38 & \textbf{100.00}
		& 62.25 & 94.66 & 98.08
		& 82.52 & 94.75 & \textbf{99.99} & \textbf{93.19} \\
		Qwen3-VL-2B & \checkmark
		& \textbf{40.91} & \textbf{41.26} & 86.02
		& \textbf{64.27} & \textbf{95.20} & \textbf{98.29}
		& \textbf{83.00} & \textbf{95.24} & 99.96 & 92.43 \\
		\midrule
		LLaVA-1.5-7B &
		& 25.58 & 26.26 & \textbf{100.00}
		& 45.99 & 88.72 & \textbf{99.43}
		& 66.94 & 88.81 & \textbf{99.95} & \textbf{91.92} \\
		LLaVA-1.5-7B & \checkmark
		& \textbf{44.14} & \textbf{43.20} & 94.95
		& \textbf{47.80} & \textbf{89.40} & 97.17
		& \textbf{68.27} & \textbf{89.39} & 99.87 & 91.23 \\
		\bottomrule
	\end{tabular*}
	\caption{Unified spectrum-understanding results on EMSpec-Instruct. The original VLMs receive image--text inputs, whereas their BATok-enhanced counterparts receive aligned image--signal--text inputs. Parse is the proportion of outputs satisfying the required task-specific format, and Empty Acc. is the accuracy of returning an empty set for unmatched grounding queries. Within each backbone, the better value between the original VLM and its BATok-enhanced counterpart is bolded.}
	\label{tab:unified_results}
\end{table*}

\section{Experiments}

\subsection{Experimental Setup}

We evaluate BATok in unified VLM and standalone
modulation-recognition settings. For the former, BATok is integrated
into Qwen3-VL-2B~\cite{bai2025qwen3vl} and
LLaVA-1.5-7B~\cite{liu2023llava15}. The original VLMs are fine-tuned
on image--text instructions, whereas their BATok-enhanced counterparts
use signal--text, image--text, and image--signal--text instructions.
The main evaluation supplies image--text inputs to the original VLMs
and aligned image--signal--text inputs to the enhanced models, using
the same observations and train--test partitions. All VLM variants
are fine-tuned for five epochs on eight NVIDIA A800 GPUs.

Standalone representation evaluation and ablations use the complete
HisarMod2019~\cite{tekbiyik2020hisarmod} and \textsc{EMSpec-Sim}
benchmarks; their aligned subsets within EMSpec-Instruct are denoted
\textsc{HisarMod-Sub} and \textsc{EMSpec-Sim-Sub}. BATok is
followed by a lightweight classification head. All standalone
classifiers and task-specific detectors are trained for 100 epochs on
one NVIDIA GeForce RTX 4090 under common dataset-specific partitions
and preprocessing protocols. Further configurations and expanded
results are provided in Supplementary Sections~S1 and~S3.

\subsection{Unified Instruction-Conditioned Spectrum Understanding}

Table~\ref{tab:unified_results} compares each image-based VLM with its
BATok-enhanced counterpart to test whether raw I/Q inputs can be
incorporated without changing the unified task interface. This comparison
reflects the combined effect of BATok integration and mixed-modality
training; Supplementary Section~S3.2 compares input modalities within
the same trained checkpoint. The enhanced models yield
the largest gains in modulation recognition: OA increases by 7.01 and
18.56 points for Qwen3-VL-2B and LLaVA-1.5-7B, respectively, with
Macro-F1 gains of 5.88 and 16.94 points. Positive-query detection and
grounding metrics also improve within both backbones, whereas
empty-set accuracy decreases slightly by 0.76 and 0.69 points,
revealing a small trade-off in rejecting unmatched queries.
Localization outputs remain highly parseable, while the lower
modulation-recognition parse rate of Qwen3-VL-2B+BATok confirms that
format compliance and recognition accuracy are distinct. The
consistent recognition gains across both backbones support BATok as a
VLM-agnostic interface, while the smaller localization gains indicate
that raw I/Q complements the time--frequency geometry already visible
in waterfall images. Supplementary Section~S3.4 attributes the
Qwen3-VL-2B gain mainly to semantics and the LLaVA-1.5-7B gain to joint
geometric--semantic improvements. Section~S3.2 also shows that signal--text
favors recognition, image--text preserves localization, and joint
input provides the best balance across the evaluated tasks.

\subsection{Dataset-Level Task Feasibility}

Task-specific models validate the learnability of the recognition and
detection subsets assembled from existing and simulated observations.
On \textsc{HisarMod-Sub}$\cup$\textsc{EMSpec-Sim-Sub}, BATok reaches
49.36\% Accuracy and 47.19\% Macro-F1; on structured detection,
YOLO26x reaches 0.9517 mIoU and 0.6925 F1@0.5. These results show that
the constructed observations and annotations support their intended
tasks. Additional results are provided in Supplementary Section~S3.3.

\subsection{Qualitative Analysis of Detection and Grounding}

\begin{figure*}[!t]
\centering
\setlength{\tabcolsep}{1.5pt}
\begin{tabular}{@{}cccccc@{}}
 \small Detection GT &
 \small Qwen3-VL-2B &
 \small Qwen3-VL-2B+BATok &
 \small Grounding GT &
 \small Qwen3-VL-2B &
 \small Qwen3-VL-2B+BATok \\
\includegraphics[width=0.158\textwidth]{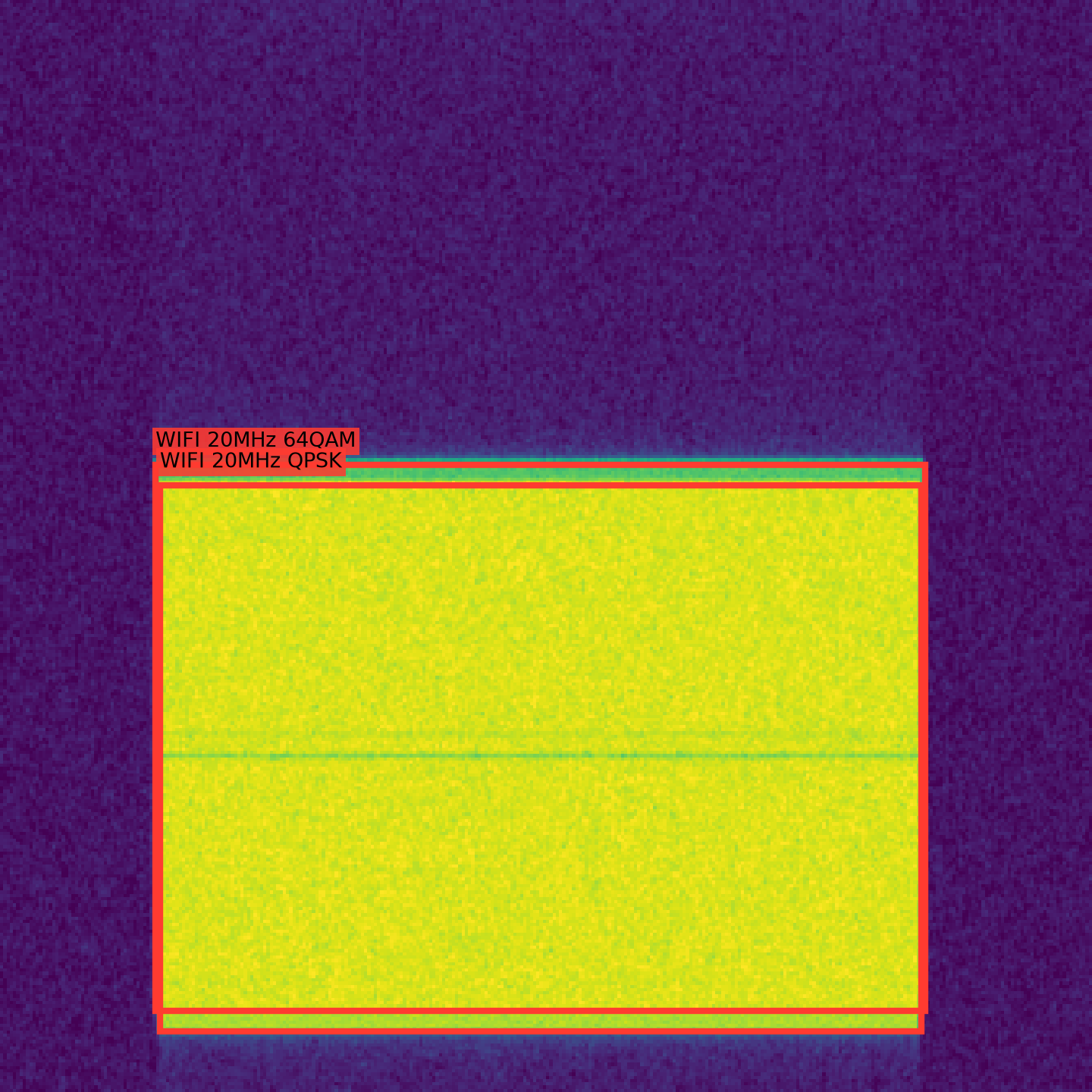} &
\includegraphics[width=0.158\textwidth]{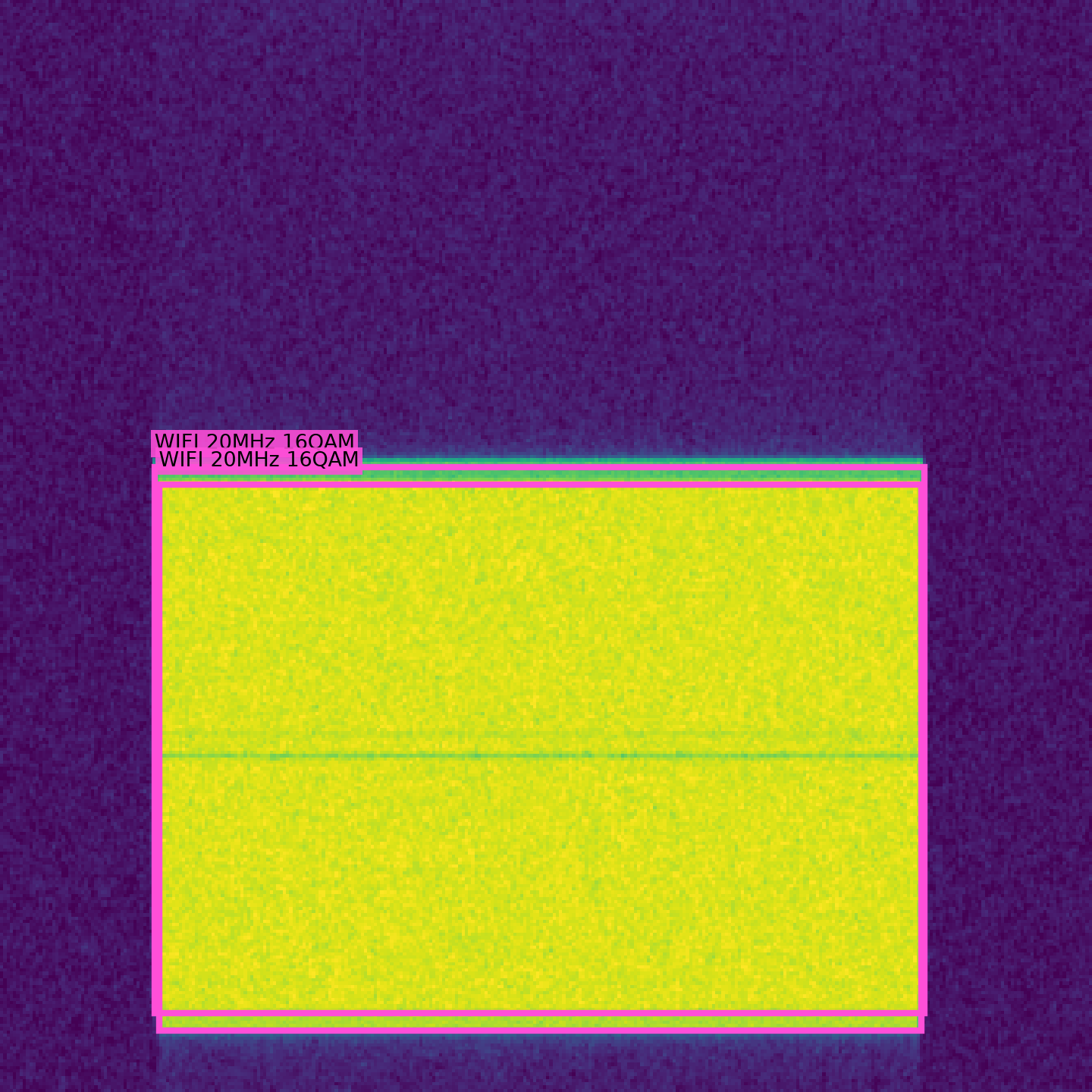} &
\includegraphics[width=0.158\textwidth]{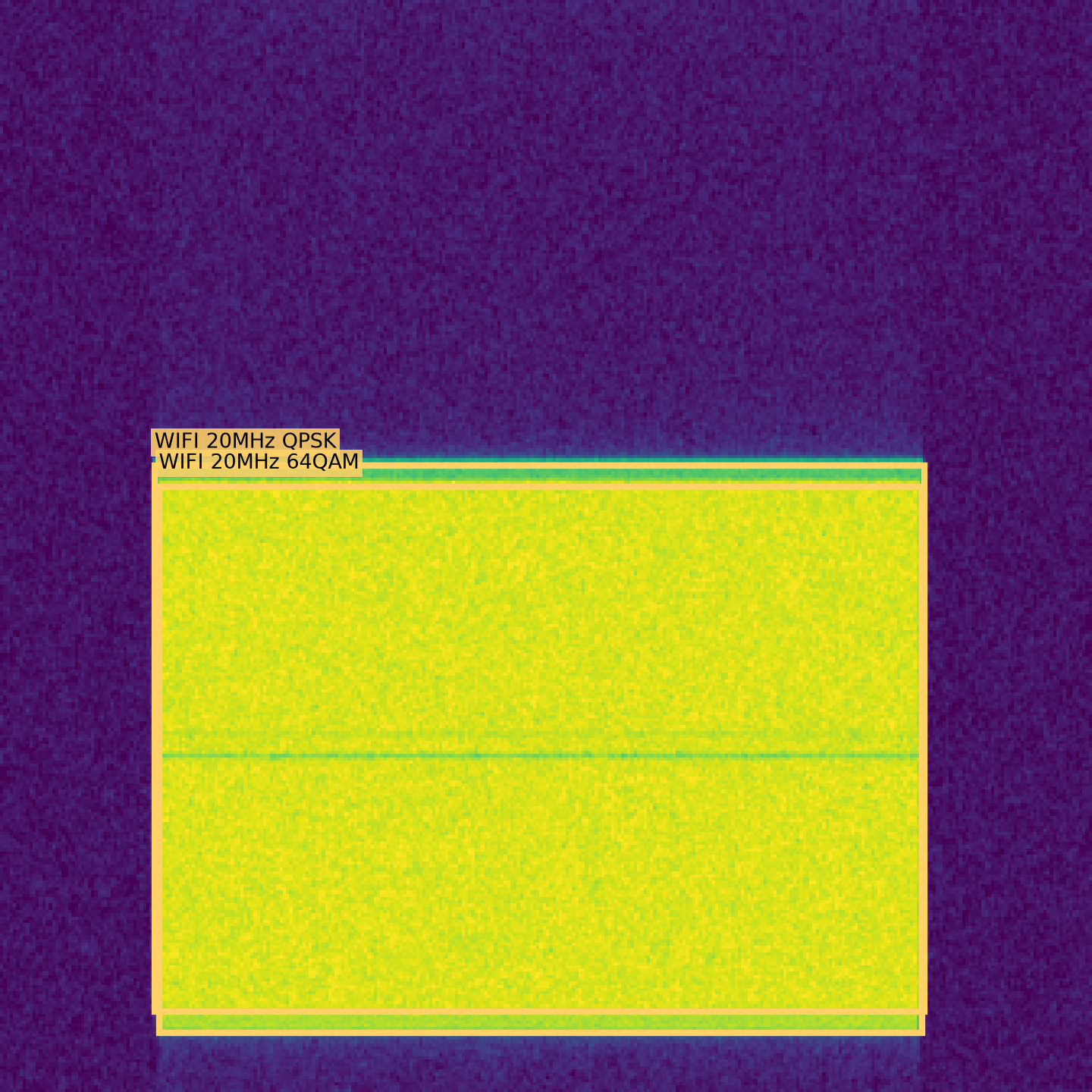} &
\includegraphics[width=0.158\textwidth]{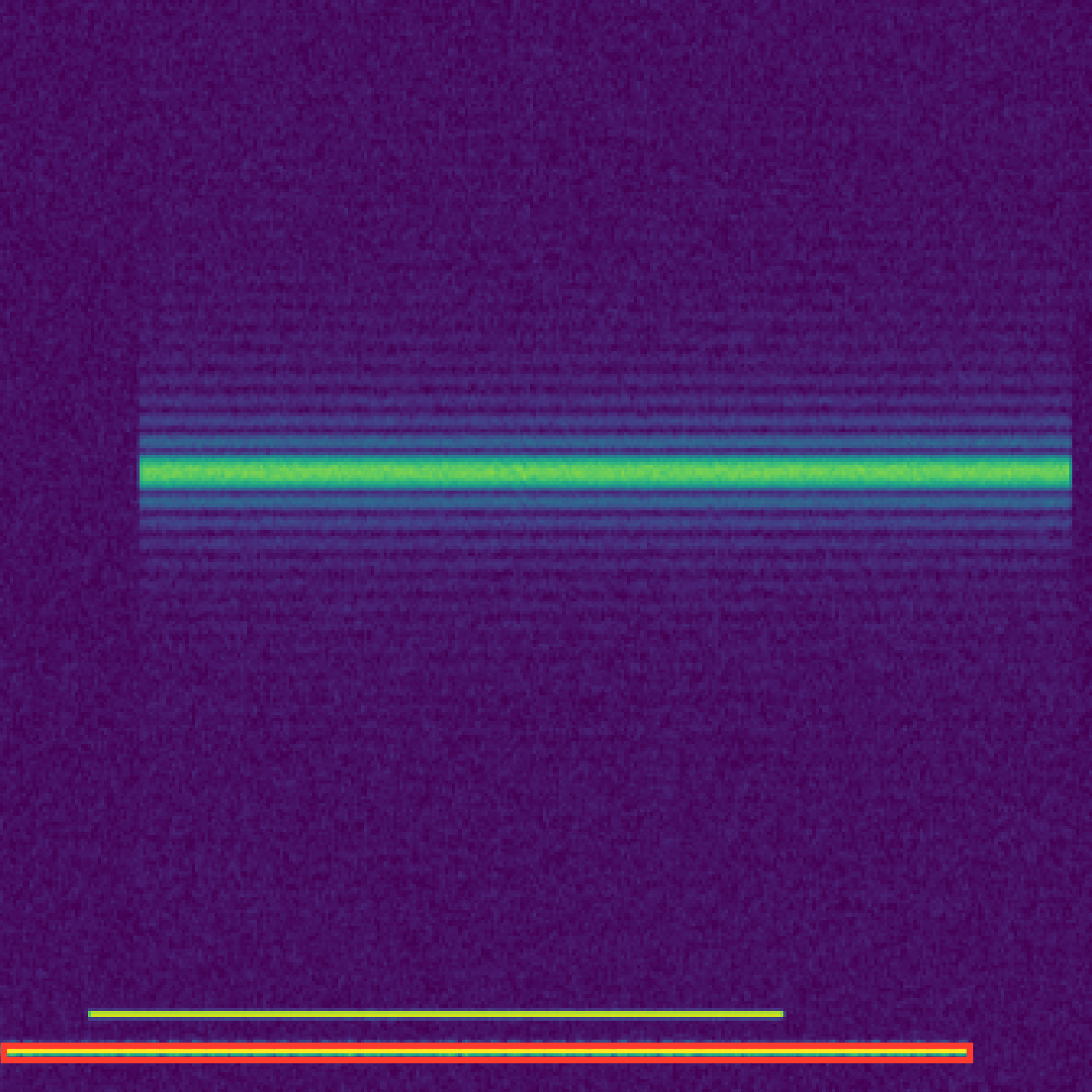} &
\includegraphics[width=0.158\textwidth]{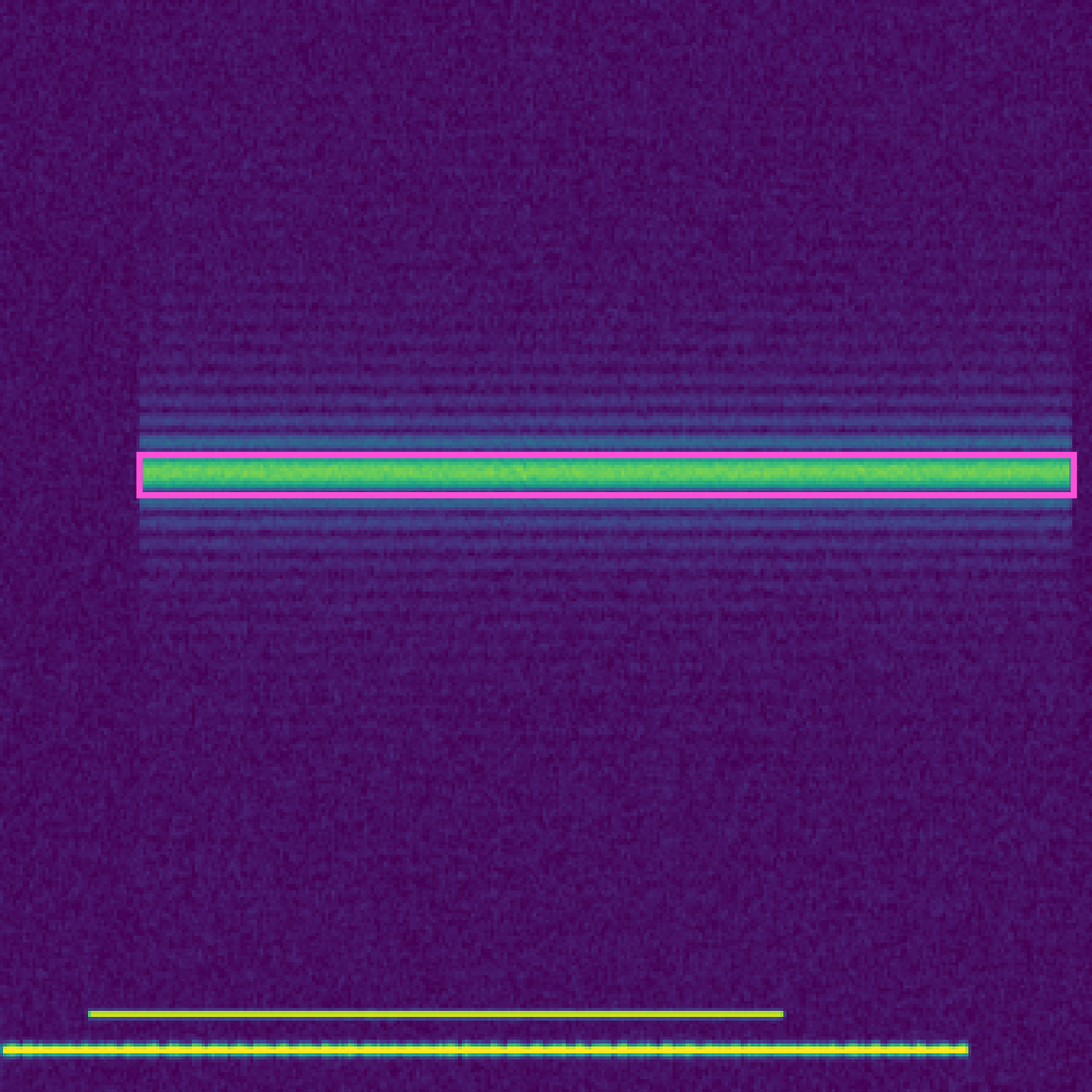} &
\includegraphics[width=0.158\textwidth]{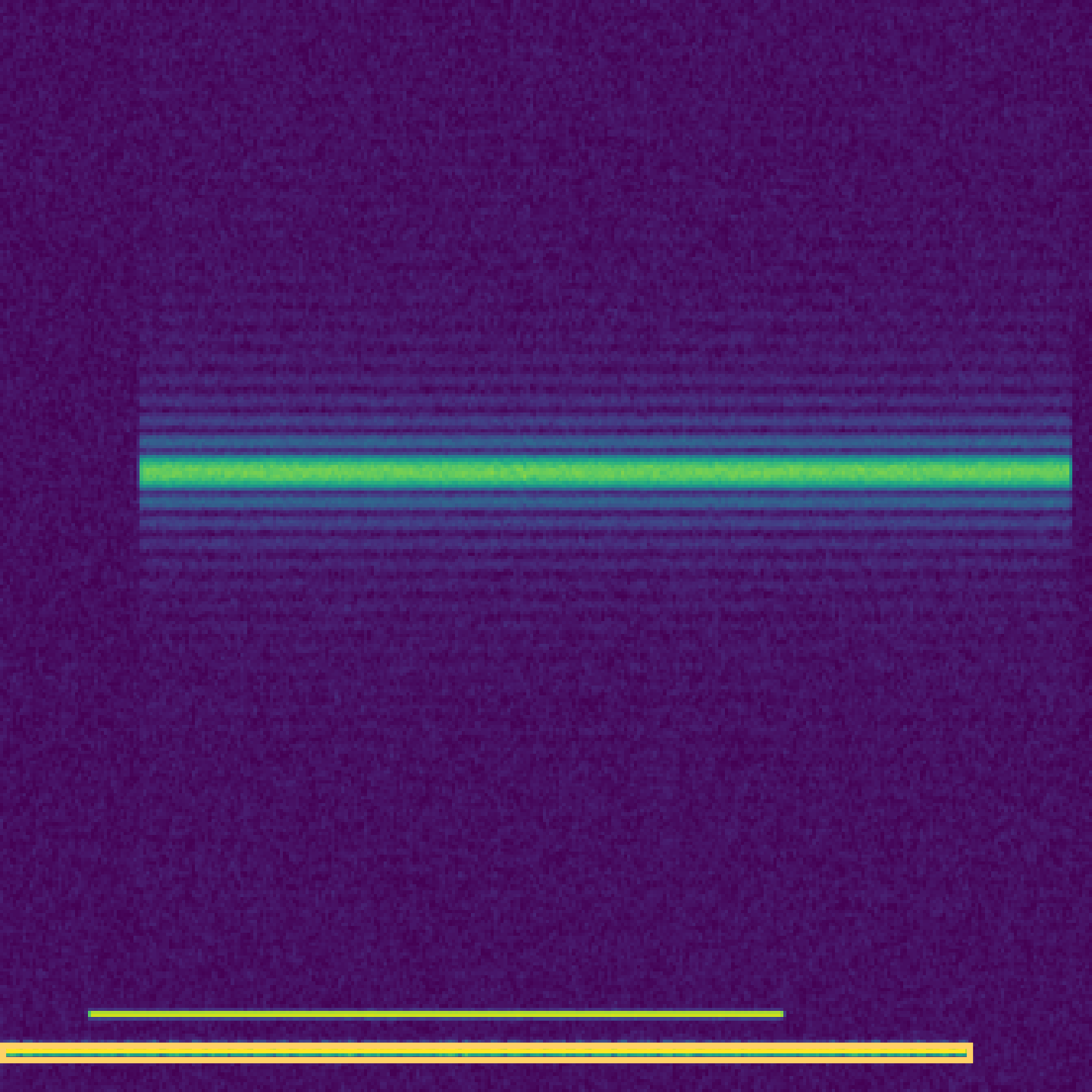}
\end{tabular}
\caption{Visualization of detection and grounding results. \textbf{Detection question:} ``Enumerate every transmission using aligned box and class arrays.'' \textbf{Ground truth:} $[141,424,846,924]$ (WiFi 20\,MHz 64QAM) and $[145,443,842,943]$ (WiFi 20\,MHz QPSK). \textbf{Grounding question:} ``Return boxes of the longest-duration transmissions.'' \textbf{Ground truth:} $[2,956,887,969]$. }
\label{fig:main-qualitative}
\end{figure*}

Figure~\ref{fig:main-qualitative} contrasts the image-only
Qwen3-VL-2B baseline with its BATok-enhanced counterpart on
representative simulated examples. In the detection example, both
models accurately localize the two overlapping transmissions, but the
image-only baseline assigns both instances to 16QAM. With access to
raw I/Q observations, the BATok-enhanced model recognizes the two
ground-truth modulation classes. In the relational grounding example,
the image-only baseline selects a different high-energy band, whereas
the BATok-enhanced model identifies the longest-duration
transmission. These examples illustrate the behaviors reflected by
the aggregate metrics rather than serving as substitutes for
quantitative evaluation. Additional cases covering both VLM
backbones and diverse query types are provided in Supplementary
Section~S5. Taken together, Figure~\ref{fig:main-qualitative} shows that
the raw I/Q branch provides complementary evidence for resolving
modulation ambiguity and selecting signals under relational
constraints, while the image stream preserves the time--frequency
structure required for localization.

\subsection{Standalone Evaluation of Signal Representations}

We next use independently trained signal classifiers to evaluate raw
I/Q representation quality without the mixed-modality training
procedure of the VLM experiments. This setting directly tests whether
BATok learns effective signal representations and how it compares
with task-specific signal encoders on complete classification
benchmarks.
Table~\ref{tab:standalone_results} compares raw-I/Q signal encoders
across two independently trained modulation-recognition benchmarks.
The references include MAMC~\cite{10705364},
the Signal Transformer (SiT)~\cite{zhai2025cdsit}, and
DNCNet with its CNN-based signal backbone~\cite{du2022dncnet}.
These task-specific models are not expected to provide the
instruction-following and grounding capabilities of the unified VLM;
instead, they provide controlled reference points for representation
quality. 

\begin{table}[!htb]
\centering
 \small
\setlength{\tabcolsep}{3pt}
\begin{tabular}{@{}c c c@{}}
\toprule
Method & Acc. (\%) $\uparrow$ &  Macro-F1 (\%) $\uparrow$ \\
\midrule
MAMC~\cite{10705364} & 43.20 & 42.31 \\
SiT~\cite{zhai2025cdsit} & 53.56 & 53.36 \\
DNCNet~\cite{du2022dncnet} & \underline{58.18} & \underline{56.89} \\
BATok & \textbf{58.64} & \textbf{58.54} \\
\bottomrule
\end{tabular}
\caption{Standalone raw-I/Q modulation-recognition results. Each average is the unweighted mean across the two primary benchmarks, HisarMod2019 and  \textsc{EMSpec-Sim}. Best and second-best averages are bolded and underlined, respectively. More results are provided in Supplementary Section~S4.1.}
\label{tab:standalone_results}
\end{table}

BATok obtains the highest mean Accuracy (58.64\%) and Macro-F1
(58.54\%) across the two benchmarks. It performs best on
HisarMod2019 and achieves the highest Macro-F1 on
\textsc{EMSpec-Sim}, while DNCNet retains the highest Accuracy
on \textsc{EMSpec-Sim}. Its average gains over the second-best method
are 0.46 Accuracy and 1.65 Macro-F1 points. Thus, BATok is not
uniformly superior on every dataset, but its bounded representation
remains competitive across distinct signal regimes. Per-benchmark and input-configuration results
are provided in Supplementary Sections~S4.1 and~S3.2.

\subsection{Ablation Studies}

To isolate the contribution of individual tokenizers from
the language model and multimodal instruction-tuning procedure, we
perform all ablations in the standalone modulation-recognition
setting. Table~\ref{tab:ablation_results} evaluates four design
questions: how token capacity should depend on signal length, whether
signal-derived features improve raw I/Q representations, how many
temporal resolutions are needed, and whether the energy prior and
learned relevance score provide complementary resampling biases.
All variants follow the same evaluation protocol, and the reported
metrics are averaged over HisarMod2019 and the complete
\textsc{EMSpec-Sim} benchmark.

\begin{table}[t]
\centering
 \small
\setlength{\tabcolsep}{2pt}
\begin{tabular}{@{}l l c c@{}}
\toprule
Group & Variant & Acc. (\%) $\uparrow$ & Macro-F1 (\%) $\uparrow$ \\
\midrule
\multirow{4}{*}{\shortstack[c]{Token\\budget}} & Fixed $K=32$
& 57.68 & 57.53 \\
& Fixed $K=48$
& 57.97 & 57.60 \\
& Linear-high
& \textbf{58.70} & \underline{58.48} \\
& Piecewise $K(T)^\dagger$
& \underline{58.64} & \textbf{58.54} \\
\midrule
\multirow{2}{*}{\shortstack[c]{Input\\features}} & I/Q only
& \underline{56.37} & \underline{56.30} \\
& I/Q + $A$ + $P$ + $\Delta\phi$ $\dagger$
& \textbf{58.64} & \textbf{58.54} \\
\midrule
\multirow{3}{*}{\shortstack[c]{Temporal\\scales}} & One branch (small)
& 54.70 & 54.53 \\
& \shortstack[l]{Two branches\\(small+mid)}
& \underline{56.06} & \underline{56.08} \\
& Three branches $\dagger$
& \textbf{58.64} & \textbf{58.54} \\
\midrule
\multirow{3}{*}{\shortstack[c]{Resampling\\bias}} & Energy prior only
& \underline{56.97} & \underline{57.22} \\
& Learned relevance only
& 56.80 & 56.45 \\
& \shortstack[l]{Energy + learned\\relevance $\dagger$}
& \textbf{58.64} & \textbf{58.54} \\
\bottomrule
\end{tabular}
\caption{Ablation study of BATok in standalone modulation recognition.
Accuracy and Macro-F1 are percentages averaged over HisarMod2019 and
\textsc{EMSpec-Sim}; $\dagger$ denotes the full BATok configuration
within each ablation group.}
\label{tab:ablation_results}
\end{table}

Among the selected operating points in Table~\ref{tab:ablation_results},
the piecewise schedule $K(T)$ attains the strongest mean Macro-F1 (58.54\%)
and remains within 0.06 points of the best mean Accuracy obtained by
Linear-high. Signal-derived features improve Accuracy/Macro-F1 by
2.27/2.24 points over I/Q alone, and the third temporal branch adds
2.58/2.46 points over two branches. A similar trend appears in
resampling, where combining energy and learned relevance exceeds the
stronger single bias by 1.67/1.32 points.
These results show that length-aware capacity, multi-resolution
encoding, and complementary resampling cues jointly support the final
representation. Full budget sweeps and per-dataset results are
reported in Supplementary Sections~S4.1--S4.4.

\section{Conclusion}

In this paper, we introduced BATok, which maps mixed-length raw I/Q observations to
bounded, content-aware tokens for MLLMs. A length-dependent budget,
multi-resolution encoding, and energy--relevance-guided resampling
adapt capacity without exposing the language model to unbounded
sequences. Across two VLMs, BATok improves modulation recognition and
positive-query detection and grounding, showing that waveform
evidence complements waterfall images. Standalone results and
ablations support its representation and design. However, the limitation is that open-set and cross-domain generalization
remain unverified. Moreover, BATok's small parameter footprint and
bounded output coexist with higher standalone latency than compact
classifiers because of its multi-resolution resampling.

\bibliography{ArxivCombined}

\clearpage
\onecolumn
\begin{center}
{\Huge\bfseries Supplementary\par}
\end{center}
\vspace{1em}
\setcounter{secnumdepth}{2}
\renewcommand{\thesection}{S\arabic{section}}
\renewcommand{\thesubsection}{\thesection.\arabic{subsection}}
\renewcommand{\thesubsubsection}{\thesubsection.\arabic{subsubsection}}
\renewcommand{\thefigure}{S\arabic{figure}}
\renewcommand{\thetable}{S\arabic{table}}
\renewcommand{\theequation}{S\arabic{equation}}
\renewcommand{\thealgorithm}{S\arabic{algorithm}}
\renewcommand{\thelisting}{S\arabic{listing}}

\setcounter{section}{0}
\setcounter{figure}{0}
\setcounter{table}{0}
\setcounter{equation}{0}
\setcounter{algorithm}{0}
\setcounter{listing}{0}

\section{Architectural and Implementation Details of BATok}
\label{sec:supp-implementation}

This section specifies the BATok configuration used in the reported
experiments. Given a variable-length I/Q sequence, BATok lifts the
waveform into signal-derived features, encodes them at multiple
temporal resolutions, and produces a bounded token sequence for VLM
integration or standalone classification.

\subsection{Input Representation and Normalization}
\label{sec:supp-input-processing}

Each signal is represented as
$X\in\mathbb{R}^{2\times T}$, with the first and second channels
containing the in-phase and quadrature components. We independently
normalize each signal by its mean complex-sample power:
\begin{equation}
  \widetilde X
  =
  \frac{X}
  {\sqrt{\frac{1}{T}\sum_{t=1}^{T}
  \left(I_t^2+Q_t^2\right)+10^{-8}}}.
  \label{eq:supp-power-normalization}
\end{equation}
Signals in a minibatch are zero-padded to the longest signal in that
minibatch, while their valid lengths are retained separately. BATok
uses these lengths to construct branch masks and to remove padded
signal tokens before insertion into the VLM.

\subsection{Feature Lifting and Multi-Resolution Encoding}
\label{sec:supp-batok-config}

BATok explicitly lifts the two I/Q channels to five signal-derived
channels:
\begin{equation}
  U_t=[I_t,Q_t,A_t,P_t,\Delta\phi_t],
  \label{eq:supp-feature-lifting}
\end{equation}
where
\begin{equation}
  A_t=\sqrt{I_t^2+Q_t^2+10^{-8}},
  \qquad
  P_t=I_t^2+Q_t^2,
\end{equation}
and the inter-sample phase difference is implemented without an
explicit phase unwrap:
\begin{equation}
  \Delta\phi_t=
  \operatorname{atan2}
  \left(
  Q_tI_{t-1}-I_tQ_{t-1},
  I_tI_{t-1}+Q_tQ_{t-1}
  \right).
  \label{eq:supp-dphase}
\end{equation}
The first phase-difference value is zero-padded. A convolutional stem
with kernel size $7$, stride $1$, Group Normalization~\cite{wu2018groupnorm}, and GELU maps
the five channels to a $d_s$-dimensional temporal representation. In
the VLM experiments, $d_s=256$.

The projected sequence is processed by three parallel temporal
branches. Table~\ref{tab:supp-branch-config} gives their exact
downsampling configurations. Each branch first applies a
depthwise-separable convolution~\cite{howard2017mobilenets}, followed by one LiteMixer block in
the VLM experiments. A LiteMixer block alternates a residual
depthwise temporal convolution with a residual gated MLP; invalid
positions are masked after each block.

\FloatBarrier
\begin{table}[htbp]
\centering
\small
\begin{tabular}{lccc}
\toprule
Branch & Kernel (samples) & Stride (samples)
& Approximate length (samples) \\
\midrule
Small & 7 & 4 & $\lceil T/4\rceil$ \\
Mid   & 31 & 16 & $\lceil T/16\rceil$ \\
Large & 127 & 64 & $\lceil T/64\rceil$ \\
\bottomrule
\end{tabular}
\caption{Temporal-branch configuration used by BATok. The
output length is approximate because convolution padding and the
valid-length mask are handled separately.}
\label{tab:supp-branch-config}
\end{table}
\FloatBarrier

Each branch computes its own local-energy sequence directly from the
I/Q input rather than from its hidden features:
\begin{equation}
  E_{b}
  =
  \operatorname{AvgPool1D}_{k_b,s_b}
  \left(I^2+Q^2\right).
  \label{eq:supp-branch-energy}
\end{equation}
The pooling kernel $k_b$ and stride $s_b$ match the corresponding
temporal branch. Consequently, the small, mid, and large branches have
separate energy sequences aligned with their own temporal
resolutions.

\subsection{Length-Adaptive Budget and Scale Allocation}
\label{sec:supp-token-budget}

BATok determines its total token budget using a monotonic
piecewise-constant function of the input length $T$, as summarized in
Table~\ref{tab:supp-token-budget}.

\begin{table}[!htbp]
\centering
\small
\begin{tabular}{rc}
\toprule
Valid input length $T$ (samples) & Total budget $K(T)$ (tokens) \\
\midrule
$T\leq 2^{12}$ & 32 \\
$2^{12}<T\leq2^{15}$ & 48 \\
$2^{15}<T\leq2^{16}$ & 64 \\
$2^{16}<T\leq2^{18}$ & 96 \\
$2^{18}<T\leq2^{20}$ & 128 \\
$2^{20}<T\leq2^{22}$ & 192 \\
$T>2^{22}$ & 256 \\
\bottomrule
\end{tabular}
\caption{BATok token schedule. }
\label{tab:supp-token-budget}
\end{table}

The three scale budgets are
\begin{equation}
  K_s=\left\lfloor K/4\right\rfloor,\qquad
  K_m=\left\lfloor K/4\right\rfloor,\qquad
  K_l=K-K_s-K_m.
  \label{eq:supp-scale-budget}
\end{equation}
Thus, one quarter of the tokens is assigned to each of the small and
mid-branches, and the remaining half is assigned to the large branch.
For example, $K=48$ yields $(K_s,K_m,K_l)=(12,12,24)$. The standalone experiments are pure
modulation-classification evaluations on the original HisarMod2019
and \textsc{EMSpec-Sim} benchmarks, from which aligned subsets are
used to construct EMSpec-Instruct. The two primary standalone
benchmarks likewise activate 32 or 48 tokens; the remaining schedule
entries specify BATok's bounded behavior for longer valid inputs.

After resampling, each branch token receives a learnable scale
embedding. The three token sets are concatenated and refined by a
shared residual projection consisting of LayerNorm, a
$d_s$-to-$d_s$ linear layer, GELU, and dropout. For mixed-length
minibatches, BATok pads this output to the largest token count in the
minibatch and returns a Boolean token mask.

\subsection{Placeholder Expansion and Modality Handling}
\label{sec:supp-vlm-training}

The textual string \texttt{<signal>} is used only as
an insertion anchor and is not added as a new atomic vocabulary item.
During the forward pass, the integration layer locates the tokenizer
sequence corresponding to the anchor and replaces its embeddings with
the valid projected signal tokens. It simultaneously updates the
attention mask, position indices, and labels. Labels at signal-token
positions are excluded from the autoregressive objective, so the loss
is computed only on the target response.

The integration layer enforces modality consistency: a sample
containing a signal anchor must supply an I/Q input, and an I/Q input
must have a corresponding anchor. This check prevents a silent shift
between the signal batch and the textual batch. It also supports
mixed-modality minibatches through per-sample indices, so image--text,
signal--text, and image--signal--text examples can be trained
together.

\subsection{Backbone-Specific Integration}
\label{sec:supp-backbone-integration}

Table~\ref{tab:supp-vlm-config} summarizes the current integration
settings documented by the training configurations. Both backbones
use the same BATok architecture: $d_s=256$, one LiteMixer block per
branch, four resampler heads, zero signal-encoder dropout, a configured
query capacity of 512, and the active budget schedule in
Table~\ref{tab:supp-token-budget}. BATok and the signal projector are
fully trainable in both models, while the backbone language-model
layers are adapted with LoRA~\cite{hu2022lora}. All VLM variants are
fine-tuned for five epochs.

\FloatBarrier
\begin{table}[htbp]
\centering
\small
\setlength{\tabcolsep}{5pt}
\begin{tabular}{lcc}
\toprule
Setting & Qwen3-VL-2B & LLaVA-1.5-7B \\
\midrule
LLM adaptation & LoRA ($r=32$) & LoRA ($r=8$) \\
Vision tower & Trainable & Trainable \\
Native multimodal projector & Trainable & Trainable \\
BATok & Trainable & Trainable \\
Signal projector & Trainable & Trainable \\
Signal position handling & Extended M-RoPE & Learned 1D embedding \\
\bottomrule
\end{tabular}
\caption{Backbone-specific VLM integration settings. The language
models are adapted with LoRA, while BATok and its projector are
trained and saved as complete modules.}
\label{tab:supp-vlm-config}
\end{table}
\FloatBarrier

For Qwen3-VL~\cite{bai2025qwen3vl}, the integration layer first obtains the native
multimodal rotary position indices for image and text tokens. It then
inserts consecutive signal positions at the expanded anchor, shifts
subsequent positions by the change in sequence length, and updates the
multimodal token types. During cached decoding, the attention mask and
position indices are reconstructed from the actual prefill-cache
length.

For LLaVA-1.5~\cite{liu2023llava15}, the projected signal tokens receive a learned
one-dimensional position embedding before insertion. The described
LLaVA experiment initializes BATok from the Qwen3-VL-2B-trained signal
encoder, but constructs a new signal projector and signal position
embedding because the two backbones need not share a language hidden
dimension. The vision tower and native multimodal
projector remain trainable.

\subsection{Standalone Classification Head}
\label{sec:supp-standalone-training}

For standalone signal classification, BATok produces
$Z\in\mathbb{R}^{B\times K_{\max}^{(B)}\times d_s}$ and a validity mask
$M$. We aggregate the valid tokens using masked mean pooling:
\begin{equation}
  \bar z_i
  =
  \frac{\sum_k M_{i,k}Z_{i,k}}
  {\max(1,\sum_k M_{i,k})}.
\end{equation}
The pooled representation is passed through LayerNorm, a
$d_s$-dimensional hidden layer, GELU, dropout, and a final
classification layer. The training objective is cross-entropy with
optional label smoothing. Dataset-specific optimization settings are
reported with the corresponding experimental results rather than
inferred from architectural defaults. BATok and all task-specific
standalone classifiers are trained for 100 epochs.

\section{EMSpec-Instruct Construction Details}
\label{sec:supp-reproducibility}

This section details the construction procedure summarized in the
main paper. EMSpec-Instruct follows an \emph{observation-centered}
design: each observation comprises an I/Q sequence, its
deterministically rendered waterfall image, and a structured list of
signal instances. We assign observations to data partitions before
deriving instructions and targets. Consequently, all instructions
generated from the same observation remain in one partition, even
when a multi-signal scene yields several detection and grounding
queries.

\subsection{Source Observations and Partitioning}
\label{sec:supp-data-sources}

We use separate single-signal and multi-signal streams. The
single-signal stream combines \textsc{HisarMod-Sub}, a subset of
HisarMod2019.1 (hereafter HisarMod2019)~\cite{tekbiyik2020hisarmod},
with \textsc{EMSpec-Sim-Sub}, a subset of the complete
\textsc{EMSpec-Sim} single-signal simulation benchmark, and supplies
modulation-recognition examples. The multi-signal stream combines
SpaceNet~\cite{spacenet_dataset_chaspark} with our
multi-signal simulation and supplies structured-detection and
signal-grounding examples. Table~\ref{tab:supp-source-counts}
summarizes the number of source observations available to the final
construction procedure. We preserve the source-provided train/test assignment and perform all
subsequent sampling within the corresponding partition. In
particular, instructions are never randomly split after generation.
The tuple consisting of the source dataset, source split, source
sample identifier, and signal path is used as the observation
identity. Thus, all instructions derived from one observation remain
in the same partition. 

\begin{table*}[!htb]
\centering
\scriptsize
\setlength{\tabcolsep}{4pt}
\begin{tabular}{p{0.17\textwidth}p{0.10\textwidth}p{0.49\textwidth}rr}
\toprule
Source & Task stream & Category coverage
& \shortstack{Train\\(observations)}
& \shortstack{Test\\(observations)} \\
\midrule
\textsc{HisarMod-Sub} & Recognition &
26 classes: PSK$_6$, QAM$_7$, FSK$_4$, PAM$_3$, and six analog
AM/FM/PM variants & 44,000 & 22,000 \\
\textsc{EMSpec-Sim-Sub} & Recognition &
34 classes: generic digital/analog modulations, WiFi$_6$, BLE LE1M/LE2M,
LoRa 125/250/500\,kHz, Zigbee, and SRRC-shaped QPSK/16QAM
& 22,000 & 11,000 \\
SpaceNet & Det./ground. &
11 wireless classes: WiFi$_6$, BLE LE1M/LE2M, LoRa 250\,kHz,
Zigbee, and SRRC-shaped QPSK & 7,500 & 2,500 \\
Multi-signal simulation & Det./ground. &
14 classes: the SpaceNet vocabulary augmented with AM, FM, and
SRRC-shaped 16QAM & 20,000 & 7,500 \\
\midrule
Total & & & 93,500 & 43,000 \\
\bottomrule
\end{tabular}
\caption{Source observations and category coverage used to construct
EMSpec-Instruct. Counts refer to electromagnetic observations before
instruction expansion. WiFi$_6$ denotes the six combinations of
20/40-MHz bandwidth and QPSK/16QAM/64QAM modulation.}
\label{tab:supp-source-counts}
\end{table*}

\paragraph{Simulation procedure.}
The simulated observations are generated with RF waveform-generation
utilities following the TorchSig simulation framework~\cite{boegner2022torchsig},
from a fixed class map and
seeded random state. For the single-signal stream, we sample a class
with approximately uniform frequency, synthesize its baseband
waveform, apply the class-specific modulation and pulse shaping, and
add complex Gaussian noise at an SNR selected from
\[
\{-20,-15,-10,-5,0,5,10,15,20,25,30\}\ {\rm dB}.
\]
Sequence length is sampled from
\[
\{1{,}024,2{,}048,4{,}096,8{,}192,16{,}384,32{,}768\},
\]
while the nominal sample rate is selected from 1, 2, 5, 10, or
20\,MHz; wider rates are used when required by WiFi signals. This
provides approximately balanced coverage of all 34 simulated classes.

For the multi-signal stream, two or more independently synthesized
waveforms are placed in a common time--frequency canvas. We sample
the class, active time interval, center frequency, occupied
bandwidth, SNR, near--far power offset, and channel-frequency offset
for every instance. Placement is stratified over non-overlap,
time-only overlap, frequency-only overlap, partial time--frequency
overlap, and near-complete overlap configurations. Signals are
shifted, gated, scaled, and summed in complex baseband; the clean
placement parameters are retained as instance metadata. Bounding
boxes are therefore obtained from generation parameters rather than
estimated from the rendered image. The simulation complements the
source datasets with controlled class balance, signal density,
acquisition length, SNR, and overlap difficulty.

\subsection{Instance Metadata}
\label{sec:supp-observation-representation}

Every signal instance is represented by
\begin{equation}
  o_i=(c_i,[t_i^s,t_i^e],[f_i^l,f_i^h],b_i,a_i),
  \label{eq:supp-instance-tuple}
\end{equation}
where $c_i$ is the canonical class name, the next two terms are its
time and frequency intervals, $b_i$ is its normalized bounding box,
and $a_i$ contains optional attributes such as SNR, signal family,
modulation, or communication standard. Class aliases are canonicalized
before sampling; for example, order-reversed names such as
\texttt{QAM16} and \texttt{16QAM} map to the same category.

\subsection{Time--Frequency Box Conversion}
\label{sec:supp-box-conversion}

Let an observation span $[0,D]$ milliseconds and
$[F_{\min},F_{\max}]$ MHz. For an instance with time interval
$[t_i^s,t_i^e]$ and frequency interval $[f_i^l,f_i^h]$, its
continuous normalized coordinates are
\begin{align}
  x_1 &= 1000\,t_i^s/D,
  &x_2 &= 1000\,t_i^e/D, \nonumber\\
  y_1 &= 1000\left(1-\frac{f_i^h-F_{\min}}
                              {F_{\max}-F_{\min}}\right),
  &y_2 &= 1000\left(1-\frac{f_i^l-F_{\min}}
                              {F_{\max}-F_{\min}}\right).
  \label{eq:supp-bbox-conversion}
\end{align}
The inversion of the frequency coordinate is consistent with the
image convention that high frequency appears at the top. Coordinates
are rounded to integers, clamped to $[0,1000]$, and sorted along each
axis. We reject boxes with non-positive width or height. The
coordinate representation is independent of the raster resolution,
which allows images to be resized by the VLM processor without
changing the targets.

All time- or frequency-conditioned questions explicitly state the
physical observation range. For image-containing examples, the
standard context describes time from left to right and frequency from
bottom to top. It also defines the box ordering as
$[x_{\min},y_{\min},x_{\max},y_{\max}]$. Signal-only prompts contain
the same numerical ranges without referring to visual axes. This
context prevents a model from having to infer milliseconds or MHz
from normalized box coordinates alone.

\subsection{Task Construction}
\label{sec:supp-task-construction}

We first construct a deterministic semantic specification for every
instruction. This specification contains the task family, query
subtype, condition, matched instance identifiers, exact target boxes,
and negative-query flag. Natural-language realization is performed
only after this specification has been fixed. 

\paragraph{Modulation recognition.}
Each single-signal observation supplies a class-prediction semantic
specification. Seeded quota sampling and modality packaging may
produce multiple instruction records from the same observation, but
all such records remain in the same data partition.
The training subset is class-balanced over 50 modulation and
communication-signal categories, with 2,400 instructions per class
and 120,000 instructions in total. Qwen3.5-Max~\cite{qwen2026qwen35} generates diverse
recognition questions from the task name and modality configuration;
the answer is then filled directly with the canonical class label
from the source annotation. The target is a one-element class list.

\paragraph{Structured detection.}
Each multi-signal observation yields one all-signal detection
instruction. The target contains every valid instance in the scene.
Boxes and class labels are sorted in the same instance order so that
the $j$-th box corresponds to the $j$-th class. The multi-signal
class universe contains 14 classes:
\texttt{AM}, \texttt{FM}, \texttt{BLE LE1M}, \texttt{BLE LE2M},
\texttt{LoRa 250KHZ}, \texttt{SRRC QPSK}, \texttt{SRRC 16QAM},
six WiFi bandwidth/modulation combinations, and \texttt{Zigbee}.
As in recognition, Qwen3.5-Max generates the instruction, whereas the
complete ordered box/class response is serialized directly from the
scene annotation.

\paragraph{Language-conditioned signal grounding.}
Grounding instructions return the boxes of all instances satisfying a
language condition. We use the following condition families.
\begin{itemize}
  \item \emph{Class}: exact match after class-name canonicalization.
  \item \emph{Time or frequency}: non-empty interval intersection
        between the instance and the requested range.
  \item \emph{Class--time--frequency composition}: conjunction of the
        class and both interval conditions.
  \item \emph{Duration and bandwidth relations}: all instances tied
        for the longest duration or widest occupied bandwidth.
  \item \emph{Waterfall strength}: strongest or weakest instances
        according to image evidence.
  \item \emph{Overlap}: instances whose time and frequency intervals
        both intersect another instance.
\end{itemize}

The waterfall-strength attribute is computed rather than obtained
from a language model. Specifically, we use the 90th percentile of
grayscale intensity inside each normalized instance box. Using a high
percentile is more robust than a maximum, which can be dominated by a
single bright pixel, and less likely than the mean to be diluted by
background pixels within a loose box.

\paragraph{Negative grounding.}
Negative examples are part of signal grounding rather than a separate
task. We construct an absent-class query by selecting a class outside
the scene's class set. We also construct a class-conditioned
time--frequency query whose conjunction matches no instance. Both
types have a deterministic empty target. This forces the model to
distinguish failure to satisfy a condition from failure to follow the
requested output format.

The training policy first creates six positive and two negative
grounding instructions per multi-signal scene, then adds positive
subtypes to meet the target task distribution. The held-out
localization set contains one or two positive questions and one
absent-class question per scene. Its positive subset additionally
contains overlap questions when overlapping instances exist; scenes
without an overlap may contribute a class-conditioned fallback
question.

\subsection{Metadata-Conditioned Instruction Realization}
\label{sec:supp-language-generation}

All natural-language instruction records are realized with
Qwen3.5-Max. Its
role depends on the task. For modulation recognition and structured
detection, the model generates only the question; the compact answer
is serialized directly from the canonical class label or complete
scene annotation. For signal grounding, it generates both the
metadata-conditioned question and answer. A grounding record is
retained only when its returned boxes and classes agree exactly with
the programmatically computed match set. Qwen3.5-Max therefore
controls linguistic realization but cannot alter the supervision.

We invoke Qwen3.5-Max through a two-message chat interface. The
\texttt{system} message assigns an RF-domain role and global
faithfulness constraints. The \texttt{user} message contains the
task-specific instruction, scene duration, observed frequency range,
and a compact list of instances. Every instance includes its
canonical class, normalized box, time/frequency intervals, and
optional SNR or protocol attributes. The user message also contains
the requested condition, expected match set, and output schema.
Accordingly, the complete input has four logical parts: system role,
task-specific instruction, scene metadata, and output constraints.
The system message is provided verbatim below:

\begin{quote}
\small
\emph{You are an expert in radio-frequency signal analysis,
electromagnetic spectrum monitoring, and time-frequency signal
grounding. Your task is to generate accurate and diverse English
query-answer pairs from the provided signal metadata. Strictly follow
the supplied class labels, time ranges, frequency ranges, SNR values,
and normalized bounding boxes. Do not infer or invent any signal,
attribute, coordinate, or physical range that is not present in the
metadata. Return valid JSON only.}
\end{quote}

For this input, a valid generated query is:
\emph{``Within the 150-ms observation spanning 2430--2480 MHz,
locate signals intersecting 2470--2474 MHz.''} The generated answer
mentions both matched classes and contains exactly
\texttt{<bbox>93,157,360,163</bbox>} and
\texttt{<bbox>547,60,1000,860</bbox>}. A recognition input is
shorter: it provides the task, available modalities, and canonical
label, and asks Qwen3.5-Max for a concise modulation-classification
question. A detection input provides the scene context and all
instance metadata and asks only for a diverse formulation of
``detect all signals''; its answer is not generated by Qwen3.5-Max.

The Qwen3.5-Max output is accepted only when all of the following checks
pass: (1) the number and order of returned items match the requested
items; (2) every query is non-empty and retains explicit
millisecond/MHz scene context when required; (3) class names and
numerical condition ranges are consistent with the semantic
specification; (4) grounding answers contain every and only expected
box, while negative answers contain no box and explicitly state that
no matching signal exists; and (5) the response parses as the
requested JSON object. The model is forbidden from inventing classes,
boxes, physical ranges, or attributes. Only validated Qwen3.5-Max records enter the dataset. 

\subsection{Compact Target Grammar}
\label{sec:supp-output-formats}

All targets use a compact, machine-parseable grammar. Let
$B=[[x_1,y_1,x_2,y_2],\ldots]$ be a list of boxes and
$C=[c_1,\ldots]$ the corresponding class list. The three task
families use
\begin{align}
  \text{modulation recognition:}\quad
  &\texttt{<c>}C\texttt{</c>}, \nonumber\\
  \text{structured detection:}\quad
  &\texttt{<d><b>}B\texttt{</b><c>}C
    \texttt{</c></d>}, \nonumber\\
  \text{signal grounding:}\quad
  &\texttt{<b>}B\texttt{</b>}.
  \label{eq:supp-output-grammar}
\end{align}
For a negative grounding query, $B$ is the empty list and the exact
target is \texttt{<b>[]</b>}. This grammar separates semantic
evaluation from free-form wording: modulation classes, boxes, and
box--class correspondence can be parsed without another language
model. Table~\ref{tab:supp-format-examples} gives representative
examples with numerical values shortened for readability.

\begin{table}[tb]
\centering
\small
\begin{tabular}{p{0.16\linewidth}p{0.42\linewidth}p{0.34\linewidth}}
\toprule
Task & Example instruction & Target \\
\midrule
Recognition &
Classify the modulation using both modalities. &
\texttt{<c>["8ASK"]</c>} \\
\addlinespace
Detection &
Detect every signal and return all boxes and classes. &
\texttt{<d><b>[[339,166,982,833]]</b>}\newline
\texttt{<c>["16QAM"]</c></d>} \\
\addlinespace
Grounding &
Find regions that overlap 2419.589--2448.466 MHz. &
\texttt{<b>[[339,166,982,833],...]</b>} \\
\addlinespace
Negative grounding &
Find all BLE LE2M regions in this scene. &
\texttt{<b>[]</b>} \\
\bottomrule
\end{tabular}
\caption{Representative task realizations and target formats. Scene
context is abbreviated in the table but is present in the actual
time/frequency-conditioned prompts.}
\label{tab:supp-format-examples}
\end{table}

\subsection{Sampling, Modalities, and Final Statistics}
\label{sec:supp-data-statistics}

The final training set contains 400,000 instruction records. We
allocate 30\% to single-signal modulation recognition and 70\% to
multi-signal understanding. The latter is divided into one detection
instruction per source scene and multiple grounding instructions.
Table~\ref{tab:supp-task-statistics} reports the exact train and test
sizes.

\begin{table}[tb]
\centering
\small
\begin{tabular}{lrrrr}
\toprule
Task & Train (records) & Train (\%) & Test (records) & Test (\%) \\
\midrule
Modulation recognition & 120,000 & 30.00 & 33,000 & 48.61 \\
Structured detection & 27,500 & 6.88 & 10,000 & 14.73 \\
Signal grounding & 252,500 & 63.12 & 24,883 & 36.66 \\
\midrule
Total & 400,000 & 100.00 & 67,883 & 100.00 \\
\bottomrule
\end{tabular}
\caption{Final task composition. The validation set is a balanced
3,000-example subset sampled from the held-out task files and is not
included in the test total.}
\label{tab:supp-task-statistics}
\end{table}

Table~\ref{tab:supp-grounding-statistics} gives the grounding
subtype distribution. The training counts reflect quota-based
sampling rather than the raw number of candidate questions. The
``class fallback'' test subtype is used only when an overlap query
cannot be constructed.

\begin{table}[H]
\centering
\small
\begin{tabular}{lrr}
\toprule
Grounding subtype & Train (records) & Test (records) \\
\midrule
Class & 27,500 & 1,631 \\
Class fallback (no overlap) & -- & 313 \\
Time & 27,500 & 1,656 \\
Frequency & 27,500 & 1,640 \\
Class--time--frequency & 27,500 & 1,656 \\
Longest duration & 27,500 & 1,638 \\
Widest bandwidth & 27,500 & 1,689 \\
Strongest brightness & 27,500 & 1,617 \\
Weakest brightness & 5,000 & 1,692 \\
Time--frequency overlap & -- & 1,351 \\
Negative class & 27,500 & 10,000 \\
Negative class--condition & 27,500 & -- \\
\midrule
Total & 252,500 & 24,883 \\
\bottomrule
\end{tabular}
\caption{Signal-grounding subtype counts. A dash denotes a subtype
that is not selected for that partition.}
\label{tab:supp-grounding-statistics}
\end{table}

We construct aligned \emph{image} and \emph{signal} JSONL
configurations. The image configuration presents every record as
waterfall-image plus text. The signal configuration is used to train
the unified model and contains 298,000 joint image--signal records,
74,000 signal-only records, and 28,000 image-only records. The
underlying task target, source sample identifier, and train/test
membership are identical across configurations. This alignment
supports controlled comparison between an image-only VLM and a VLM
equipped with BATok. Including image--text, signal--text, and
image--signal--text instructions in one training set exposes the
model to the same task interface under different modality configurations.

The data cover a wide range of observation lengths. After
deduplicating repeated instructions by source observation, the
training partition contains 93,500 signal observations with lengths
from 1,024 to 15,000,000 complex samples. Its median is 1,024, the
95th percentile is 6,000,000, and the mean is approximately
1.05 million samples.

\section{Extended Experimental Evaluation}
\label{sec:supp-additional-experiments}

\subsection{Evaluation Metrics}
\label{sec:supp-evaluation-metrics}

For modulation recognition, we report overall accuracy (OA) and
Macro-F1. OA measures the proportion of correctly classified samples,
whereas Macro-F1 averages class-wise F1 scores and therefore reflects
performance across modulation classes without allowing frequent
classes to dominate the result. For VLM outputs, we additionally
report the parse rate, defined as the proportion of responses that
conform to the required task-specific output format. This separates
semantic task performance from the ability to follow the structured
response protocol.

For structured detection and positive-query grounding, we use
F1@0.5 and mean intersection over union (mIoU). A predicted box is
considered matched when its IoU with a ground-truth box is at least
0.5. Detection matching is class-aware and additionally requires the
correct signal label, whereas grounding evaluates the boxes selected
by the query and does not involve a class prediction. F1@0.5 jointly
captures missed and spurious targets through precision and recall,
while mIoU measures the spatial agreement of matched boxes. We report
precision and recall separately for the task-specific detectors to
expose their false-positive and false-negative trade-off. For
unmatched grounding queries, where no target box exists, empty-set
accuracy measures whether the model correctly returns no box.

The gain-decomposition analysis further reports class-agnostic
F1@0.5 and matched-box mIoU together with conditional class accuracy,
allowing target discovery and box geometry to be separated from
signal-category prediction. Class-agnostic mAP@0.5 and
mAP@0.5:0.95 provide complementary precision--recall summaries across
localization thresholds; because the generated boxes do not carry
confidence scores, their output order is used for this secondary
analysis. For efficiency comparisons, parameter count measures model
size, latency measures batch-size-one inference time in milliseconds
per sample, and the realized token count \(K\) measures the
representation budget presented to the downstream model.

\subsection{Spectrum Understanding Across Input Configurations}
\label{sec:supp-input-results}

Table~\ref{tab:supp-modality-breakdown} isolates the contribution of
each input pathway. Each BATok-enhanced VLM is trained as one
checkpoint with image--text, signal--text, and
image--signal--text instructions. At test time, the three input
configurations share the same held-out observations, prompts, and
targets, so their differences directly reflect the evidence supplied
by each modality.

\begin{table*}[t]
\centering
\scriptsize
\setlength{\tabcolsep}{2.1pt}
\begin{tabular}{l l c c c c c c c c c c}
\toprule
& &
\multicolumn{3}{c}{Mod. Rec.} &
\multicolumn{3}{c}{Detection} &
\multicolumn{4}{c}{Grounding} \\
\cmidrule(lr){3-5}\cmidrule(lr){6-8}\cmidrule(lr){9-12}
Backbone & Test Input &
\shortstack{OA\\(\%) $\uparrow$} &
\shortstack{Macro-F1\\(\%) $\uparrow$} &
\shortstack{Parse\\(\%) $\uparrow$} &
\shortstack{F1@0.5\\(\%) $\uparrow$} &
\shortstack{mIoU\\(\%) $\uparrow$} &
\shortstack{Parse\\(\%) $\uparrow$} &
\shortstack{F1@0.5\\(\%) $\uparrow$} &
\shortstack{mIoU\\(\%) $\uparrow$} &
\shortstack{Parse\\(\%) $\uparrow$} &
\shortstack{Empty Acc.\\(\%) $\uparrow$} \\
\midrule
Qwen3-VL-2B + BATok & Image + Text
& 16.52 & 11.93 & 100.00
& 64.12 & 95.20 & 97.88
& 81.78 & 95.23 & 99.94 & 92.04 \\
Qwen3-VL-2B + BATok & Signal + Text
& 40.36 & 40.33 & 88.96
& 3.64 & 72.48 & 99.44
& 13.36 & 74.37 & 99.90 & 89.39 \\
Qwen3-VL-2B + BATok & Image + Signal + Text
& 40.91 & 41.26 & 86.02
& 64.27 & 95.20 & 98.29
& 83.00 & 95.24 & 99.96 & 92.43 \\
\midrule
LLaVA-1.5-7B + BATok & Image + Text
& 3.08 & 2.54 & 99.35
& 47.69 & 89.28 & 98.19
& 67.75 & 89.29 & 99.88 & 85.15 \\
LLaVA-1.5-7B + BATok & Signal + Text
& 43.68 & 43.02 & 94.90
& 3.06 & 74.68 & 98.80
& 12.85 & 75.05 & 99.81 & 89.11 \\
LLaVA-1.5-7B + BATok & Image + Signal + Text
& 44.14 & 43.20 & 94.95
& 47.80 & 89.40 & 97.17
& 68.27 & 89.39 & 99.87 & 91.23 \\
\bottomrule
\end{tabular}
\caption{BATok-enhanced VLM performance by input configuration.
Values are percentages, and all input configurations for a backbone
use the same trained checkpoint. Parse denotes successful task-format
parsing, and Empty Acc. is reported for unmatched grounding queries.}
\label{tab:supp-modality-breakdown}
\end{table*}

The same-checkpoint comparison separates the contribution of test-time
signal input from the overall effect of BATok integration and
mixed-modality training. For Qwen3-VL-2B+BATok, image--text input
already achieves 64.12\% detection F1, and adding signal input raises
it to 64.27\%. The increase from 62.25\% to 64.27\% in
Table~\ref{tab:unified_results} therefore reflects the overall change
in the model and training setup, whereas the increase from 64.12\%
to 64.27\% measures the additional contribution of waveform evidence
within the enhanced checkpoint.

The modalities play different roles across tasks. Signal--text input
achieves recognition OA of 40.36\% and 43.68\% for Qwen3-VL-2B and
LLaVA-1.5-7B, respectively, while its detection and grounding F1
remain low. Image--text input provides the main spatial evidence for
localization. Joint input raises recognition OA to 40.91\% and
44.14\% and improves grounding F1 over image--text input by 1.22 and
0.52 percentage points, respectively. These results support
complementary roles for waveform evidence and waterfall geometry
within the unified instruction interface.

\subsection{Validation of EMSpec-Instruct}
\label{sec:supp-dataset-validation}

We evaluate whether the recognition and detection subsets of
EMSpec-Instruct provide valid, learnable supervision for their
intended tasks. Task-specific classifiers and detectors assess the
observations, class labels, and bounding-box annotations assembled
from existing and simulated signal sources, while VLM+BATok evaluates
whether the same data can support a unified instruction-conditioned
interface. The aligned recognition experiment additionally reveals
how the raw I/Q observations and their derived waterfall images
preserve modulation-class evidence.

\subsubsection{Multi-Task Instruction Supervision}

We evaluate whether the three task families in EMSpec-Instruct provide
complementary supervision for language-conditioned signal grounding.
The Qwen3-VL-2B+BATok architecture and grounding test set are fixed,
while the instruction-tuning data contain grounding alone or
grounding augmented with modulation recognition, structured
detection, or both. Table~\ref{tab:task_composition} measures how
these complementary task annotations improve grounding.

\begin{table}[!htb]
\centering
\small
\setlength{\tabcolsep}{4pt}
\begin{tabular}{ccc cc}
\toprule
\multicolumn{3}{c}{Training tasks} &
\multicolumn{2}{c}{Grounding} \\
\cmidrule(lr){1-3}\cmidrule(lr){4-5}
Mod. Rec. & Detection & Grounding &
F1@0.5 (\%) $\uparrow$ & mIoU (\%) $\uparrow$ \\
\midrule
 &  & \checkmark
& 81.97 & 95.15 \\
\checkmark &  & \checkmark
& 81.99 & \textbf{95.24} \\
 & \checkmark & \checkmark
& 81.47 & 95.06 \\
\checkmark & \checkmark & \checkmark
& \textbf{83.00} & \textbf{95.24} \\
\bottomrule
\end{tabular}
\caption{Grounding performance under different instruction-tuning
task compositions. All variants use Qwen3-VL-2B+BATok and are
evaluated on the same grounding test set.}
\label{tab:task_composition}
\end{table}

The complete three-task mixture achieves the highest grounding
F1@0.5 (83.00\%) and ties the highest mIoU (95.24\%).
Compared with grounding-only training, jointly incorporating
modulation-recognition and structured-detection supervision improves
F1@0.5 by 1.03 percentage points and mIoU by 0.09 percentage points.
These results indicate that joint training on the three tasks helps
improve language-conditioned signal grounding in this experimental
setting.

\subsubsection{Modulation-Recognition Learnability}

Table~\ref{tab:supp-modality-comparison} evaluates task-specific
modulation recognition on the union of \textsc{HisarMod-Sub} and
\textsc{EMSpec-Sim-Sub}, the two aligned modulation-recognition
subsets of EMSpec-Instruct. Its primary purpose, together with the
structured-detection evaluation below, is to verify that the
constructed observations and annotations support learnable
recognition and detection tasks. We include classifiers operating on
both waterfall images and raw I/Q observations. Because the two
modalities are derived from the same source observations and use the
same partitions, their results also indicate how much
modulation-class evidence is retained by each representation. The
image references use ResNet-50~\cite{he2016resnet} and
ViT~\cite{dosovitskiy2021vit}; the raw-I/Q references include
MAMC~\cite{10705364}, the Signal Transformer
(SiT)~\cite{zhai2025cdsit}, and DNCNet with its ResNet18-1D signal
backbone~\cite{du2022dncnet}.

\begin{table}[H]
\centering
\small
\setlength{\tabcolsep}{5pt}
\begin{tabular}{l l c c c c}
\toprule
Method & Input & Acc. (\%) $\uparrow$ & Macro-F1 (\%) $\uparrow$
& Params (M) $\downarrow$ & Latency (ms/sample) $\downarrow$ \\
\midrule
ResNet-50 & Waterfall
& 32.43 & 34.33 & 23.610 & 4.105 \\
ViT & Waterfall
& 31.06 & 33.13 & 85.837 & 3.424 \\
MAMC~\cite{10705364} & I/Q
& 29.62 & 27.64 & \textbf{0.009} & 1.466 \\
SiT~\cite{zhai2025cdsit} & I/Q
& 41.56 & 41.43 & 0.654 & \textbf{0.967} \\
DNCNet~\cite{du2022dncnet} & I/Q
& 44.17 & 42.80 & 3.870 & 1.860 \\
BATok & I/Q
& \textbf{49.36} & \textbf{47.19} & 1.047 & 5.443 \\
\bottomrule
\end{tabular}
\caption{Waterfall-image and raw-I/Q modulation recognition on the
aligned EMSpec-Instruct subset, defined as the union of
\textsc{HisarMod-Sub} and \textsc{EMSpec-Sim-Sub}. Latency is
measured in milliseconds per sample with batch size 1 on one NVIDIA
GeForce RTX 4090.}
\label{tab:supp-modality-comparison}
\end{table}

The image- and raw-I/Q-based classifiers both recover modulation
structure from the aligned subsets, confirming that the constructed
observations and class annotations support task-specific recognition.
BATok records 49.36\% accuracy and 47.19\% Macro-F1, whereas the
strongest waterfall-image result reaches 32.43\% accuracy and 34.33\%
Macro-F1. Beyond validating the dataset, this consistent advantage of
the raw-I/Q classifiers supports the observation that the source
signals retain modulation-class evidence that is not fully preserved
in their derived waterfall images. BATok obtains this result with
1.047M parameters, fewer than
DNCNet and far fewer than the image encoders, while its 5.443-ms
latency is higher. The difference arises because parameter count
measures stored learnable weights, whereas latency is dominated by
operations over the input sequence: BATok applies compact shared
weights across multiple temporal resolutions and then performs
evidence-aware resampling over the resulting features. These
length-dependent operations and their memory access make BATok
small in parameter count but more compute-intensive at the tokenizer
stage.
The dedicated BATok classifier also exceeds Qwen3-VL-2B+BATok and
LLaVA-1.5-7B+BATok by 8.45 and 5.22 Accuracy points, respectively,
providing a direct measure of the signal representation before it is
used by the unified generative interface.

\subsubsection{Structured-Detection Learnability}

We train an Ultralytics YOLO implementation~\cite{jocher2023ultralytics}
and DETR~\cite{carion2020detr} for 100 epochs on the structured
detection subset of EMSpec-Instruct. Their dedicated detection heads
provide task-specific references for evaluating the consistency and
learnability of the constructed boxes and class annotations.

\begin{table}[H]
\centering
\small
\setlength{\tabcolsep}{4pt}
\begin{tabular}{l c c c c}
\toprule
Method & mIoU (ratio) $\uparrow$ & F1@0.5 (ratio) $\uparrow$
& Precision@0.5 (ratio) $\uparrow$ & Recall@0.5 (ratio) $\uparrow$ \\
\midrule
YOLO26n
& 0.9291 & 0.6466 & 0.6288 & 0.6654 \\
YOLO26l
& \underline{0.9475} & \underline{0.6865}
& \underline{0.6514} & \underline{0.7255} \\
YOLO26x
& \textbf{0.9517} & \textbf{0.6925}
& \textbf{0.6577} & \textbf{0.7311} \\
DETR
& 0.8597 & 0.4127 & 0.2918 & 0.7044 \\
\bottomrule
\end{tabular}
\caption{Task-specific detection results on the EMSpec-Instruct
detection test set. Best and second-best values are bolded and
underlined, respectively. These systems use dedicated detection
heads.}
\label{tab:supp-detection-specialists}
\end{table}

YOLO26x provides the strongest task-specific detection result,
reaching an mIoU of 0.9517 and F1@0.5 of 0.6925. DETR obtains
0.8597 and 0.4127, respectively, while its higher recall than
precision indicates more false-positive predictions. The consistent
performance across YOLO variants and DETR confirms that the
constructed boxes and class annotations provide learnable supervision
for structured detection. Detection through the unified generative
interface is evaluated separately in
Table~\ref{tab:supp-modality-breakdown}.

\subsection{Analysis of BATok Performance Gains}
\label{sec:supp-gain-source}

We further separate modulation semantics from box geometry in the
structured-detection results. The main metric is class-aware: a
prediction is a true positive only when its class is correct and its
IoU is at least 0.5. We additionally recompute class-agnostic metrics
by matching boxes solely by IoU, and report conditional class
accuracy over predictions already matched to a ground-truth box at
IoU$\geq0.5$. Table~\ref{tab:supp-gain-decomposition} applies all
three views to the same predictions used in the main comparison.

\begin{table}[H]
\centering
\small
\setlength{\tabcolsep}{4pt}
\begin{tabular}{l l c c}
\toprule
Backbone & Detection measure
& Image + Text (\%) & Image + Signal + Text (\%) \\
\midrule
Qwen3-VL-2B
& Class-aware F1@0.5 & 62.25 & 64.27 \\
& Class-agnostic F1@0.5 & 90.74 & 90.68 \\
& Conditional class accuracy & 68.57 & 70.72 \\
& Class-agnostic matched-box mIoU & 95.02 & 95.48 \\
& Class-agnostic mAP@0.5 & 82.87 & 82.66 \\
& Class-agnostic mAP@0.5:0.95 & 71.23 & 72.19 \\
\midrule
LLaVA-1.5-7B
& Class-aware F1@0.5 & 45.99 & 47.80 \\
& Class-agnostic F1@0.5 & 72.58 & 75.30 \\
& Conditional class accuracy & 61.43 & 62.38 \\
& Class-agnostic matched-box mIoU & 88.48 & 89.20 \\
& Class-agnostic mAP@0.5 & 54.89 & 58.60 \\
& Class-agnostic mAP@0.5:0.95 & 37.02 & 40.68 \\
\bottomrule
\end{tabular}
\caption{Decomposition of structured-detection gains into class
semantics and box geometry. Class-agnostic metrics ignore the
predicted class during IoU matching; conditional class accuracy is
computed only over boxes matched at IoU$\geq0.5$.}
\label{tab:supp-gain-decomposition}
\end{table}

For Qwen3-VL-2B, adding BATok raises class-aware F1 by 2.02 points
while class-agnostic F1 remains essentially unchanged. Conditional
class accuracy instead increases by 2.15 points, and matched-box mIoU
rises by only 0.46 points. Its detection gain therefore comes
primarily from more accurate signal-category predictions, with a
smaller improvement in box tightness. For LLaVA-1.5-7B,
class-agnostic F1 and mAP@0.5 increase by 2.72 and 3.71 points,
whereas conditional class accuracy improves by 0.95 points. Here,
BATok mainly improves target discovery and localization while also
providing a smaller semantic benefit. The source of the gain is thus
backbone dependent: predominantly semantic for Qwen3-VL-2B and
jointly geometric--semantic for LLaVA-1.5-7B.

Grounding contains no class prediction and is therefore outside this
class--geometry decomposition. In the main results, BATok produces
smaller, non-uniform grounding changes: matched-box mIoU increases by
0.49 and 0.58 points for Qwen3-VL-2B and LLaVA-1.5-7B, while
empty-set accuracy decreases by 0.76 and 0.69 points. BATok should
therefore be interpreted as complementary signal evidence rather
than a uniform improvement to every localization behavior. Because
the generated boxes have no confidence scores, the mAP calculation
uses output order; we consequently interpret it jointly with F1,
conditional class accuracy, and matched-box mIoU.

\subsection{Fine-Grained Grounding Analysis}
\label{sec:supp-grounding-breakdown}

Empty-set accuracy is reported only for unmatched queries, for which
no ground-truth box exists; it is not substituted for an IoU-based
localization metric.
Table~\ref{tab:supp-grounding-breakdown} reports this breakdown for
both BATok-enhanced VLM backbones.

\begin{table*}[!htb]
\centering
\small
\setlength{\tabcolsep}{5pt}
\begin{tabular}{l c c c c c c c c}
\toprule
& \multicolumn{4}{c}{Qwen3-VL-2B + BATok}
& \multicolumn{4}{c}{LLaVA-1.5-7B + BATok} \\
\cmidrule(lr){2-5}\cmidrule(lr){6-9}
Query Type &
\shortstack{F1@0.5\\(\%) $\uparrow$} &
\shortstack{mIoU\\(\%) $\uparrow$} &
\shortstack{Empty Acc.\\(\%) $\uparrow$} &
\shortstack{Parse\\(\%) $\uparrow$} &
\shortstack{F1@0.5\\(\%) $\uparrow$} &
\shortstack{mIoU\\(\%) $\uparrow$} &
\shortstack{Empty Acc.\\(\%) $\uparrow$} &
\shortstack{Parse\\(\%) $\uparrow$} \\
\midrule
Modulation-conditioned
& 83.51 & 94.85 & -- & 99.94
& 66.59 & 87.95 & -- & 100.00 \\
Time-conditioned
& 90.28 & 95.22 & -- & 99.88
& 74.07 & 89.24 & -- & 99.58 \\
Frequency-conditioned
& 87.02 & 94.82 & -- & 100.00
& 74.16 & 88.91 & -- & 99.88 \\
Compositional
& 89.46 & 94.50 & -- & 99.94
& 79.49 & 88.80 & -- & 99.88 \\
Overlap / relational
& 78.00 & 95.58 & -- & 99.92
& 66.13 & 90.01 & -- & 99.87 \\
Unmatched / negative
& -- & -- & 92.43 & 100.00
& -- & -- & 91.23 & 99.88 \\
\bottomrule
\end{tabular}
\caption{Language-conditioned grounding results by query type for the
image--signal--text test configuration. Values are percentages; a
dash marks an inapplicable metric, and Parse denotes successful
required-format parsing. Overlap/relational pools duration, bandwidth,
brightness, and overlap queries, while unmatched pools negative-class
and class-fallback-no-overlap queries.}
\label{tab:supp-grounding-breakdown}
\end{table*}

Qwen3-VL-2B+BATok exceeds its LLaVA counterpart in F1 across every
positive grounding subtype, with the largest F1 among its own
subtypes on time-conditioned queries (90.28\%). Both models maintain
near-perfect parse rates, showing reliable instruction following
across class-, time-, frequency-, compositional, relational, and
negative queries. The variation across query types is therefore
driven primarily by query-conditioned localization.

\section{Extended Ablation and Efficiency Analysis}
\label{sec:supp-ablations}

\subsection{Benchmark-Level Classification Performance and Efficiency}
\label{sec:supp-standalone-efficiency}

Table~\ref{tab:supp-standalone-per-benchmark} expands the averaged
results in the main paper into HisarMod2019 and \textsc{EMSpec-Sim}
results, together with parameter count and measured runtime. All
models are independently trained for modulation classification on the
complete source benchmarks; the aligned source subsets used for
dataset validation are evaluated separately in
Table~\ref{tab:supp-modality-comparison}.

\begin{table}[H]
\centering
\small
\resizebox{\textwidth}{!}{%
\begin{tabular}{l cc cc cc}
\toprule
& \multicolumn{2}{c}{HisarMod2019}
& \multicolumn{2}{c}{\textsc{EMSpec-Sim}}
& & \\
\cmidrule(lr){2-3}\cmidrule(lr){4-5}
Method & Acc. (\%) $\uparrow$ & Macro-F1 (\%) $\uparrow$
& Acc. (\%) $\uparrow$ & Macro-F1 (\%) $\uparrow$
& Params (M) $\downarrow$ & Latency (ms/sample) $\downarrow$ \\
\midrule
MAMC~\cite{10705364}
& 55.22 & 53.23
& 31.18 & 31.39
& \textbf{0.008} & 1.470 \\
SiT~\cite{zhai2025cdsit}
& 71.34 & 71.30
& 35.78 & 35.41
& 0.649 & \textbf{0.978} \\
DNCNet~\cite{du2022dncnet}
& \underline{75.25} & \underline{75.19}
& \textbf{41.11} & \underline{38.58}
& 3.856 & 1.868 \\
BATok
& \textbf{77.53} & \textbf{77.53}
& \underline{39.75} & \textbf{39.54}
& 1.044 & 5.509 \\
\bottomrule
\end{tabular}}
\caption{Per-benchmark standalone raw-I/Q modulation-recognition
results and classification efficiency. Best and second-best
performance values within each benchmark are bolded and underlined,
respectively. Latency is measured with batch size 1 on one NVIDIA
GeForce RTX 4090.}
\label{tab:supp-standalone-per-benchmark}
\end{table}

BATok performs best on HisarMod2019 and obtains the highest Macro-F1
on \textsc{EMSpec-Sim}, while DNCNet obtains the highest Accuracy on
\textsc{EMSpec-Sim}. With 1.044M parameters, BATok is substantially
smaller than DNCNet while achieving stronger Macro-F1 on both
benchmarks. Its 5.509-ms standalone latency is higher because runtime
depends on sequence-level computation rather than parameter count
alone. BATok repeatedly applies compact shared operators along long
I/Q sequences at multiple resolutions and performs content-aware
resampling; these operations increase computation and memory movement
without introducing many learnable weights. BATok is therefore
small in parameter count and bounded in output length, while
allocating additional tokenizer computation to preserve multi-scale
signal evidence.

The parameter and latency values differ slightly from those in the
aligned-subset validation because the two experiments use different
class heads and input-length distributions.
Table~\ref{tab:supp-modality-comparison} evaluates the aligned
\textsc{HisarMod-Sub}$\cup$\textsc{EMSpec-Sim-Sub} union, whereas
Table~\ref{tab:supp-standalone-per-benchmark} evaluates the complete
standalone benchmarks.

\subsection{Comparison of Token-Budget Strategies}
\label{sec:supp-budget}

We compare five fixed budgets, three length-dependent linear maps, and
the piecewise schedule used by BATok. Table~\ref{tab:supp-budget-sweep}
reports average and maximum token counts, cross-dataset mean
Accuracy/Macro-F1, and measured latency.
The three linear baselines represent increasingly aggressive growth
in token capacity with signal length.

\begin{table}[H]
\centering
\begin{tabular*}{\textwidth}{@{\extracolsep{\fill}}lrrrrr@{}}
\toprule
Strategy &
Avg. $K$ (tokens) & Max $K$ (tokens) &
Avg. Acc. (\%) & Avg. F1 (\%) &
Latency (ms/sample) \\
\midrule
Fixed-16
& 16.00 & 16 & 57.91 & 57.62 & 5.358 \\
Fixed-32
& 32.00 & 32 & 57.68 & 57.53 & \textbf{5.294} \\
Fixed-48
& 48.00 & 48 & 57.97 & 57.60 & 5.300 \\
Fixed-64
& 64.00 & 64 & 58.04 & 57.70 & 5.310 \\
Fixed-128
& 128.00 & 128 & 58.08 & 57.90 & 5.469 \\
\midrule
Linear-low
& 32.50 & 48 & 58.60 & 58.34 & 5.469 \\
Linear-mid
& 48.50 & 64 & \textbf{59.05} & \textbf{58.67} & 5.536 \\
Linear-high
& 81.50 & 128 & \underline{58.70} & 58.48 & 5.456 \\
\midrule
Piecewise $K(T)^\dagger$
& 40.00 & 48 & 58.64 & \underline{58.54} & 5.509 \\
\bottomrule
\end{tabular*}
\caption{Unified performance--budget--latency ablation. Mean Accuracy and Macro-F1 are the unweighted means across
HisarMod2019 and \textsc{EMSpec-Sim}.
Latency is measured in milliseconds per sample
on one NVIDIA GeForce RTX 4090.
$\dagger$ denotes the full BATok configuration. Best and second-best
mean results are bolded and underlined, respectively.}
\label{tab:supp-budget-sweep}
\end{table}

The budget sweep shows that increasing the token count alone does not
consistently improve classification performance. Across the fixed-budget
settings, mean Macro-F1 ranges from 57.53\% to 57.90\%. Linear-mid achieves
the highest mean Accuracy and Macro-F1, at 59.05\% and 58.67\%,
respectively. The piecewise schedule achieves 58.64\% Accuracy and
58.54\% Macro-F1 with 40 tokens on average, using 17.5\% fewer tokens
than Linear-mid at decreases of 0.41 and 0.13 percentage points.
Compared with Fixed-48, it reduces the average token count by 16.7\%
while improving Accuracy and Macro-F1 by 0.67 and 0.94 percentage points.
These results support the piecewise schedule as a compact configuration
that balances token budget and classification performance.

\subsection{Length-Stratified Comparison under Matched Token Budgets}
\label{sec:supp-controlled-length}

We further isolate signal length on \textsc{EMSpec-Sim-VarLen}, a
variable-length subset of \textsc{EMSpec-Sim}. Although samples are stored with
zero-padding to 32,768 points, their natural valid lengths are
$T\in\{1{,}024,2{,}048,4{,}096,8{,}192,16{,}384,32{,}768\}$.
All strategies use the same stratified training/validation split and
the same untouched test split. Equal-length samples are grouped only
to accelerate minibatch processing; no sample is cropped or moved
between splits. The piecewise schedule assigns 32 tokens to the first
three length groups and 48 tokens to the remaining groups, yielding a
sample-weighted test-set average of 39.23 tokens. Fixed $K=39$ is
therefore the matched-budget reference.

\begin{table}[H]
\centering
\setlength{\tabcolsep}{3.0pt}
\begin{tabular}{l r c c c c c c}
\toprule
Strategy & Avg. $K$ (tokens)
& \shortstack{$T=1{,}024$\\(samples)\\Acc./F1 (\%)}
& \shortstack{$T=2{,}048$\\(samples)\\Acc./F1 (\%)}
& \shortstack{$T=4{,}096$\\(samples)\\Acc./F1 (\%)}
& \shortstack{$T=8{,}192$\\(samples)\\Acc./F1 (\%)}
& \shortstack{$T=16{,}384$\\(samples)\\Acc./F1 (\%)}
& \shortstack{$T=32{,}768$\\(samples)\\Acc./F1 (\%)} \\
\midrule
Fixed-32
& 32.00
& \textbf{31.99}/30.56
& 34.69/33.34
& \textbf{38.72}/36.74
& 41.28/39.43
& \textbf{40.39}/38.67
& 40.76/39.24 \\
Fixed-39
& 39.00
& 31.73/\textbf{30.77}
& 36.19/\textbf{35.39}
& 38.10/\textbf{37.21}
& 41.40/39.97
& 38.93/37.96
& 41.18/\textbf{40.70} \\
Fixed-48
& 48.00
& 29.23/28.76
& 34.27/34.04
& 35.05/34.77
& 40.62/\textbf{40.35}
& 39.54/\textbf{39.25}
& 39.86/39.23 \\
Piecewise $K(T)^\dagger$
& 39.23
& 31.84/29.70
& \textbf{36.73}/34.54
& 37.35/35.44
& \textbf{42.11}/38.93
& \textbf{40.39}/37.97
& \textbf{41.42}/38.99 \\
\bottomrule
\end{tabular}
\caption{Controlled within-dataset token-budget comparison.
\textsc{EMSpec-Sim-VarLen} is a variable-length subset of
\textsc{EMSpec-Sim}. Each length column reports Accuracy/Macro-F1 in
percent; Avg. $K$ is weighted by the number of test samples, and bold
marks the best result for each metric at each length.}
\label{tab:supp-controlled-length}
\end{table}

At nearly the same average budget, the piecewise schedule attains
37.67\% overall Accuracy versus 37.31\% for Fixed-39. Its Accuracy is
higher in all three long-input groups, by 0.71, 1.46, and 0.24 points
at $T=8{,}192$, 16,384, and 32,768, respectively. Fixed-39 obtains
higher overall Macro-F1 (36.46\% versus 35.46\%), while the piecewise
schedule consistently improves Accuracy for long observations.
Together, the matched-budget results confirm the intended effect of
length-dependent allocation: it directs additional capacity to long
signals without increasing the average token budget.

\subsection{Component-Level Ablation Analysis}
\label{sec:supp-components}

Table~\ref{tab:supp-component-ablations} reports per-dataset
classification results for input features, temporal scales, and
resampling bias. The temporal-scale rows compare the strongest
single- and two-branch configurations with the full three-branch
model. In each group, $\dagger$ marks the full BATok configuration.

\begin{table}[H]
\centering
\resizebox{\textwidth}{!}{%
\begin{tabular}{ll c c r r}
\toprule
Group & Variant
& \shortstack{HisarMod2019\\Acc./F1 (\%)}
& \shortstack{\textsc{EMSpec-Sim}\\Acc./F1 (\%)}
& Avg. Acc. (\%) & Avg. F1 (\%) \\
\midrule
Input features
& I/Q only & 74.07/73.97 & 38.66/38.62
& 56.37 & 56.30 \\
& I/Q+$A$+$P$+$\Delta\phi$ $\dagger$
& 77.53/77.53 & 39.75/39.54
& \textbf{58.64} & \textbf{58.54} \\
\midrule
Temporal scales
& Small & 68.89/68.88 & 40.51/40.18
& 54.70 & 54.53 \\
& Small+mid & 72.64/72.53 & 39.48/39.63
& 56.06 & 56.08 \\
& Small+mid+large $\dagger$
& 77.53/77.53 & 39.75/39.54
& \textbf{58.64} & \textbf{58.54} \\
\midrule
Resampling bias
& Energy prior only & 76.99/77.07 & 36.95/37.36
& 56.97 & 57.22 \\
& Learned relevance only & 76.88/76.83 & 36.71/36.06
& 56.80 & 56.45 \\
& Energy+learned relevance $\dagger$
& 77.53/77.53 & 39.75/39.54
& \textbf{58.64} & \textbf{58.54} \\
\bottomrule
\end{tabular}}
\caption{Per-dataset component ablations for standalone BATok
classification. Dataset columns report Accuracy/Macro-F1 in percent;
Avg. columns are arithmetic means across HisarMod2019 and
\textsc{EMSpec-Sim}. $\dagger$
denotes the full BATok configuration, and bold marks the best
cross-dataset mean within each ablation group.}
\label{tab:supp-component-ablations}
\end{table}

The three resampling-bias variants retain the same classifier
architecture and each has 1.044M parameters; the ablation changes the
evidence score used during resampling.

The full configuration achieves the strongest two-benchmark mean in
all three ablation groups. Signal-derived amplitude, power, and phase
difference features improve mean Accuracy/Macro-F1 by 2.27/2.24
points over I/Q alone; the third temporal branch adds 2.58/2.46
points over two branches; and combining energy with learned relevance
improves by 1.67/1.32 points over the stronger single-bias variant.
These consistent gains validate enhanced signal features,
multi-resolution encoding, and complementary resampling cues as the
three key components of BATok.

\section{Visualization Analysis}
\label{sec:supp-qualitative}

The main paper provides one detection case and one grounding case for
the Qwen3-VL-2B backbone. Here, we expand the comparison to both
Qwen3-VL-2B and LLaVA-1.5-7B and cover narrow-band targets, crowded
multi-class scenes, and class-, frequency-, time-, and
overlap-conditioned queries. We also retain a case in which all
models succeed (Figure~\ref{fig:supp-grounding-b}) instead of
selecting only examples favorable to BATok. These examples are
intended to expose representative output behavior; performance claims
are based on the full-test-set metrics reported in the main paper.
The visualizations are read along two complementary axes: whether the
image pathway preserves the scene geometry needed for localization,
and whether the raw-I/Q pathway resolves signal semantics that remain
ambiguous in a waterfall image. Accordingly, each group shows the
ground truth and paired predictions from the same prompt and
observation, with and without BATok.

\subsection{Structured Detection}
\label{sec:supp-detection-examples}

\begin{figure}[!htb]
\centering
\setlength{\tabcolsep}{1pt}
\begin{tabular}{@{}ccccc@{}}
\scriptsize GT & \scriptsize LLaVA & \scriptsize LLaVA+BATok &
\scriptsize Qwen3-VL & \scriptsize Qwen3-VL+BATok \\
\includegraphics[width=0.192\linewidth]{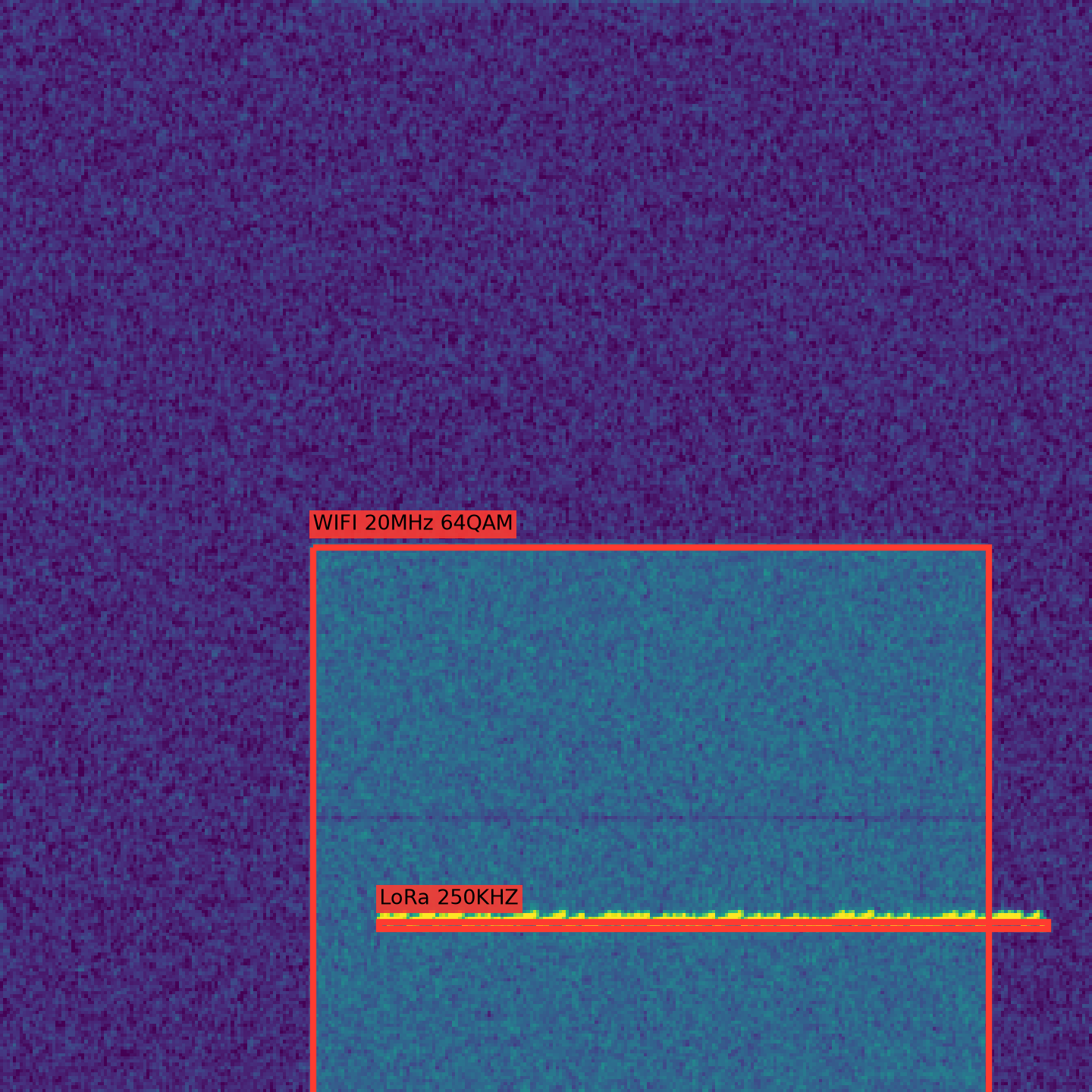} &
\includegraphics[width=0.192\linewidth]{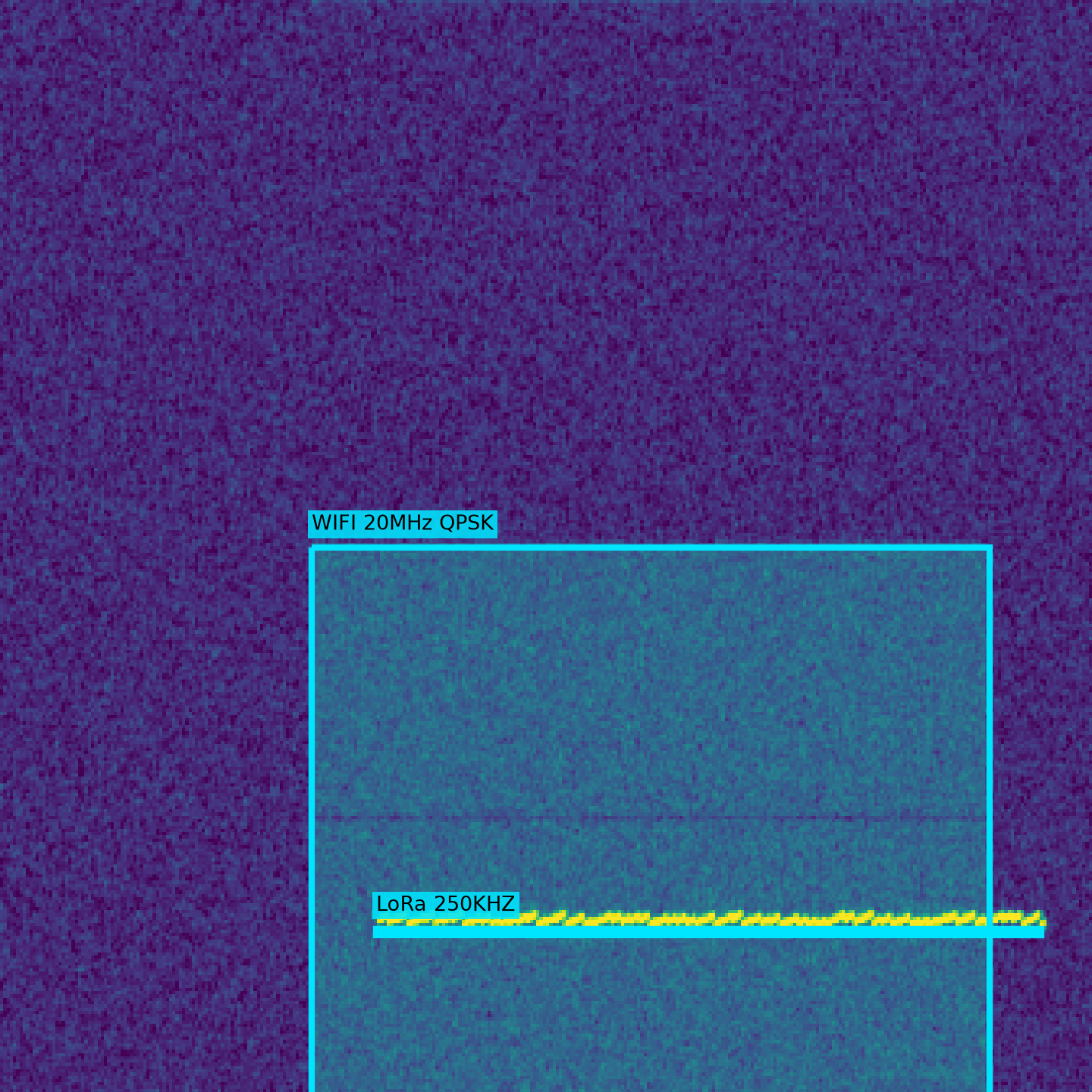} &
\includegraphics[width=0.192\linewidth]{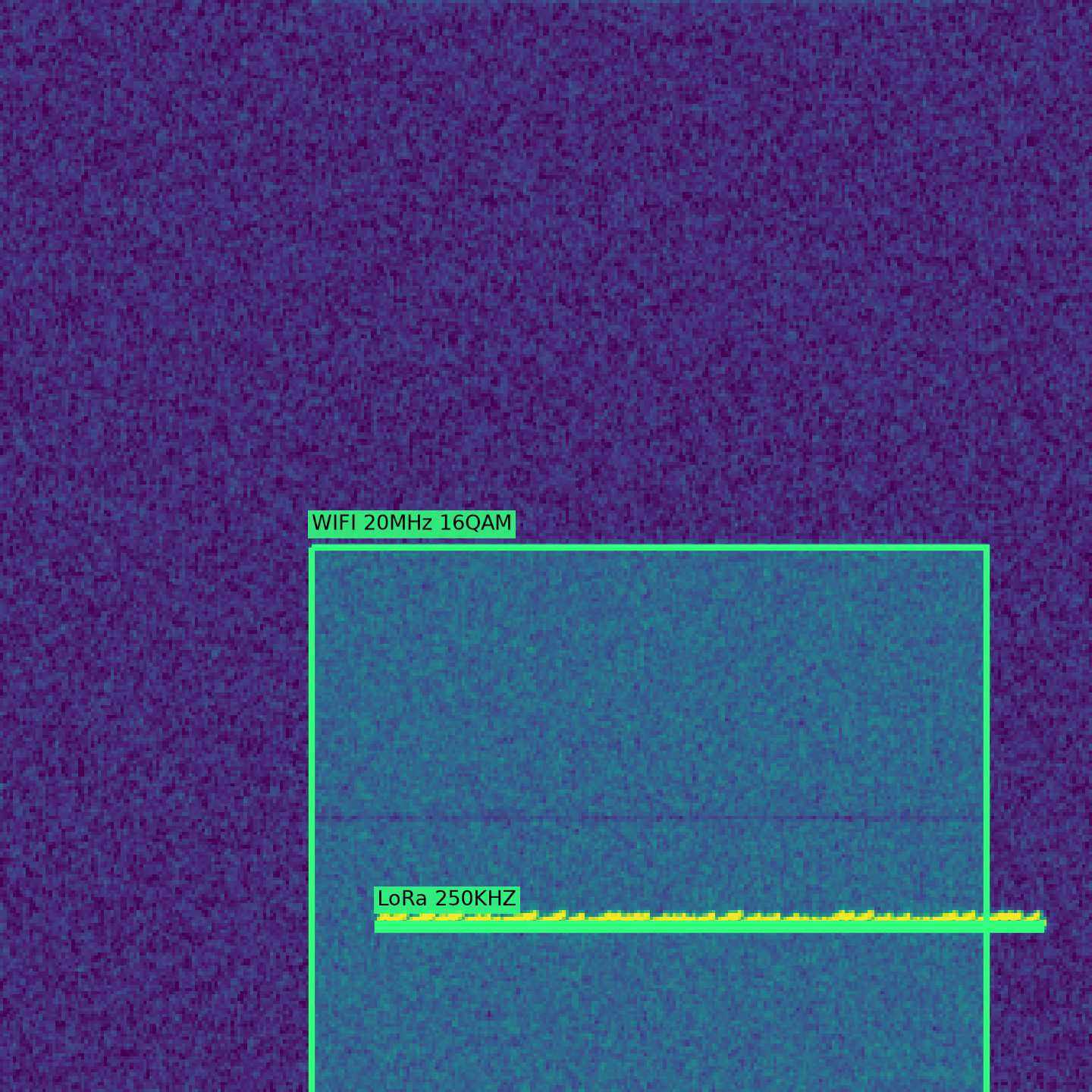} &
\includegraphics[width=0.192\linewidth]{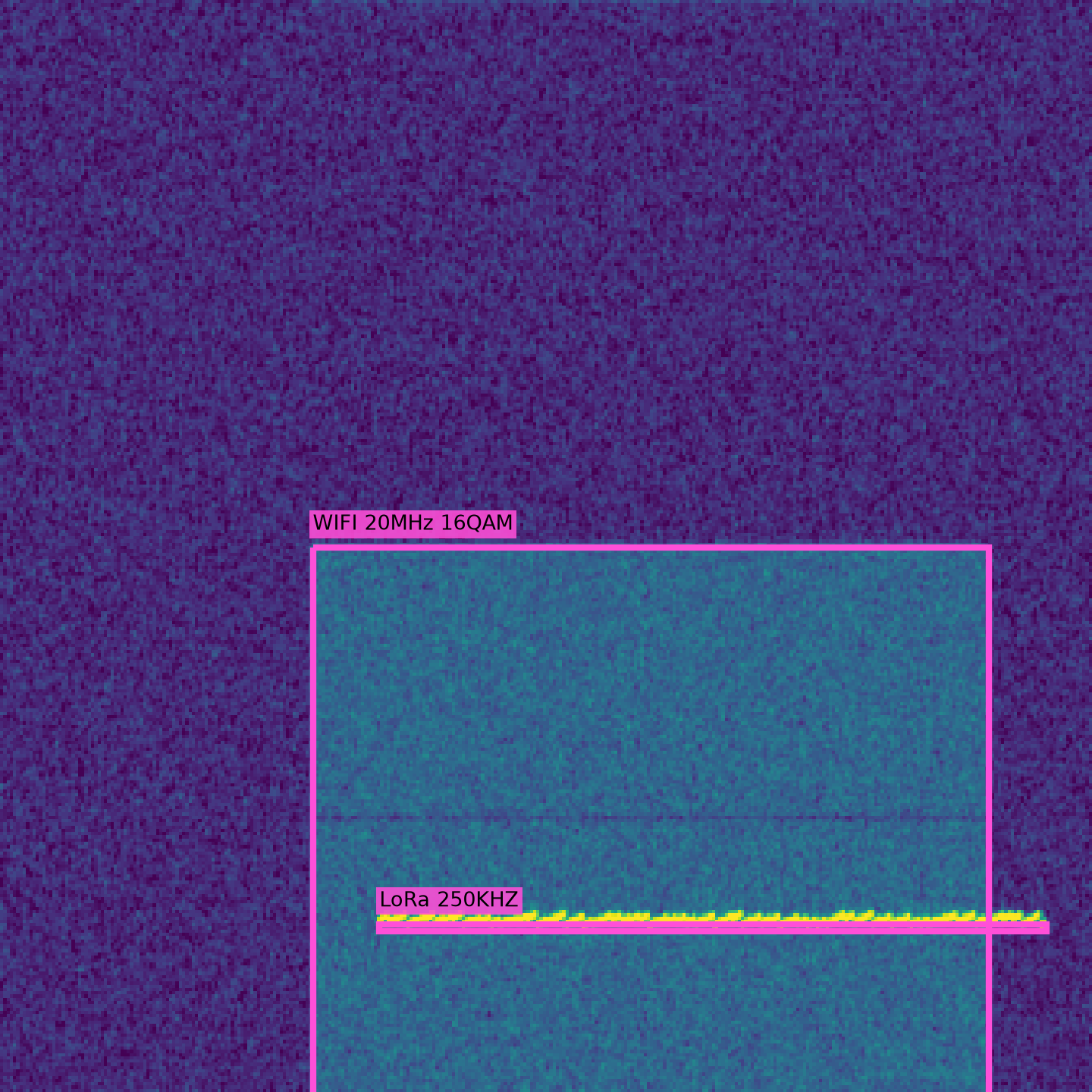} &
\includegraphics[width=0.192\linewidth]{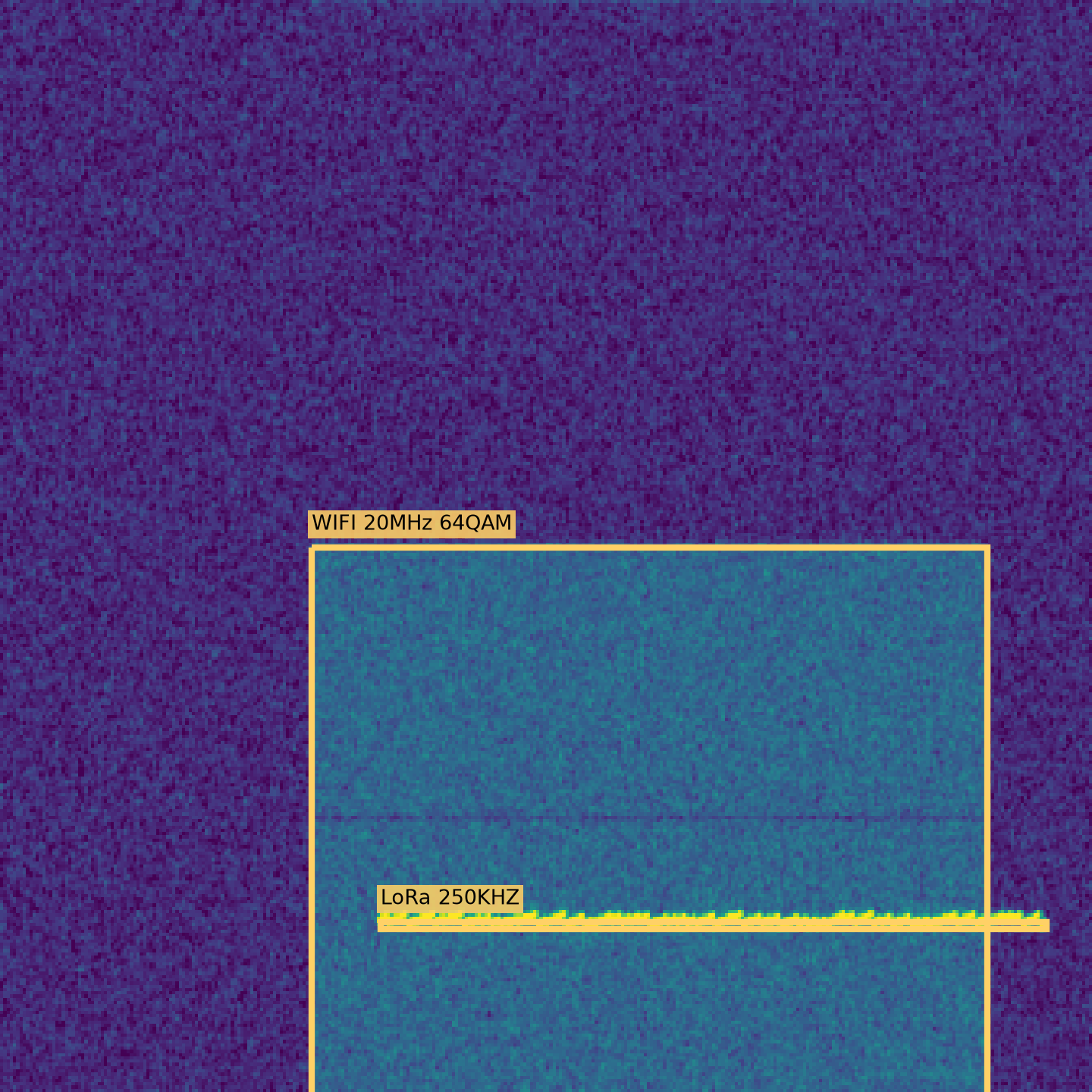}
\end{tabular}
\caption{Detection of a broadband transmission and a narrow-band
target. \textbf{Question:} ``Detect all occupied signal regions and
pair each bounding box with the matching class label.''
\textbf{Ground truth:} $[285,500,904,1000]$ (WiFi 20\,MHz 64QAM) and
$[346,843,958,849]$ (LoRa 250\,kHz). Coordinates are normalized to
$[0,1000]$; the GT panel uses red boxes.}
\label{fig:supp-detection-a}
\end{figure}

\begin{figure}[!htb]
\centering
\setlength{\tabcolsep}{1pt}
\begin{tabular}{@{}ccccc@{}}
\scriptsize GT & \scriptsize LLaVA & \scriptsize LLaVA+BATok &
\scriptsize Qwen3-VL & \scriptsize Qwen3-VL+BATok \\
\includegraphics[width=0.192\linewidth]{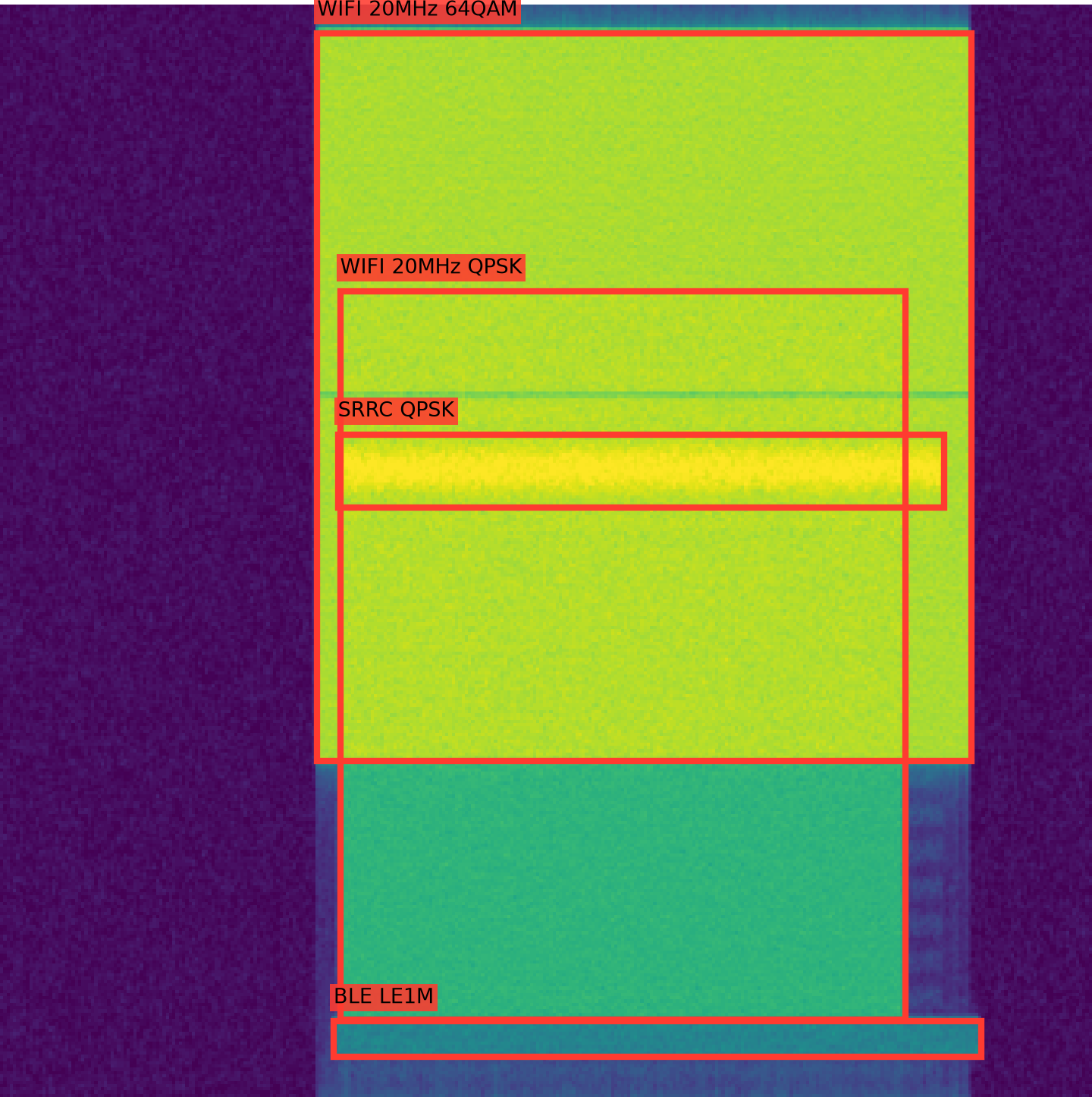} &
\includegraphics[width=0.192\linewidth]{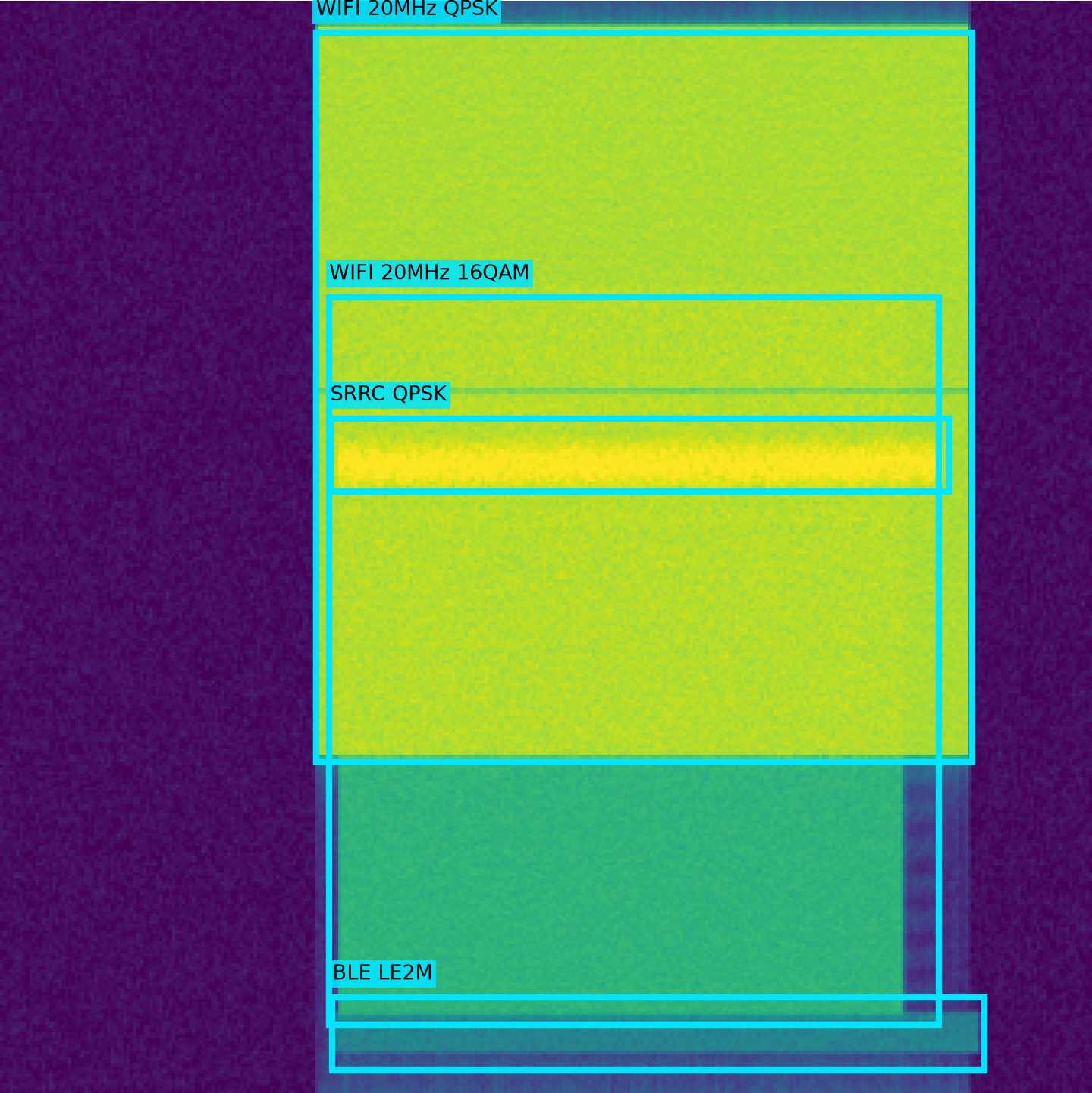} &
\includegraphics[width=0.192\linewidth]{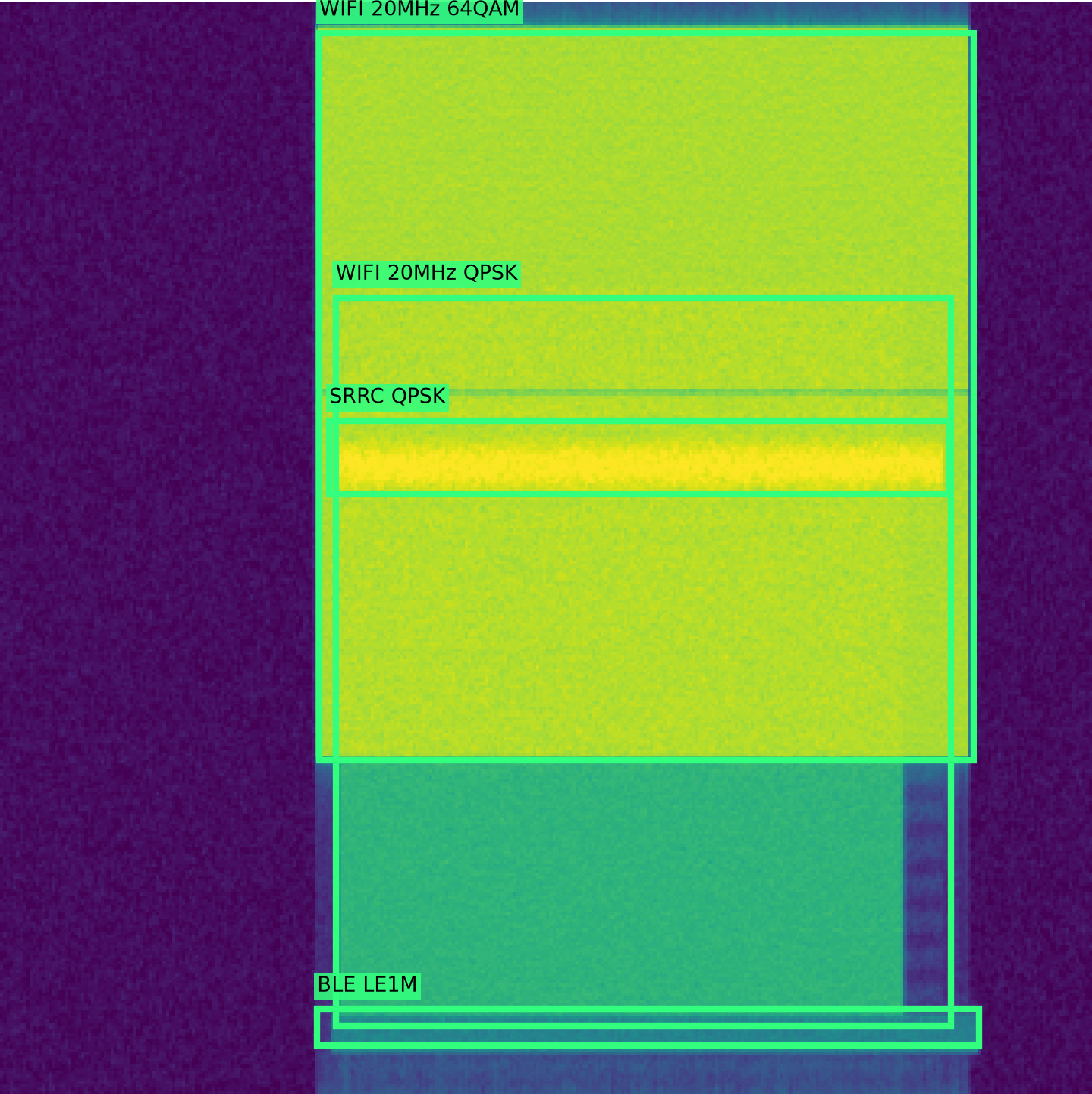} &
\includegraphics[width=0.192\linewidth]{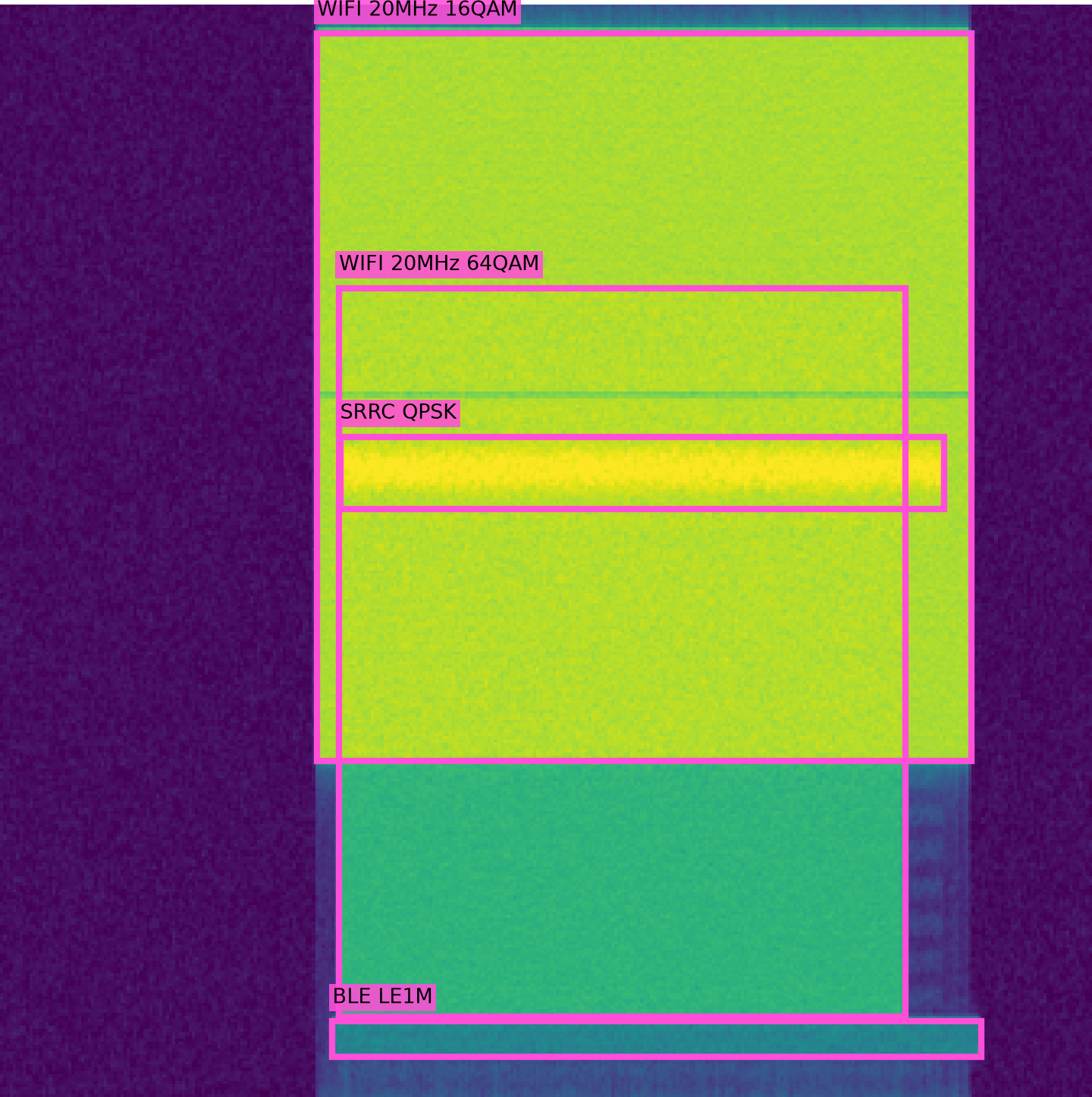} &
\includegraphics[width=0.192\linewidth]{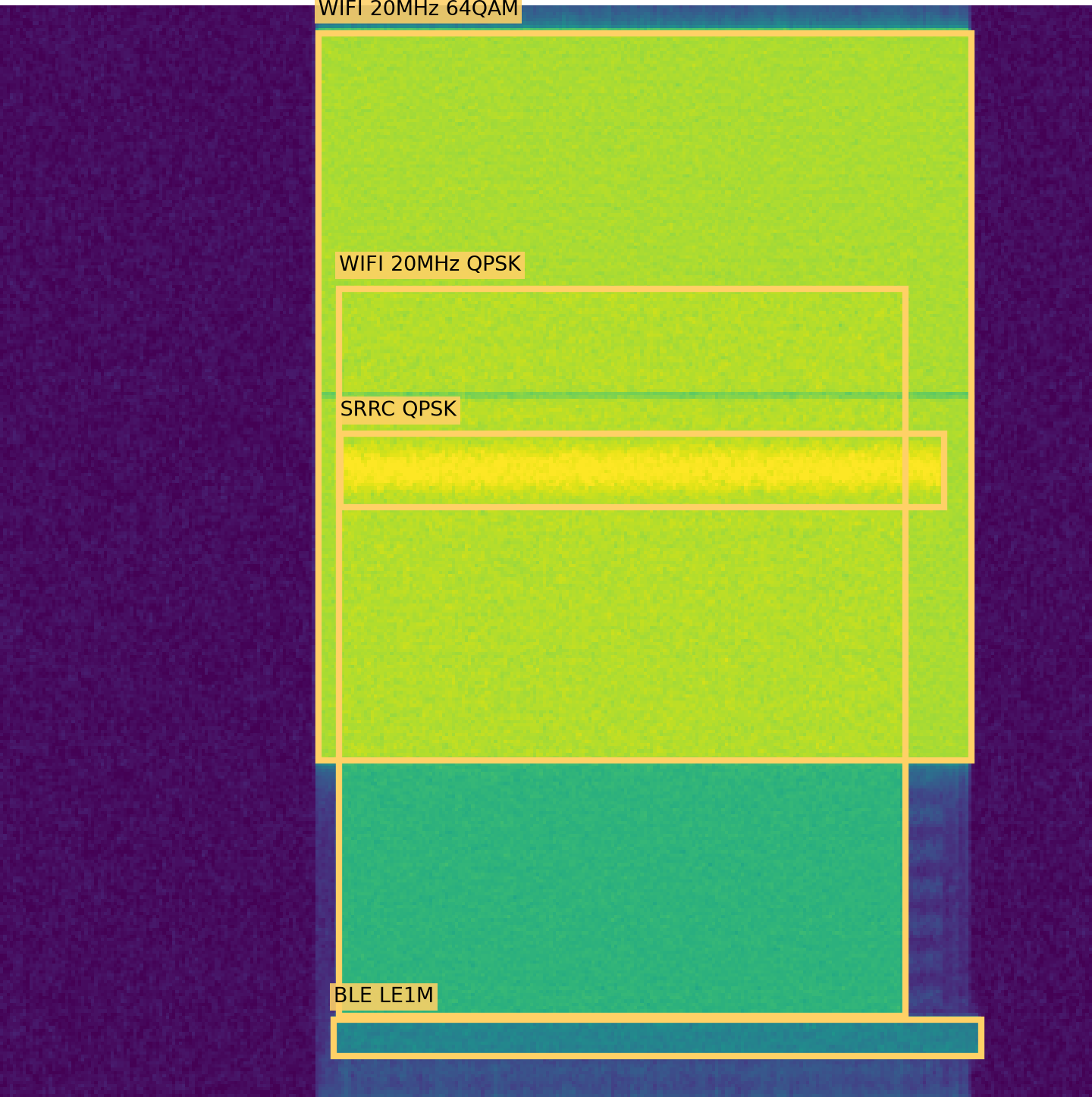}
\end{tabular}
\caption{Detection in a crowded heterogeneous scene.
\textbf{Question:} ``Perform exhaustive detection and respond only
with the requested tags.'' \textbf{Ground truth:}
$[289,25,888,691]$ (WiFi 20\,MHz 64QAM),
$[304,929,897,962]$ (BLE LE1M), $[308,392,863,459]$ (SRRC QPSK), and
$[310,261,828,928]$ (WiFi 20\,MHz QPSK). Coordinates are normalized
to $[0,1000]$; the GT panel uses red boxes.}
\label{fig:supp-detection-b}
\end{figure}

Figures~\ref{fig:supp-detection-a} and
\ref{fig:supp-detection-b} show two complementary challenges. The
first combines a broadband WiFi transmission with an extremely narrow
LoRa signal, while the second contains four transmissions from three
signal families with substantial time--frequency overlap. Each model
is evaluated on the same observation and prompt as its paired
counterpart.

Across both scenes, the image pathway generally recovers the
time--frequency extent of broadband, narrow-band, and overlapping
transmissions, while the remaining errors are concentrated in their
class assignments. Adding BATok preserves these localized regions and
provides additional evidence for distinguishing modulation variants.
This effect is particularly clear for Qwen3-VL-2B: in both figures,
the image-only model labels the WiFi 20\,MHz 64QAM transmission as
16QAM, whereas Qwen3-VL-2B+BATok recovers the correct class without
degrading the corresponding box. The examples therefore illustrate
the semantic contribution of raw I/Q observations alongside the
geometric evidence supplied by the waterfall image.

\subsection{Language-Conditioned Grounding}
\label{sec:supp-grounding-examples}

\begin{figure}[H]
\centering
\setlength{\tabcolsep}{1pt}
\begin{tabular}{@{}ccccc@{}}
\scriptsize GT & \scriptsize LLaVA & \scriptsize LLaVA+BATok &
\scriptsize Qwen3-VL & \scriptsize Qwen3-VL+BATok \\
\includegraphics[width=0.192\linewidth]{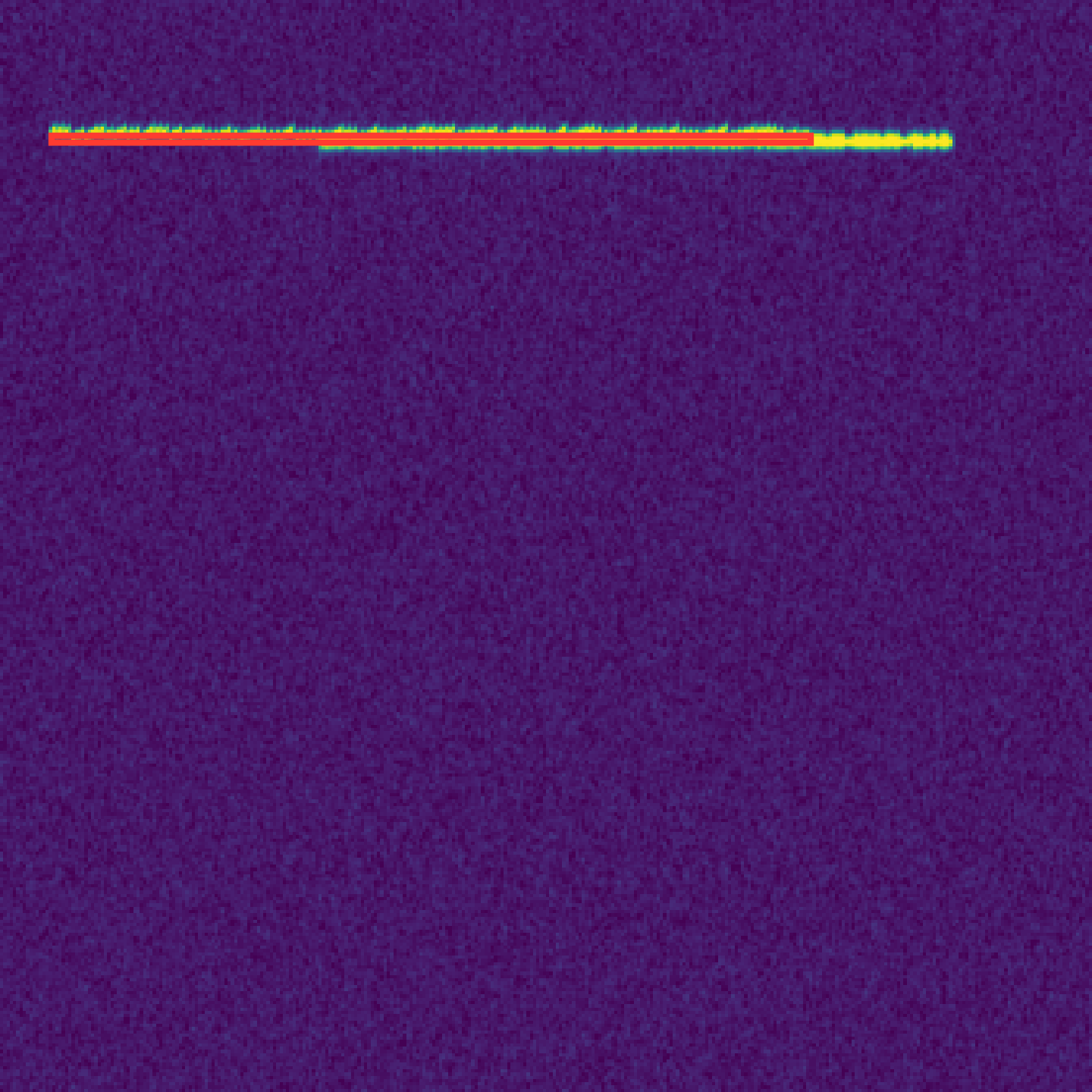} &
\includegraphics[width=0.192\linewidth]{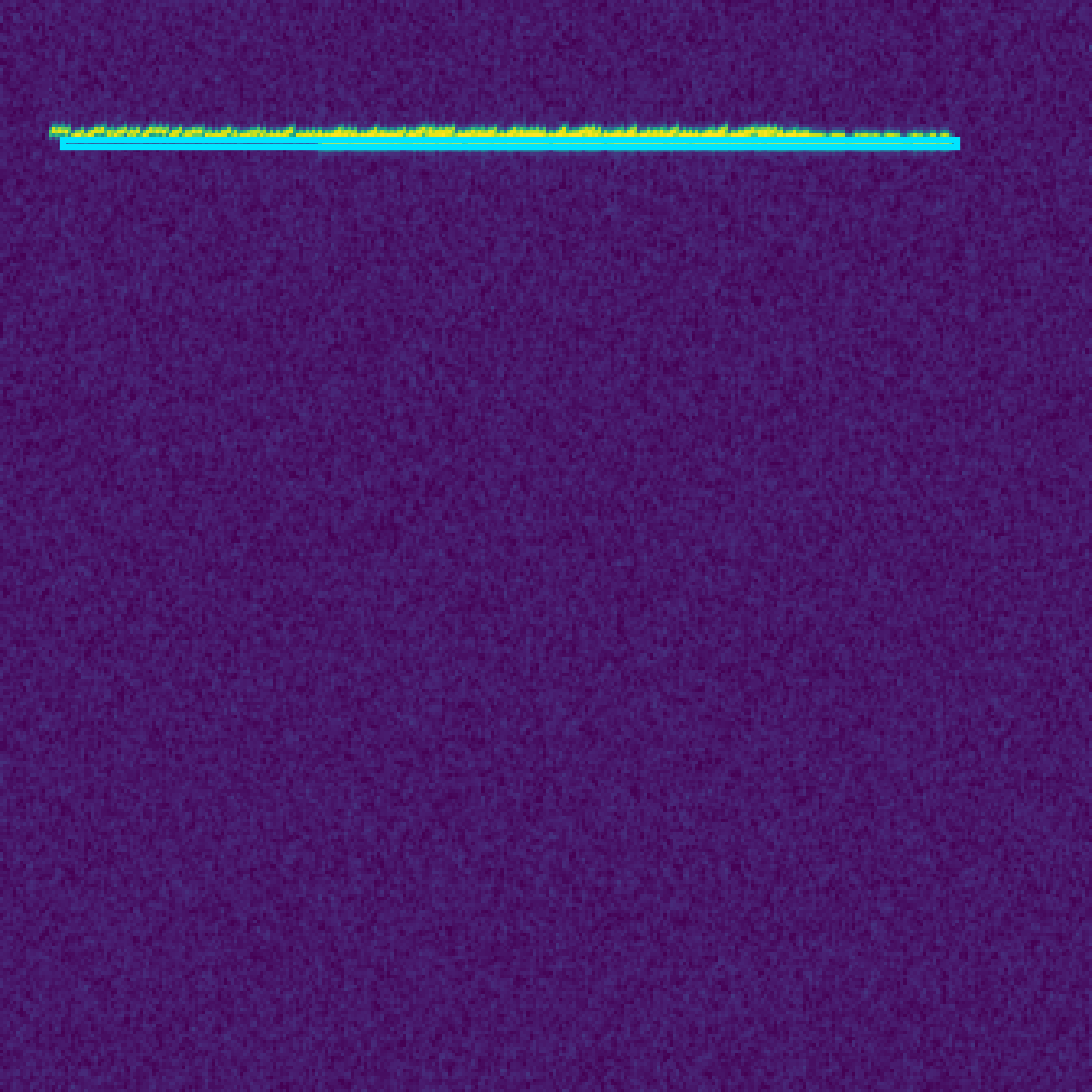} &
\includegraphics[width=0.192\linewidth]{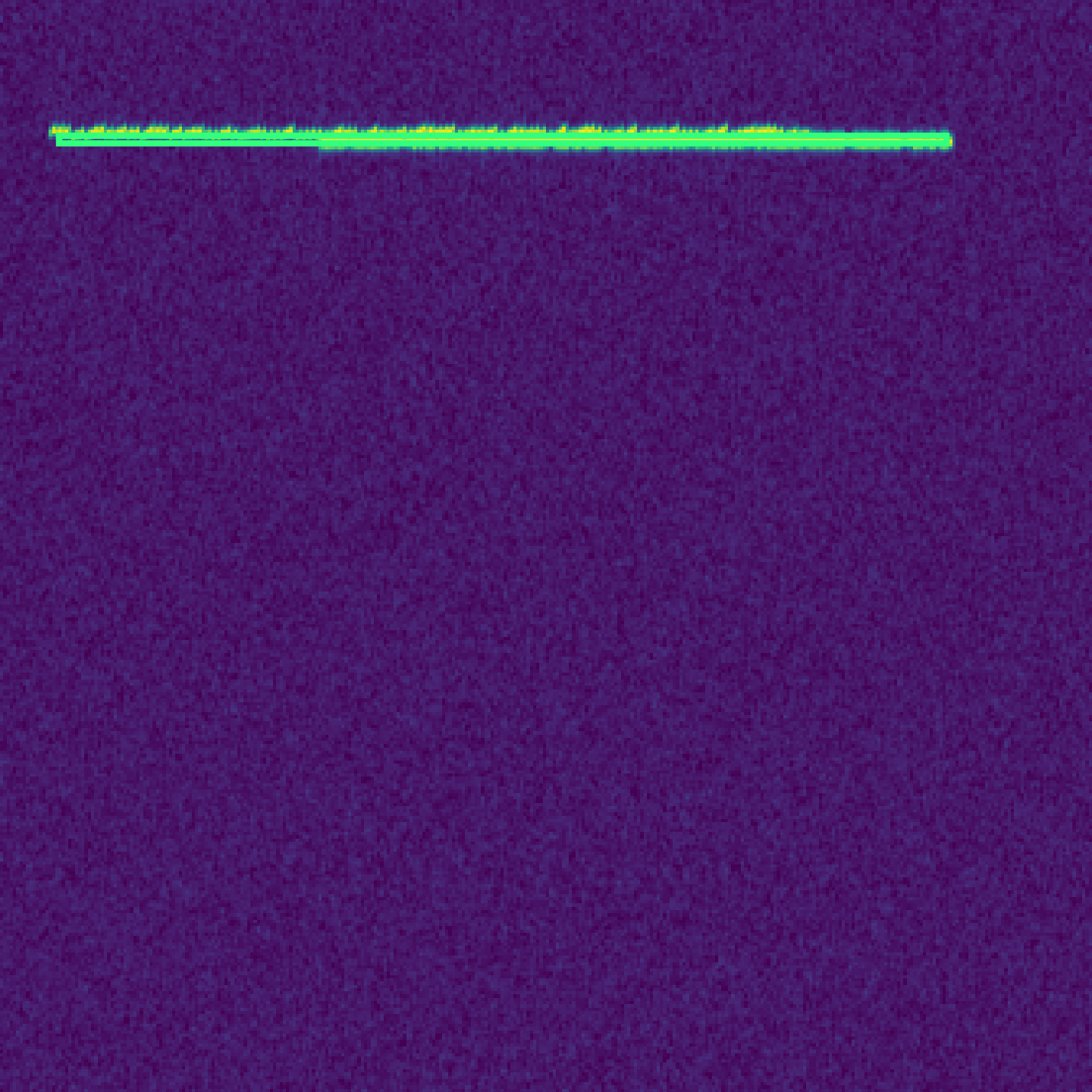} &
\includegraphics[width=0.192\linewidth]{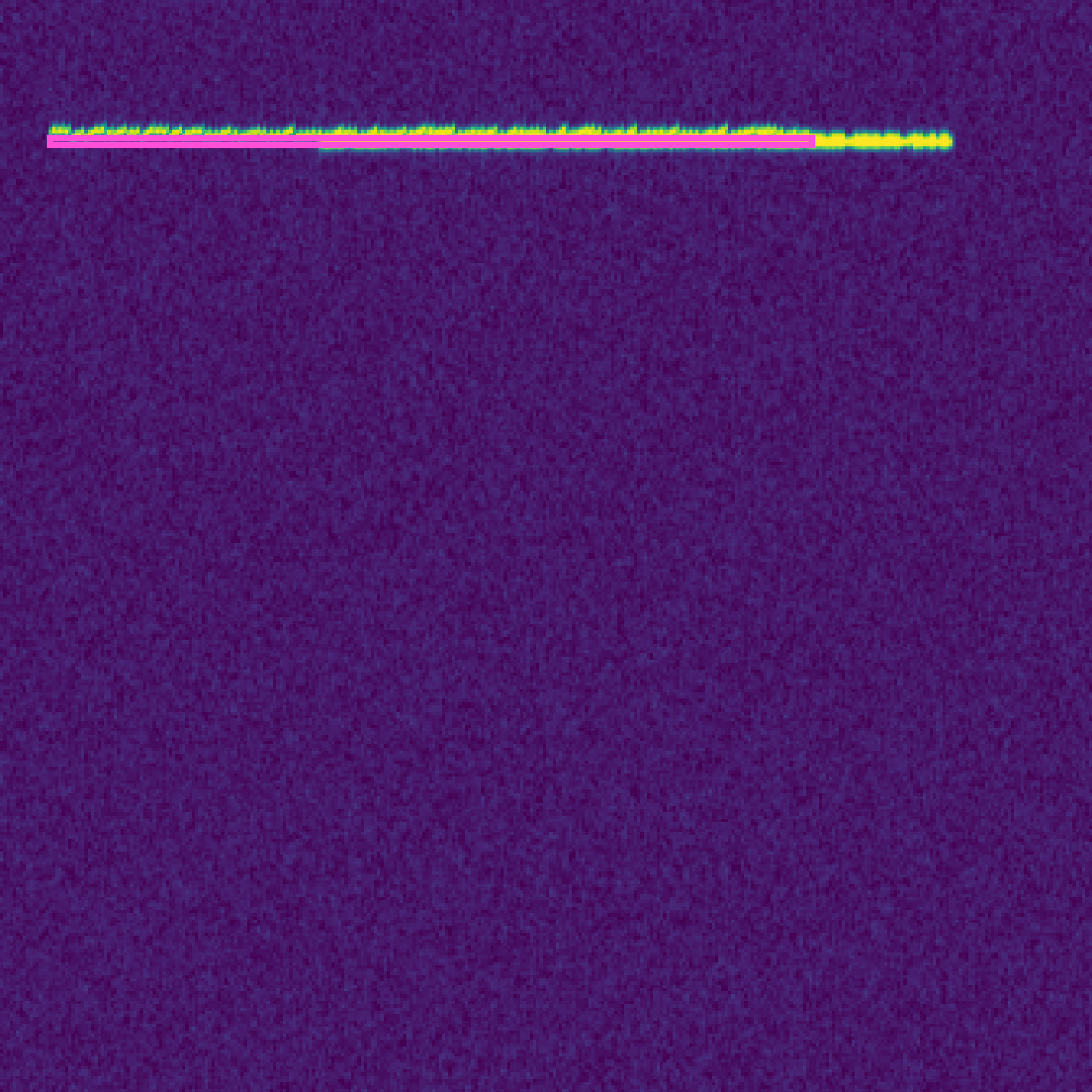} &
\includegraphics[width=0.192\linewidth]{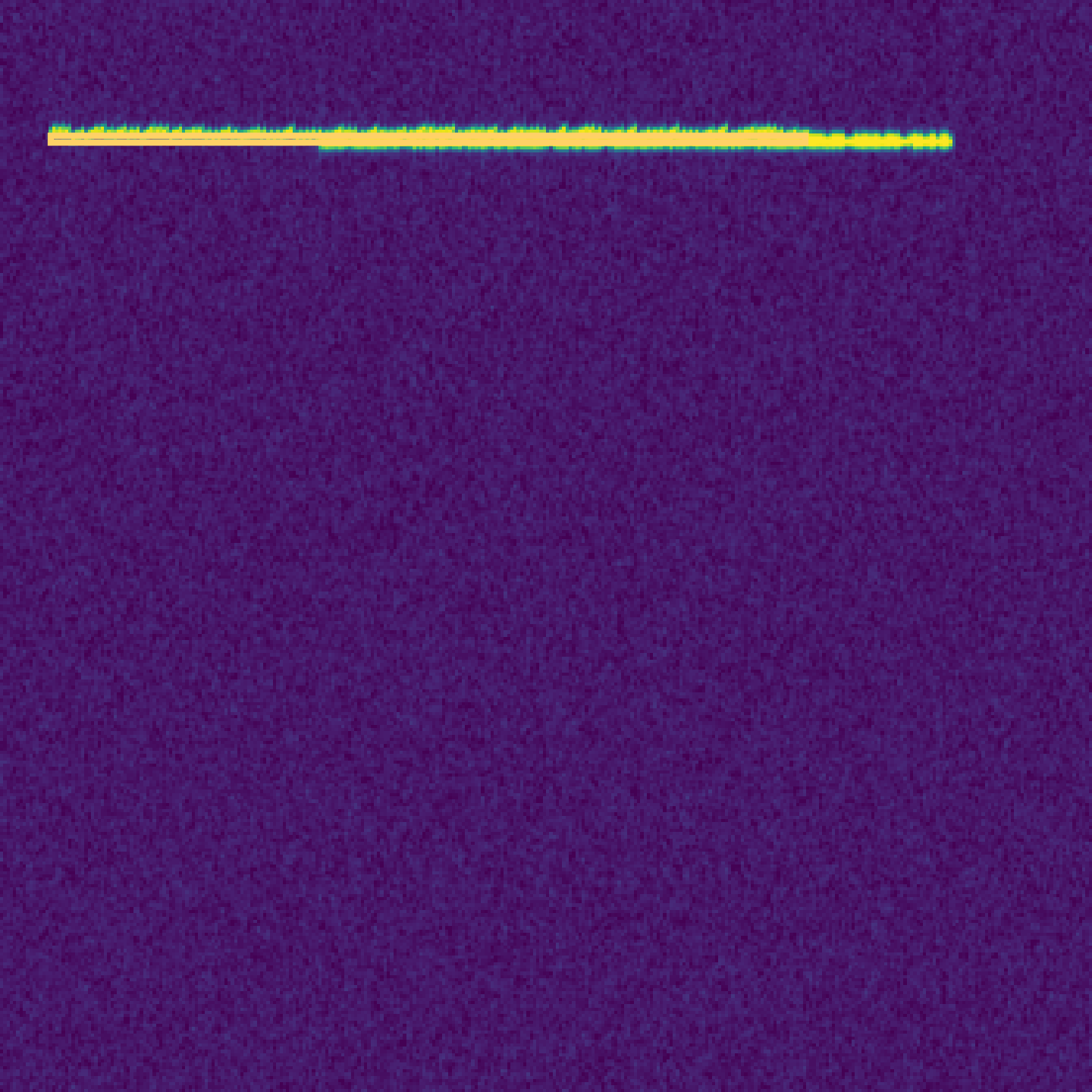} \\
\multicolumn{5}{c}{\scriptsize (a) Class-conditioned query} \\[2pt]
\includegraphics[width=0.192\linewidth]{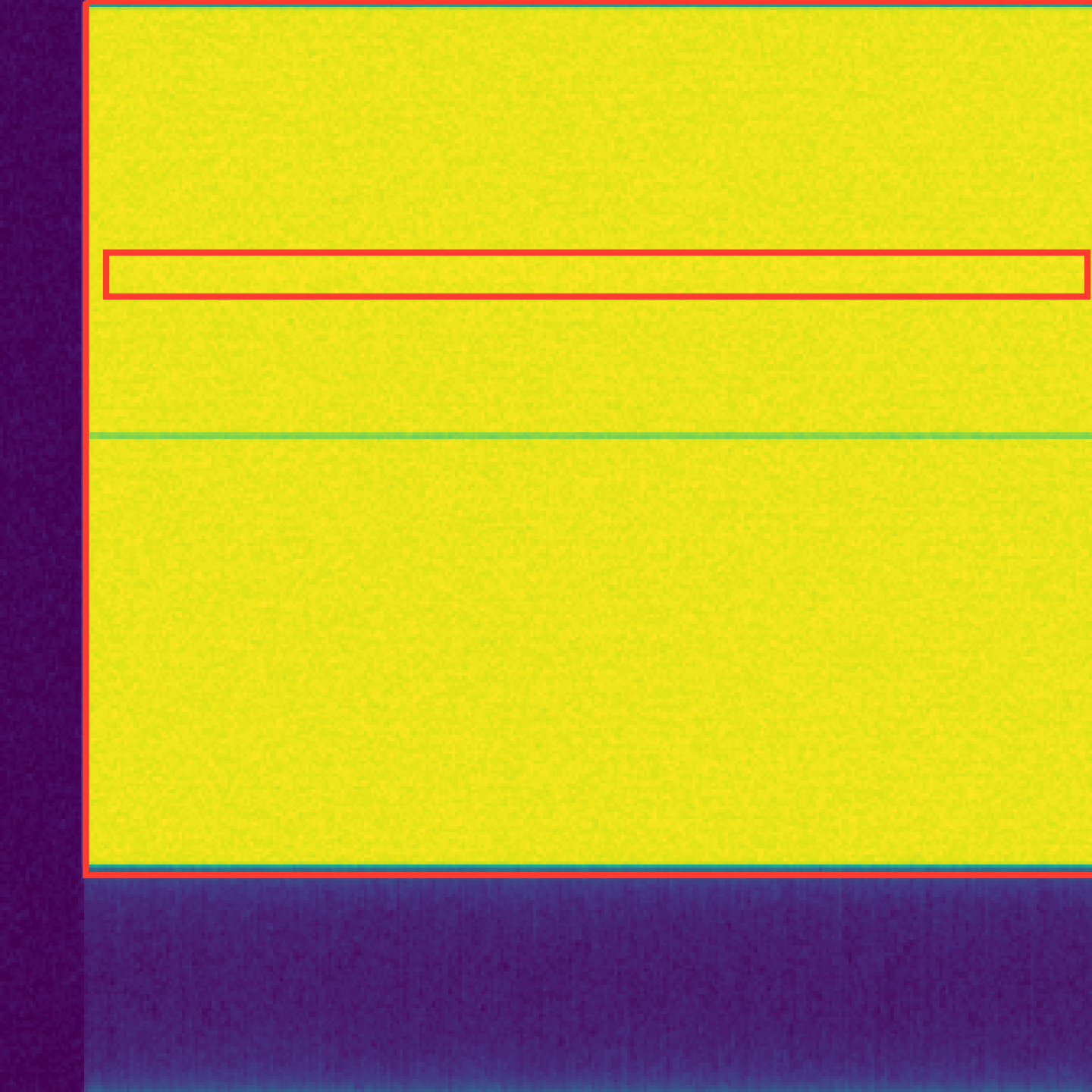} &
\includegraphics[width=0.192\linewidth]{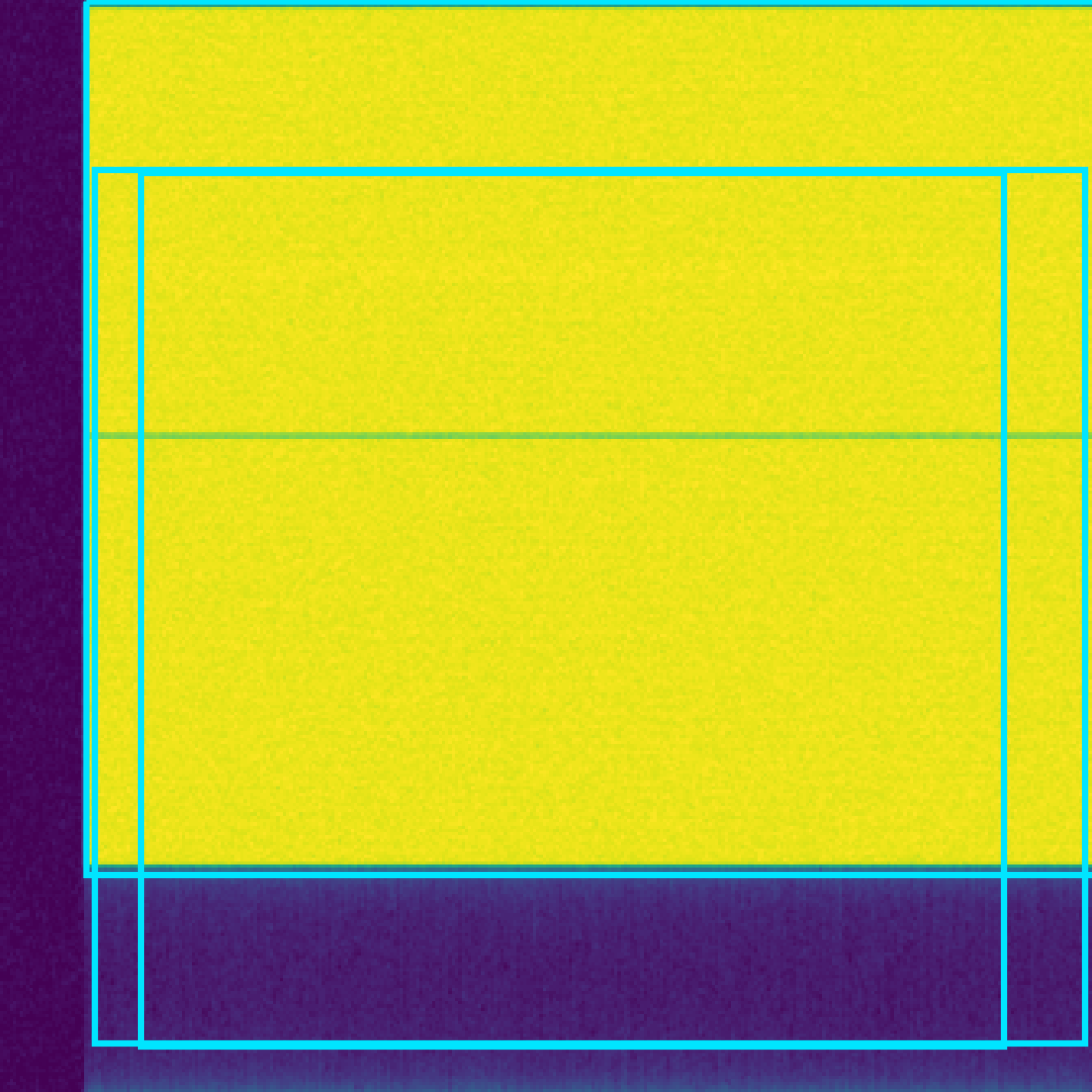} &
\includegraphics[width=0.192\linewidth]{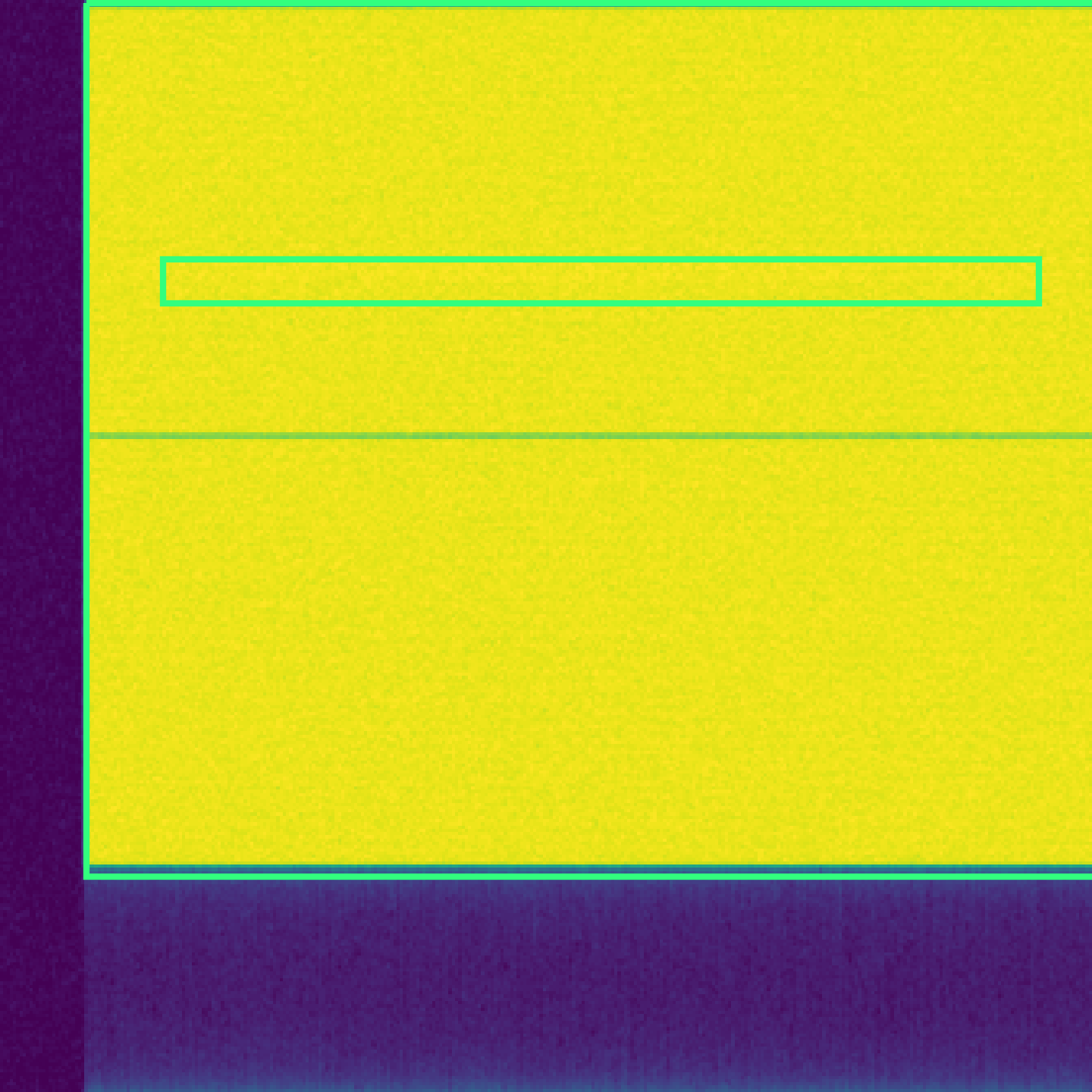} &
\includegraphics[width=0.192\linewidth]{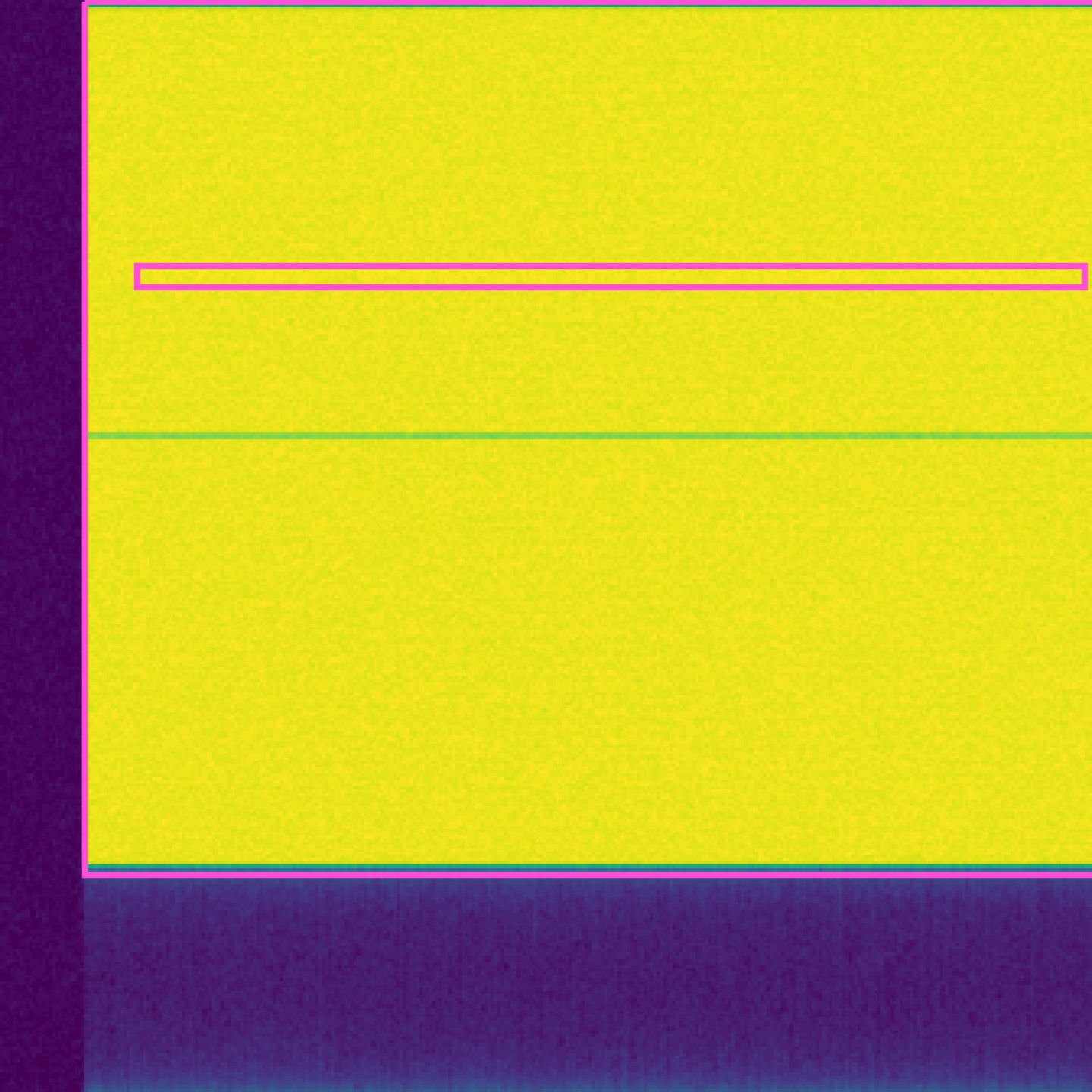} &
\includegraphics[width=0.192\linewidth]{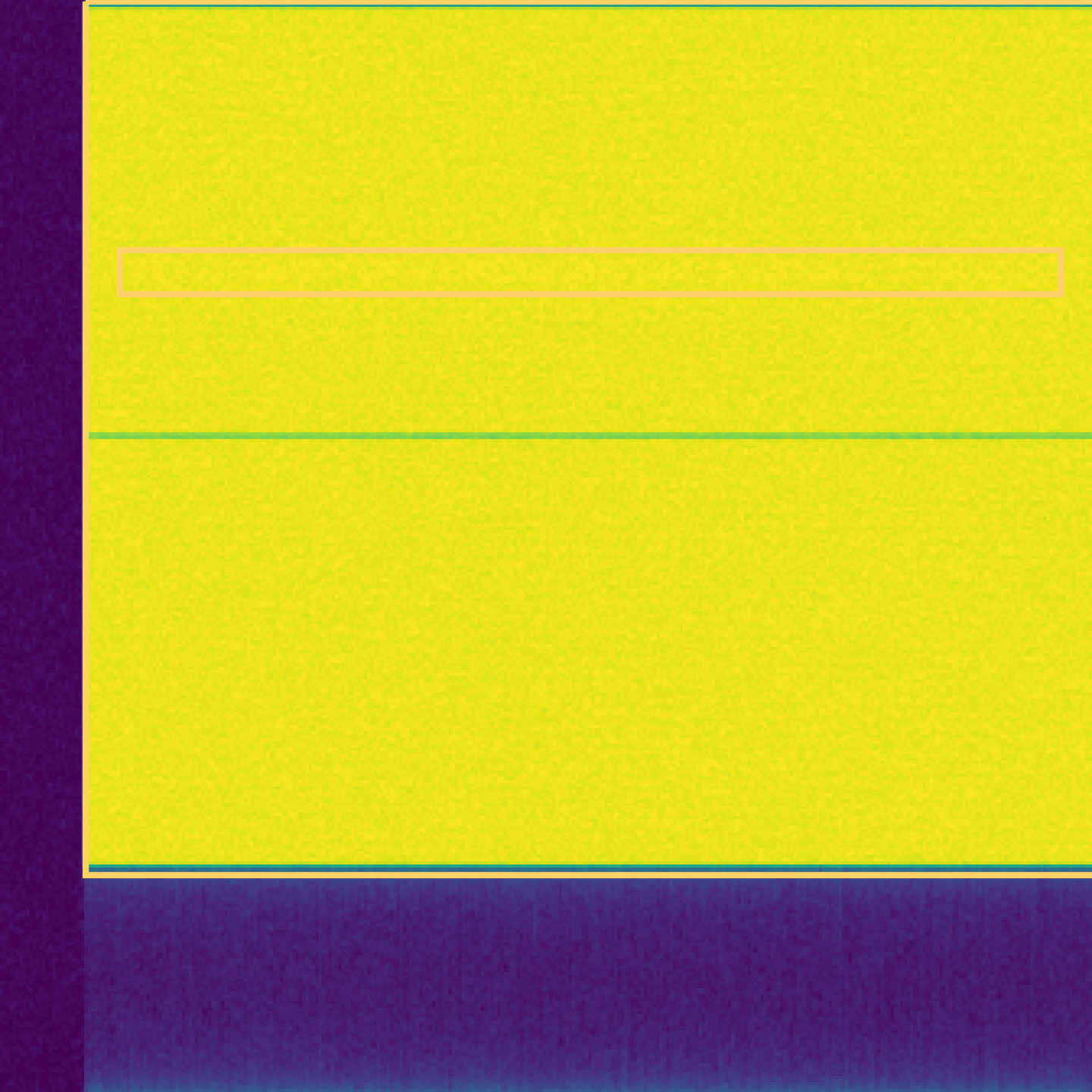} \\
\multicolumn{5}{c}{\scriptsize (b) Frequency-conditioned query}
\end{tabular}
\caption{Grounding under class and frequency constraints.
\textbf{(a) Question:} ``Which regions correspond to the LoRa
250\,kHz class?'' \textbf{GT:} $[46,123,741,129]$.
\textbf{(b) Question:} ``Find signal regions that overlap the
frequency band 2455.015597--2457.837200\,MHz.''
\textbf{GT:} $[77,0,1000,800]$ and $[96,230,994,270]$.
Coordinates are normalized to $[0,1000]$; GT panels use red boxes.}
\label{fig:supp-grounding-a}
\end{figure}

\begin{figure}[H]
\centering
\setlength{\tabcolsep}{1pt}
\begin{tabular}{@{}ccccc@{}}
\scriptsize GT & \scriptsize LLaVA & \scriptsize LLaVA+BATok &
\scriptsize Qwen3-VL & \scriptsize Qwen3-VL+BATok \\
\includegraphics[width=0.192\linewidth]{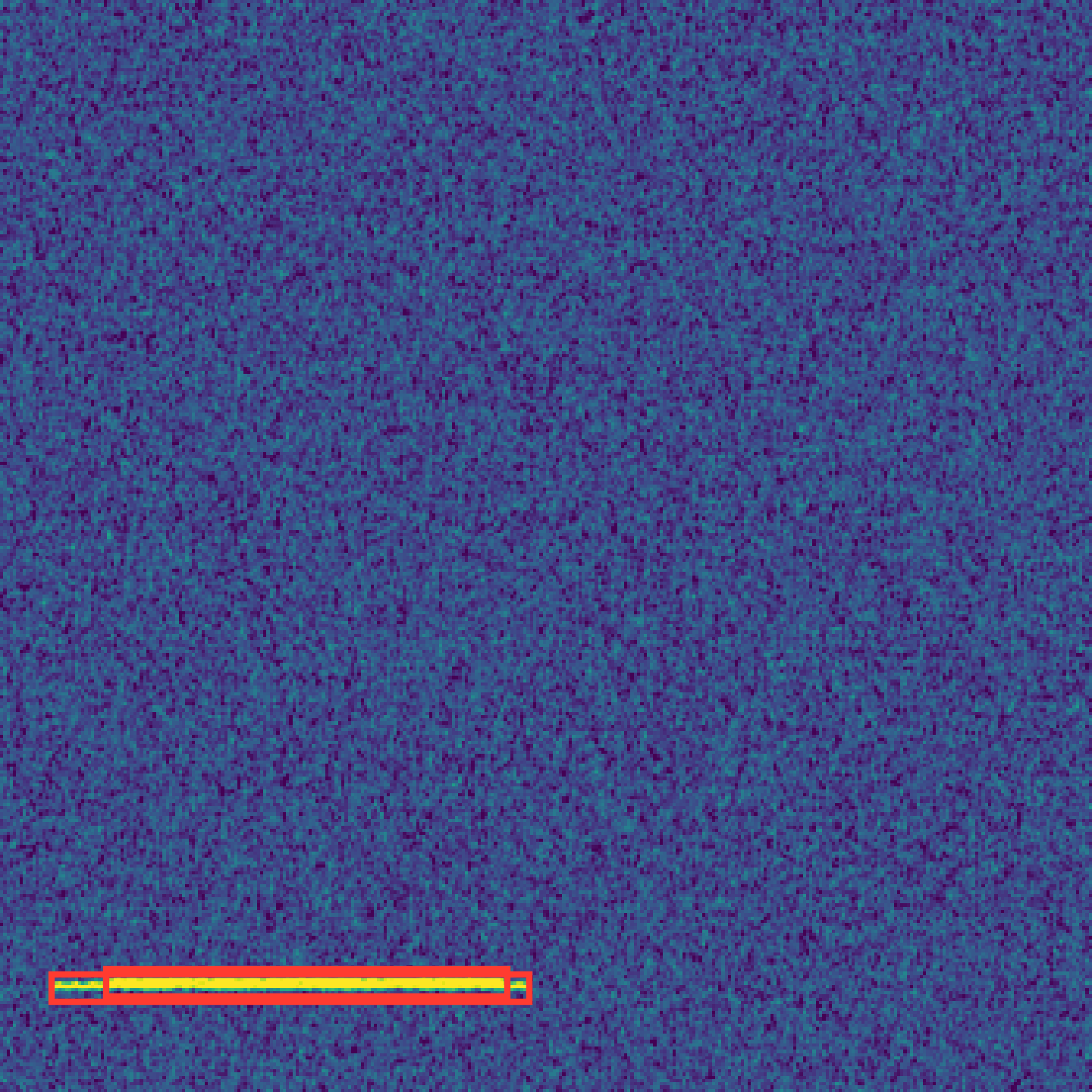} &
\includegraphics[width=0.192\linewidth]{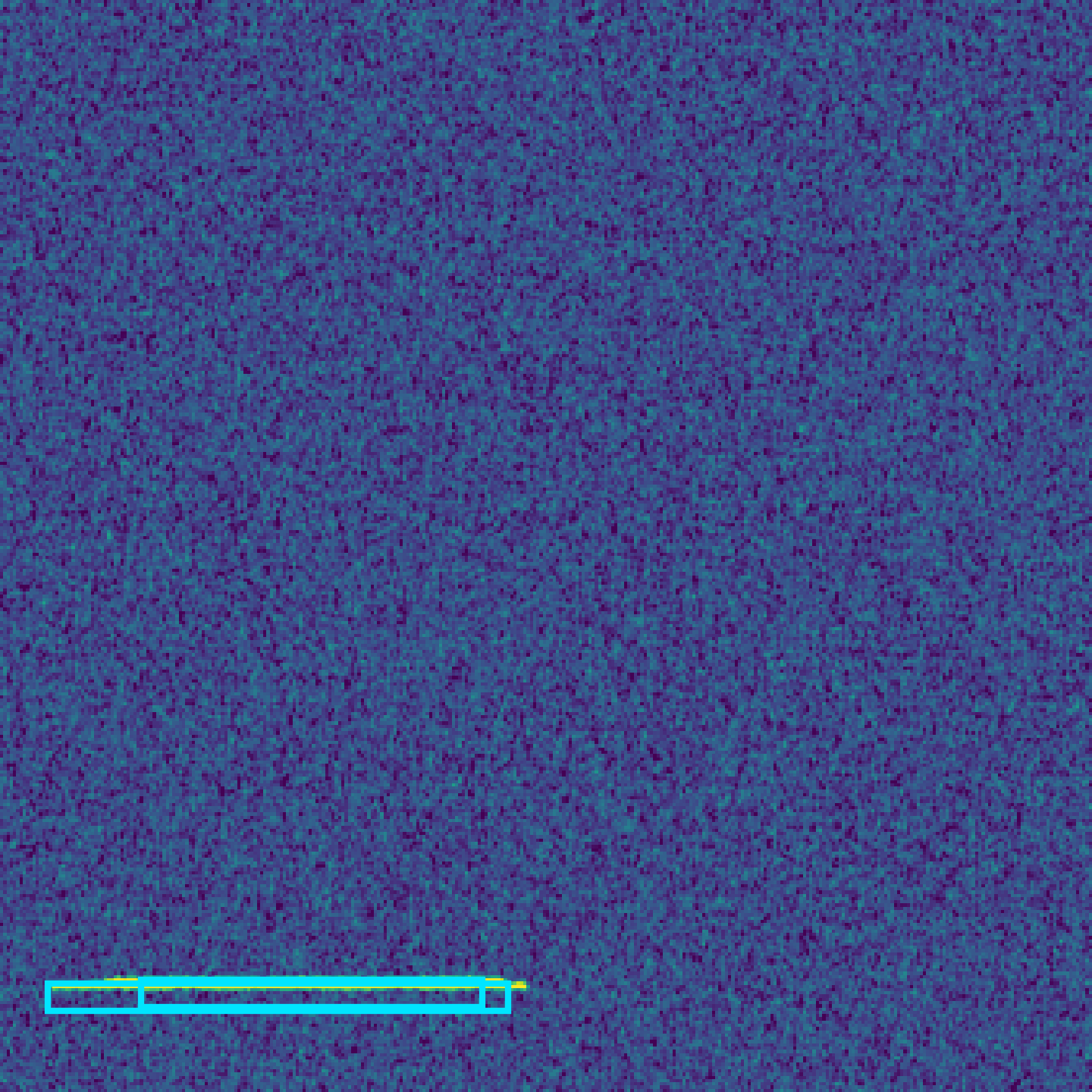} &
\includegraphics[width=0.192\linewidth]{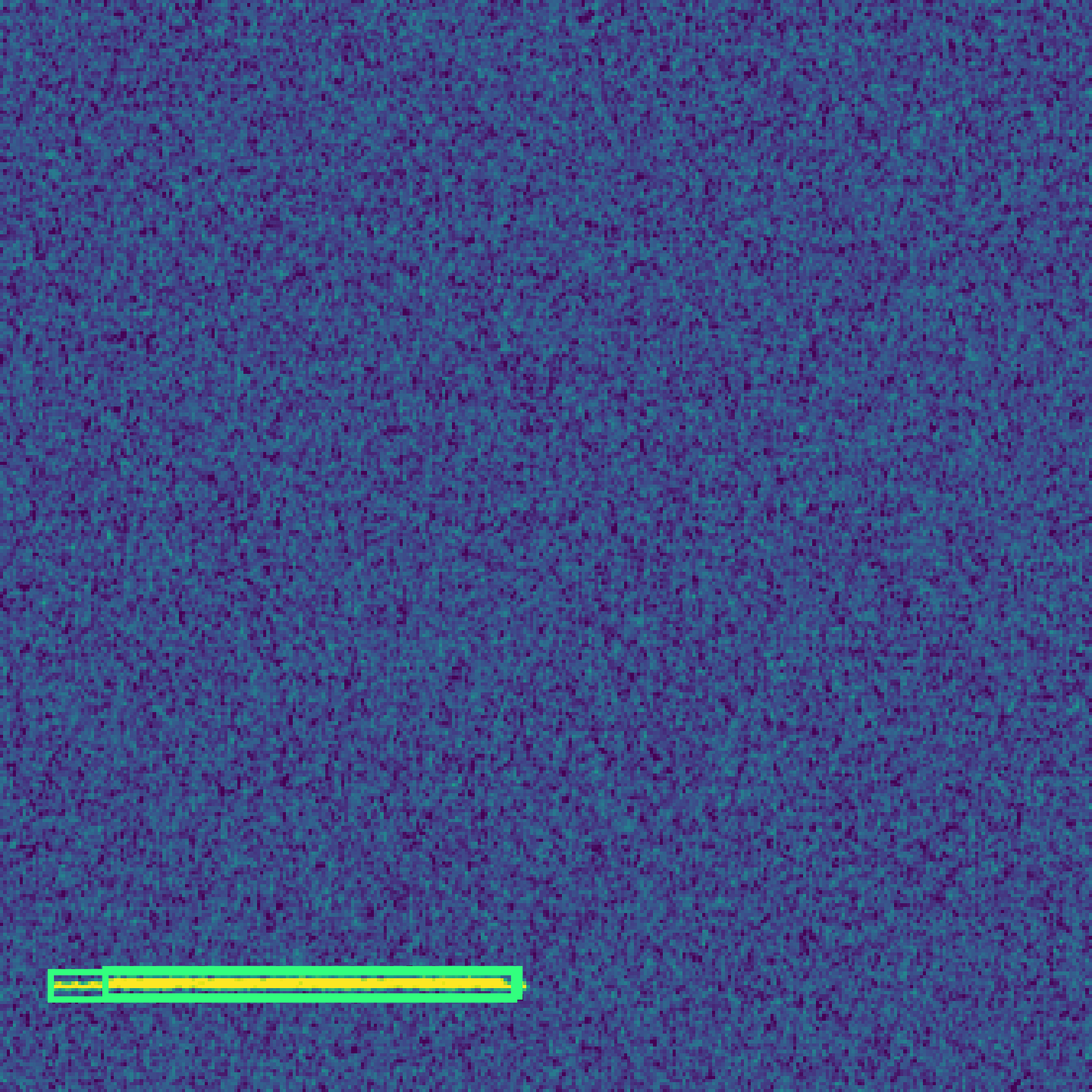} &
\includegraphics[width=0.192\linewidth]{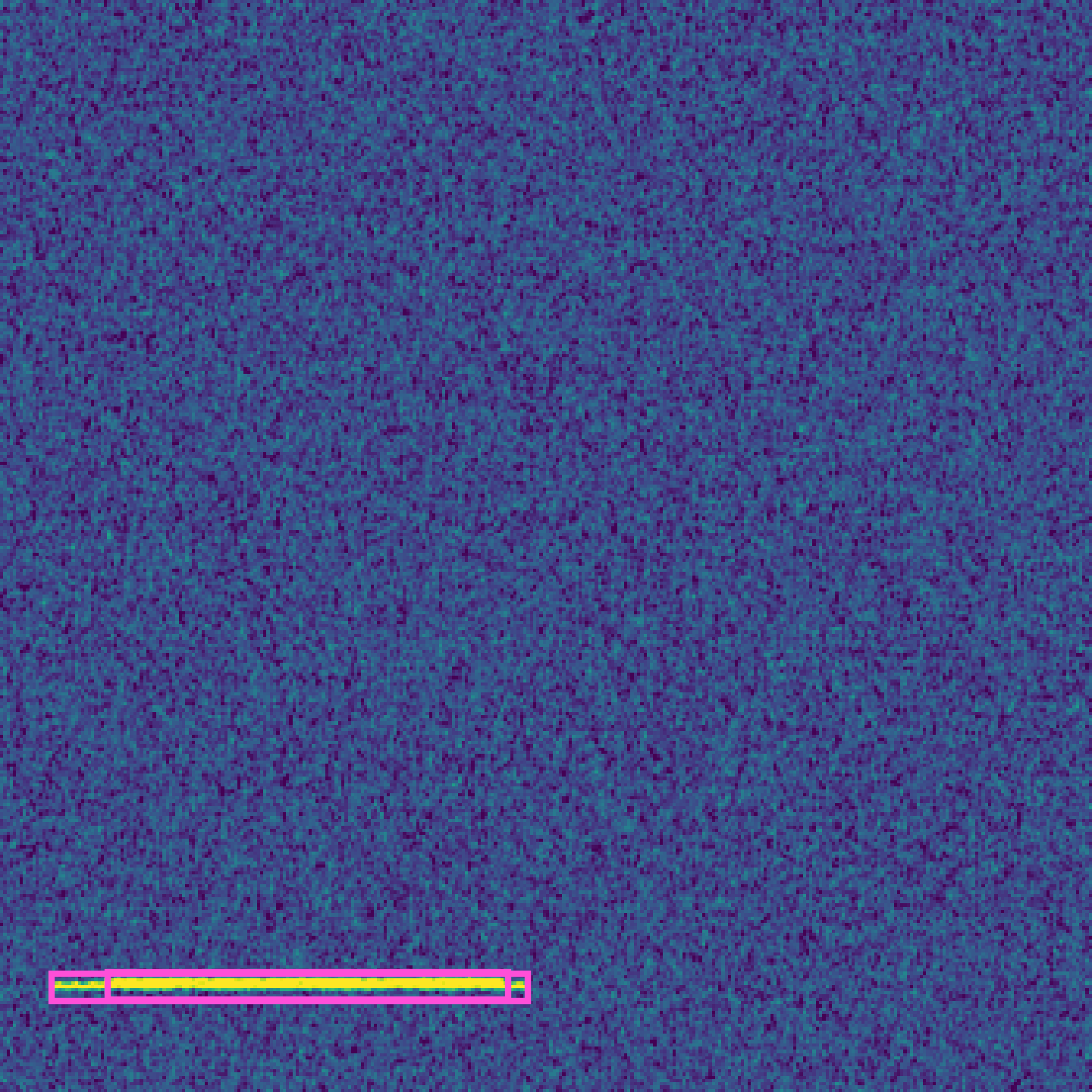} &
\includegraphics[width=0.192\linewidth]{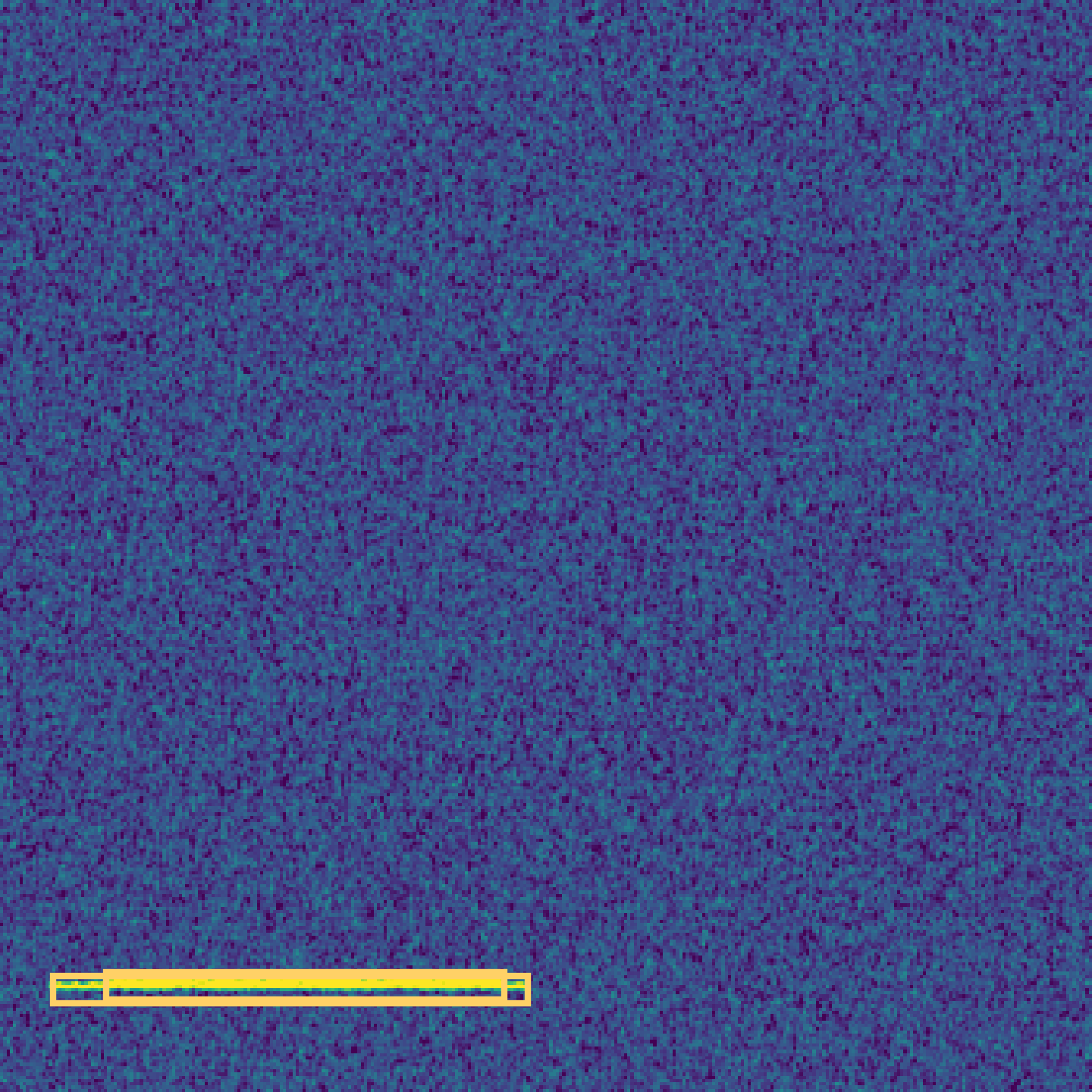} \\
\multicolumn{5}{c}{\scriptsize (a) Time-conditioned query} \\[2pt]
\includegraphics[width=0.192\linewidth]{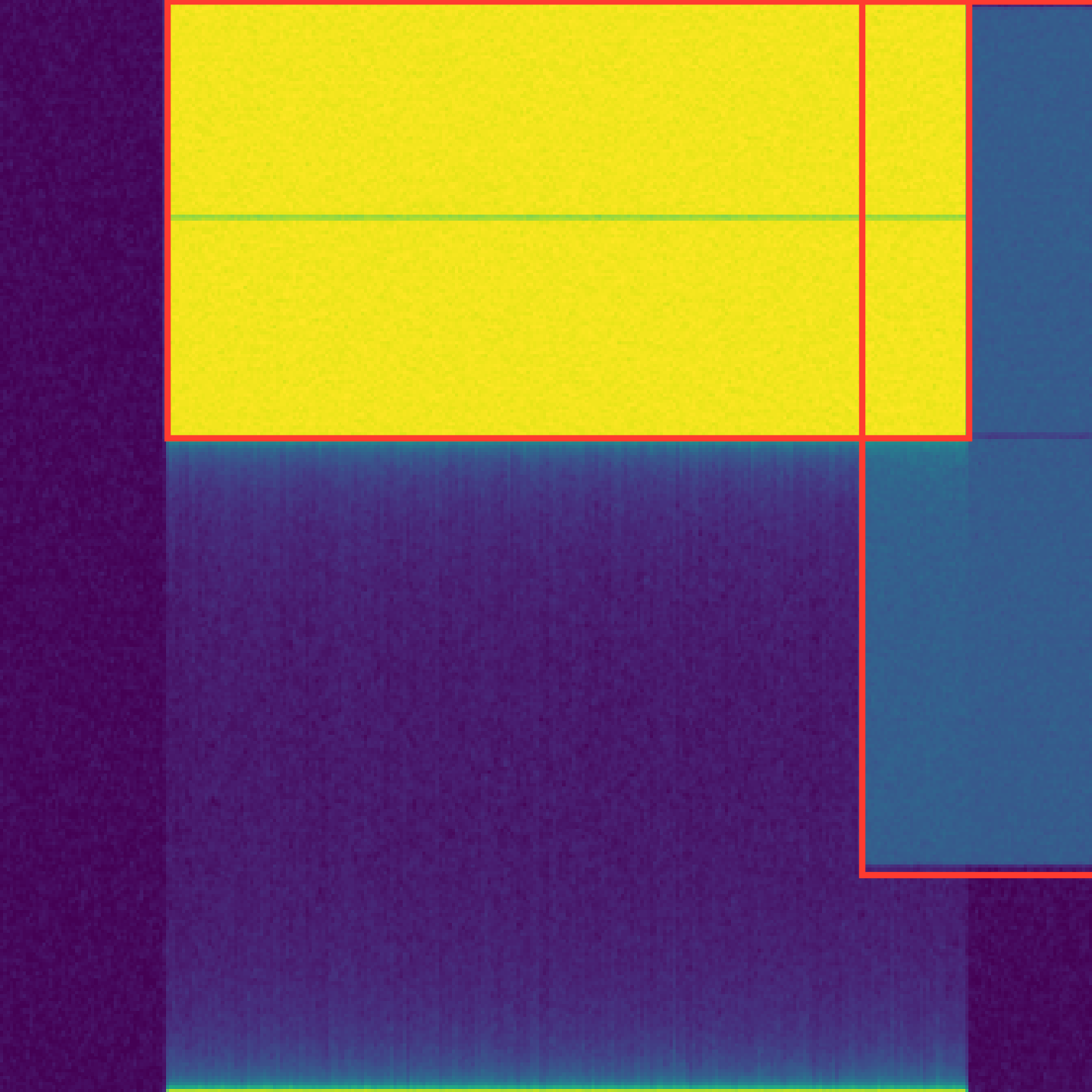} &
\includegraphics[width=0.192\linewidth]{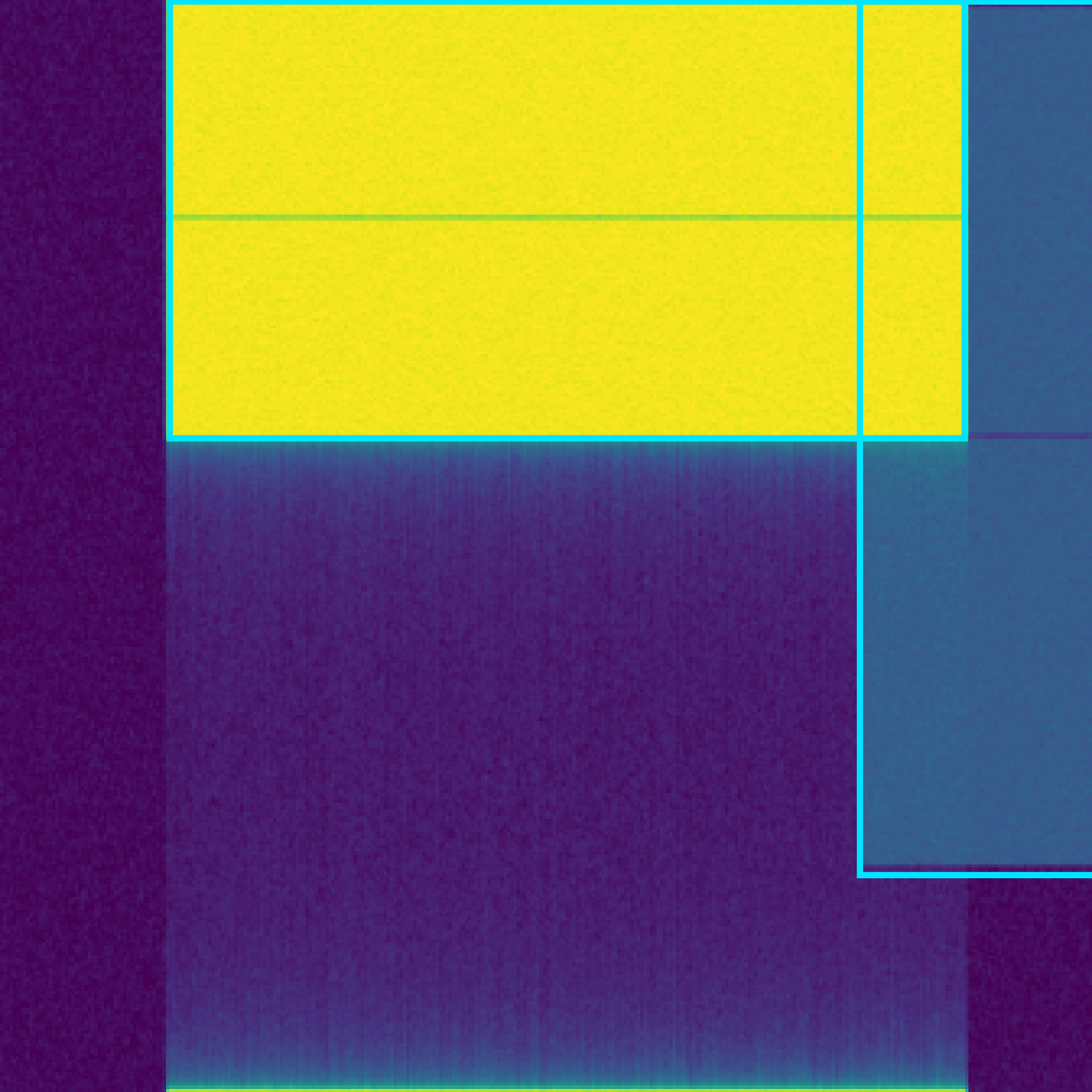} &
\includegraphics[width=0.192\linewidth]{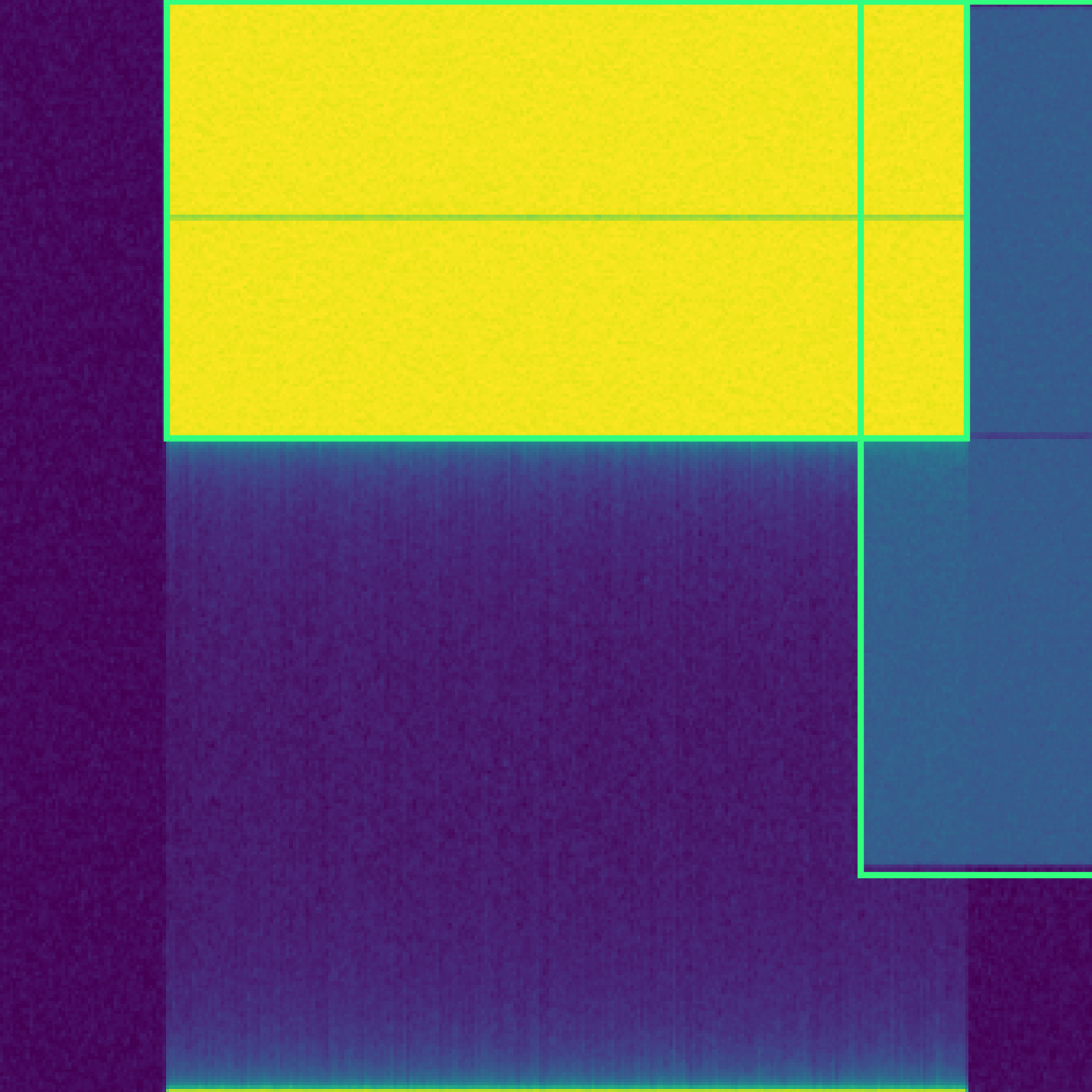} &
\includegraphics[width=0.192\linewidth]{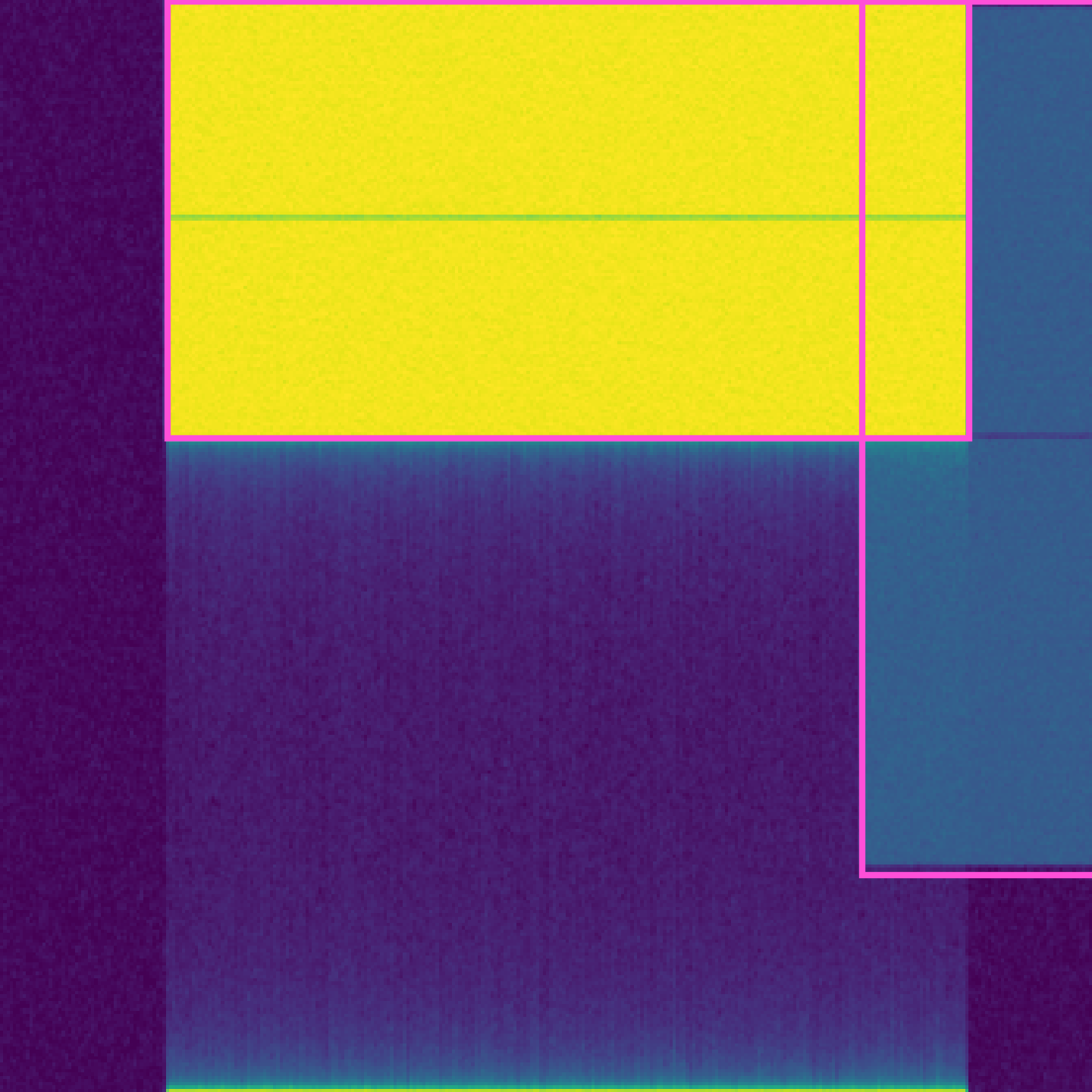} &
\includegraphics[width=0.192\linewidth]{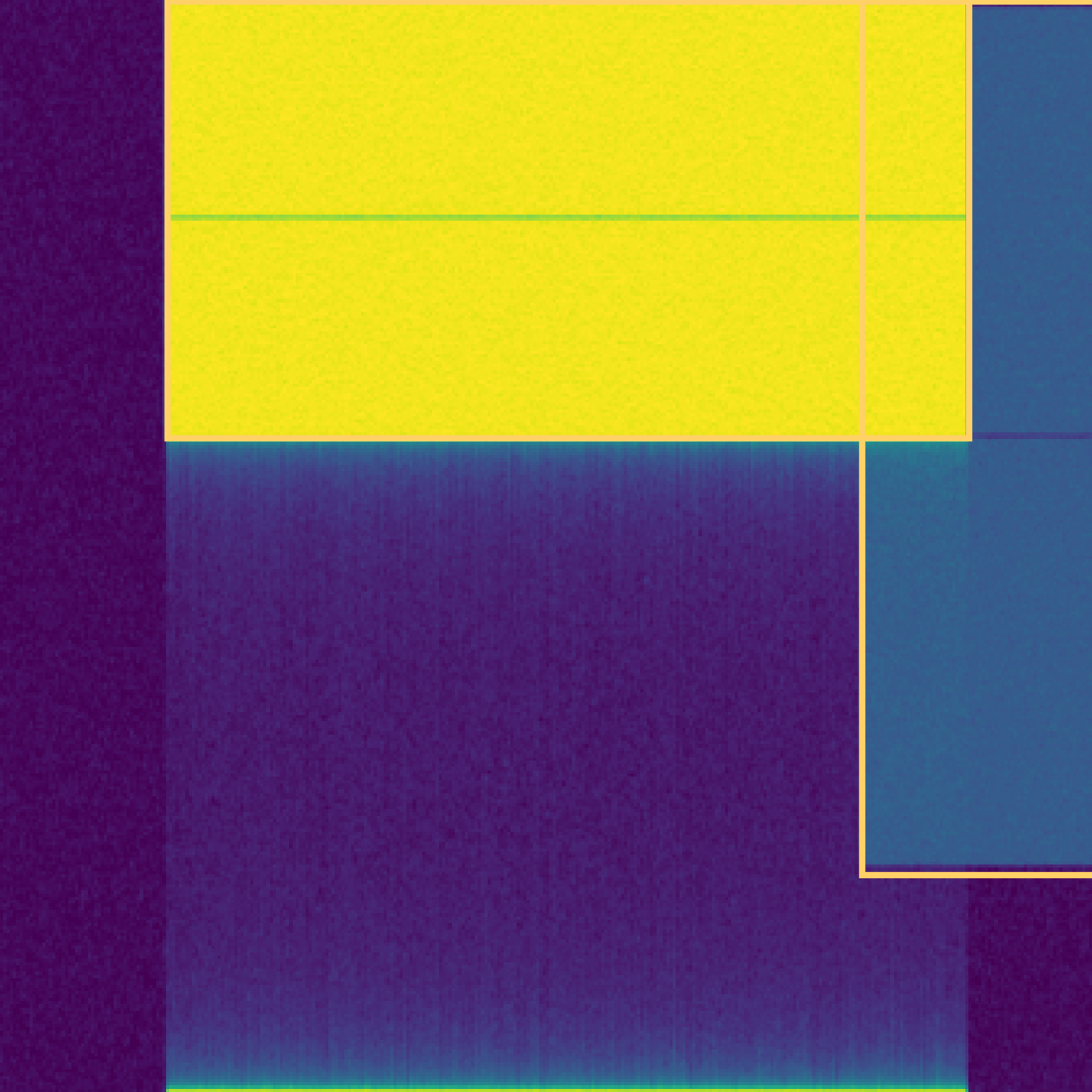} \\
\multicolumn{5}{c}{\scriptsize (b) Overlap-conditioned query}
\end{tabular}
\caption{Grounding under time and overlap constraints.
\textbf{(a) Question:} ``Find signal regions that overlap the time
interval 1.092--10.124\,ms.'' \textbf{GT:}
$[46,891,483,916]$ and $[96,886,463,911]$.
\textbf{(b) Question:} ``Find the signal regions involved in
time--frequency overlap.'' \textbf{GT:} $[152,0,886,400]$ and
$[788,0,1000,800]$. The overlap case is deliberately included because
all compared variants succeed on it. Coordinates are normalized to
$[0,1000]$; GT panels use red boxes.}
\label{fig:supp-grounding-b}
\end{figure}

The following cases vary the linguistic condition while keeping the
same normalized box-output format. Figures~\ref{fig:supp-grounding-a}
and \ref{fig:supp-grounding-b} cover class and frequency constraints,
followed by time and overlap relations.

Taken together, the grounding examples show that the models can map
class, frequency, time, and overlap conditions to localized signal
instances under a common output format. The paired predictions further
indicate that BATok primarily changes condition-dependent target
selection while retaining the time--frequency geometry available from
the image pathway. Figure~\ref{fig:supp-grounding-b} also shows that
the additional signal pathway preserves correct predictions when the
visual and linguistic evidence is already sufficient. These behaviors
are consistent with the quantitative results: raw I/Q tokens
complement waterfall images by supplying signal-level semantic
evidence, while the visual pathway remains effective for spatial
localization.

\end{document}